\documentclass[conference]{IEEEtran}
\IEEEoverridecommandlockouts
\usepackage{cite}
\usepackage{amsmath,amssymb,amsfonts}
\usepackage{algorithmic}
\usepackage{graphicx}
\usepackage{textcomp}
\usepackage{xcolor}
\usepackage{listings}
\usepackage{float}
\def\BibTeX{{\rm B\kern-.05em{\sc i\kern-.025em b}\kern-.08em
    T\kern-.1667em\lower.7ex\hbox{E}\kern-.125emX}}
\begin{document}

\title{Lighting and Rotation Invariant Real-time Detector Based on YOLOv5: Vehicle Wheel Detector}

\author{
\IEEEauthorblockN{Michael Shenoda}
\IEEEauthorblockA{\textit{College of Computing \& Informatics} \\
\textit{Drexel University}\\
Philadelphia, PA, United States \\
michael.shenoda@drexel.edu}
}

\maketitle

\begin{abstract}
Creating an object detector, in computer vision, has some common challenges when initially developed based on Convolutional Neural Network (CNN) architecture. These challenges are more apparent when creating model that needs to adapt to images captured by various camera orientations, lighting conditions, and environmental changes. The availability of the initial training samples to cover all these conditions can be an enormous challenge with a time and cost burden. While the problem can exist when creating any type of object detection, some types are less common and have no pre-labeled image datasets that exists publicly. Sometime public datasets are not reliable nor comprehensive for a rare object type. Vehicle wheel is one of those example that been chosen to demonstrate the approach of creating a lighting and rotation invariant real-time detector based on YOLOv5 architecture. The objective is to provide a simple approach that could be used as a reference for developing other types of real-time object detectors.
\end{abstract}

\section{Introduction}
Surprisingly humans are really good at understanding visual content of an image and  instantly provide information about the objects within. Unfortunately, this task is a hard problem in computer vision. Since 2012 Convolutional Neural Network (CNN) has became so popular for object detection and classification tasks, but yet the challenges to create reliable models can get overwhelming. Looking at the YOLO architecture, it has taken a leap forward as being the architecture of choice for real-time detectors. The original version of YOLO was designed for real-time object detection as primary focus. Then the latest version YOLOv5 has retained the same focus with the simplification of the development process of creating custom models with the power of image augmentation techniques. It provides the ability to visualize metrics, and get insights of changes made on each model update using integration with Weights and Biases machine learning tool. While the end result of a well designed model works well in terms of performance and accuracy, the amount of initial development can be daunting. The objective in this paper is to outline a general guideline to achieve a model that can be adaptive to lighting and rotation changes with minimal initial images samples. These guidelines are being presented with a practical example by implementing a model that is capable of detecting vehicle wheels in images captured with various camera orientations, and lighting conditions. After getting an overview of the YOLOv5 architecture and methodology, the basic approach is summarized as series of steps that can be easily followed:\\ \\
1. Understand the immediate use case of the detector. \\
2. Select the model size of YOLOv5 for the detector. \\
3. Understand the relationship of the object to the surroundings and how to propose specific views captured to optimize the visual appearance of the object.\\
4. Collect initial image samples to cover the various camera orientations and lighting conditions.\\ 
5. Train the initial model with the initial image samples, with weights transferred from an existing YOLOv5 model, then evaluate. \\ 
6. Collect 3D synthetic images to cover the variety of visual appearances of the object and also covers various rotations. \\
7. Use ground truth labels extracted from the 3D models, if available. Or use initial model to pre-label the 3D synthetic images, then manually review them to remove any false positives and fine tune any bounding box as necessary. \\
8. Train the model with the 3D synthetic images with weights transferred from initial training session then evaluate\\
9. Collect sample images from publicly available image datasets that represents the desired views of the object with various lighting conditions. \\
10. Use previous model to pre-label the public image samples, then manually review them to remove any false positives and fine tune any bounding box as necessary. \\
11. Train the model with the public image samples with weights transferred from previous training session then evaluate\\
12. Finally the model should be sufficient to automatically label images collected specifically for the use case.
\begin{figure}[H]
    \centering
    \includegraphics[width=\linewidth]{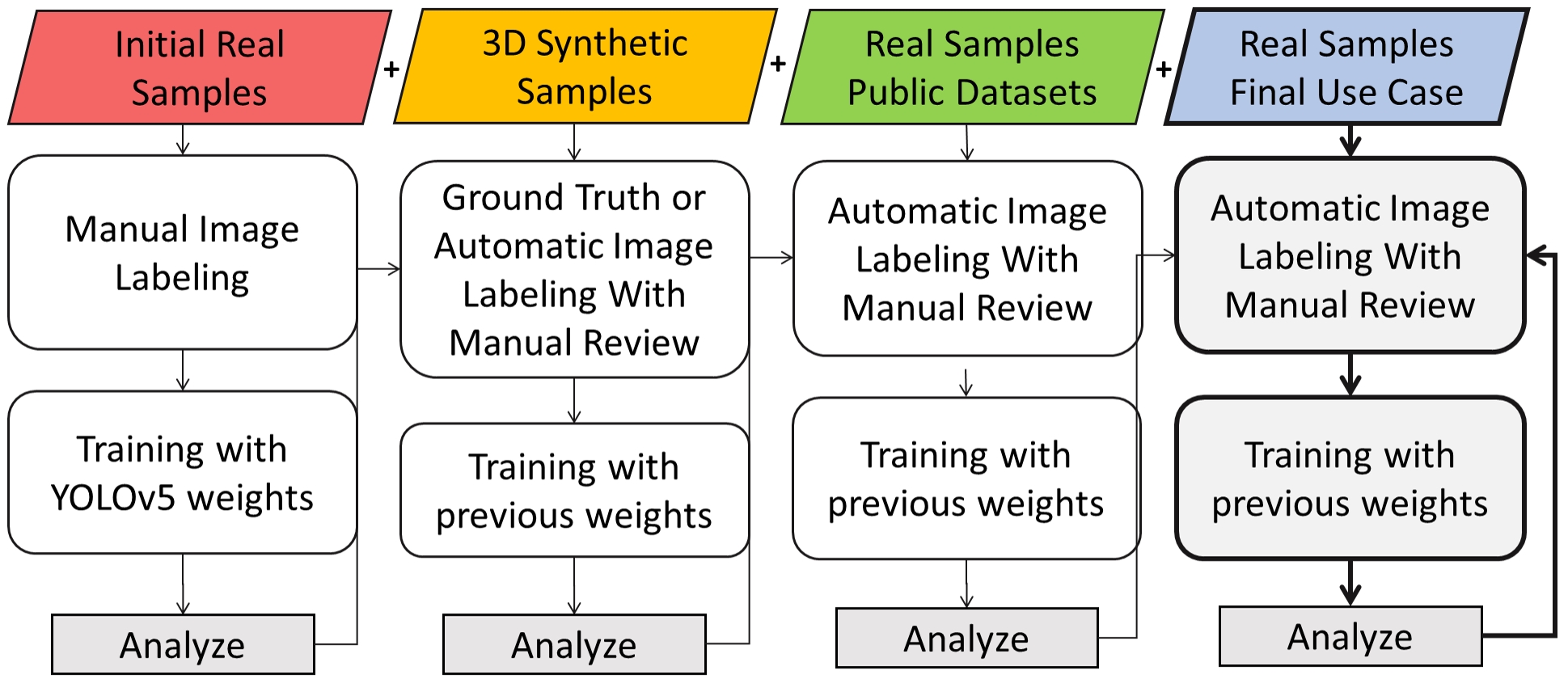}
    \caption{The Proposed General Guideline for Model Development}
    \label{fig:fig1}
\end{figure}

The work presented here is a humble approach towards a simple way to develop a custom model that can be easily prototyped and speed up the effort of creating an initial model that is reliable enough for providing machine labeled images with minimal manual review. The main goal is to allow the machine learning engineer to carefully craft the initial model than can be used later on to automatically label images for a non-technical person to manually review and perform any refinement to the bounding boxes. 

\section{Overview of YOLOv5}
YOLO is one of the best architecture and family of object detection models providing sate of the art performance with its focus on real-time detection and classification. YOLO for stands for You Only Look Once. The initial YOLO\cite{b1} was created by Joseph Redmon and later published YOLOv2\cite{b2} and YOLOv3\cite{b3} papers. Joseph's work was further advanced by Alexey Bochkovskiy who has published YOLOv4 in 2020. \cite{b4}
YOLOv5, created by Ultralytics, is a family of object detection architectures and models pretrained on the COCO dataset that represents their open-source research into computer vision AI methods, and incorporating lessons learned and best practices. It is a continuation on the great work that has been initially developed in the earlier versions\cite{b5}. It's important to note that YOLOv5 is built using PyTorch while previous YOLOv4 was built using Darknet. Using the PyTorch approach has simplified the development since it uses python by default.\\

The YOLOv5 model structure consists of two main blocks and the overall architecture is shown below:\\
\begin{figure}[h]
    \centering
    \includegraphics[width=\linewidth]{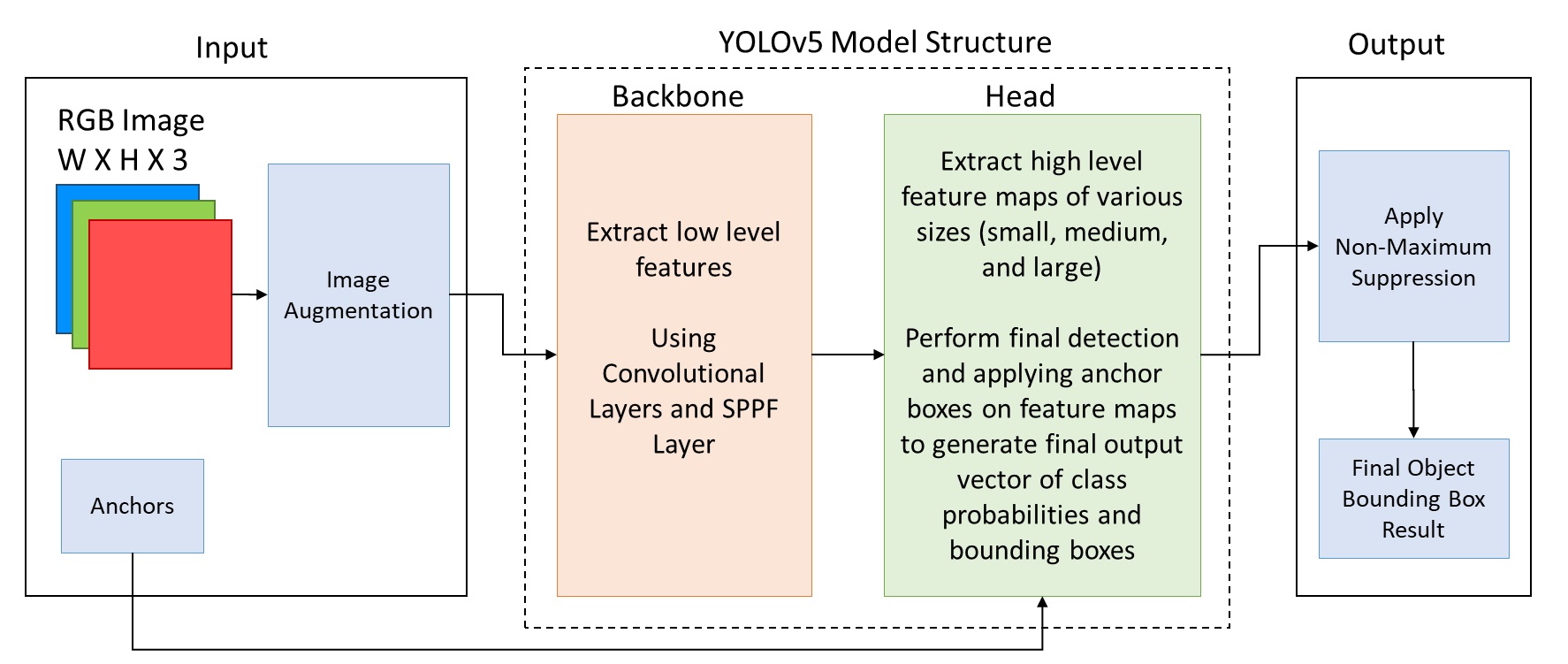}
    \caption{YOLOv5 architecture overview}
    \label{fig:fig2}
\end{figure}\\

The model consists of two parts, Backbone and Head.
Backbone is responsible for extracting low level features using convolutional layers in combination with Spatial Pixel Pair Features (SPPF) layer.\\
Head is responsible for extracting the high level feature maps and performing the detections, applying anchor boxes on the feature maps to generate the final output vector of classes probabilities and bounding boxes.\\

The YOLOv5 models originally came in four different sizes which are small, medium, large, and xlarge. The size of the model depends on the complexity of the detection and classification task. Recent version 6.0 has introduced the nano size which was primarily developed to be very lightweight to fit ultra small devices.
\begin{figure}[H]
    \centering
    \includegraphics[width=\linewidth]{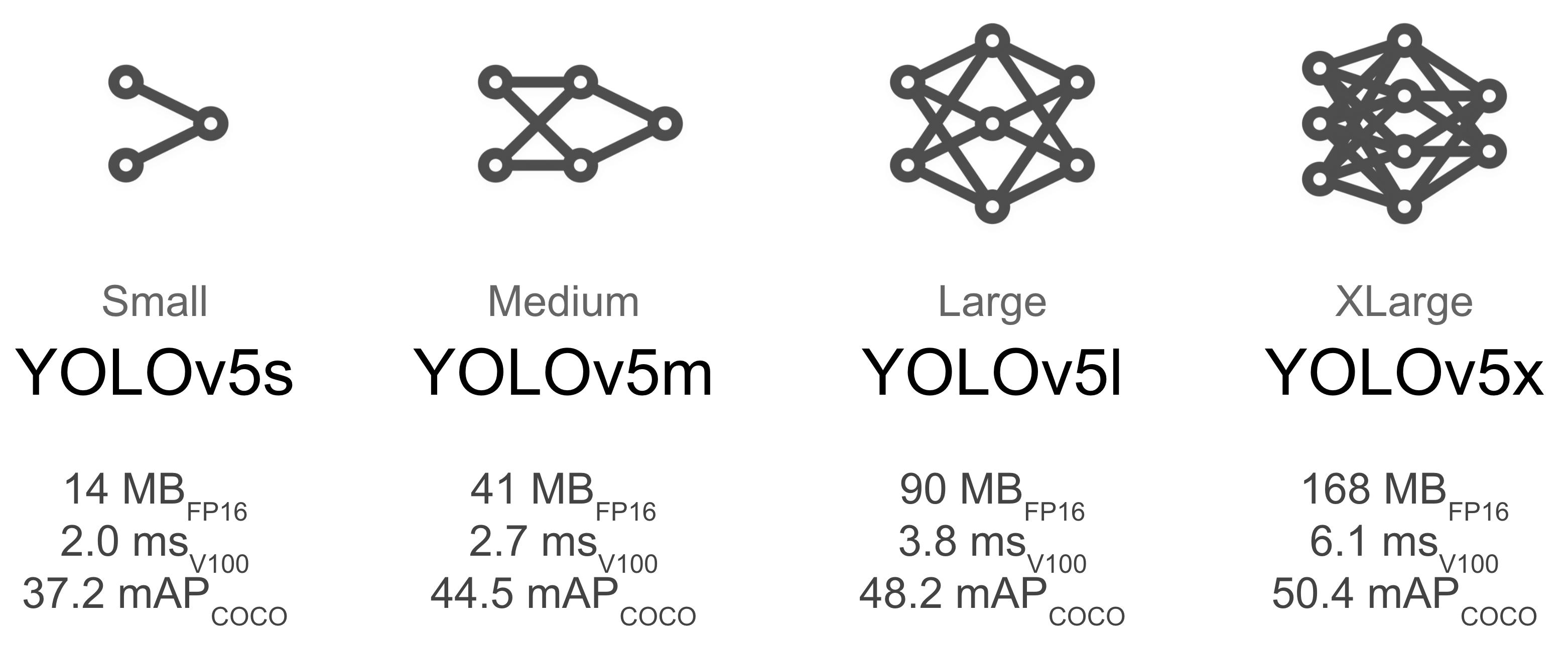}
    \caption{YOLOv5 model size comparison}
    \label{fig:fig3}
\end{figure}

The nice aspect about YOLOv5 that models are defined declaratively using yaml configuration files. Yaml configuration describes the model definition without the need to programmatically define it in Python. This aspect makes it superb for rapid model development. The other powerful aspect of YOLOv5 is the image augmentation capability that it provides, especially when starting with very small number of samples. \\

It's important to note when training custom dataset, a new model file needs to be created for the detector. One of the important parameters is the number of classes. In this case, we need to set to 1 for a simple detector of single object type like vehicle wheel.

\section{Proposed General Guideline of Model Development}
\subsection{Understanding the use case}
The proposed wheel detector has direct use case of providing vehicle axle count that can be used in application such as traffic analysis, and tolling systems. The end goal is to provide a reliable method of detecting the vehicle wheels regardless of camera orientation and lighting conditions. 
\begin{figure}[H]
    \centering
    \includegraphics[width=\linewidth]{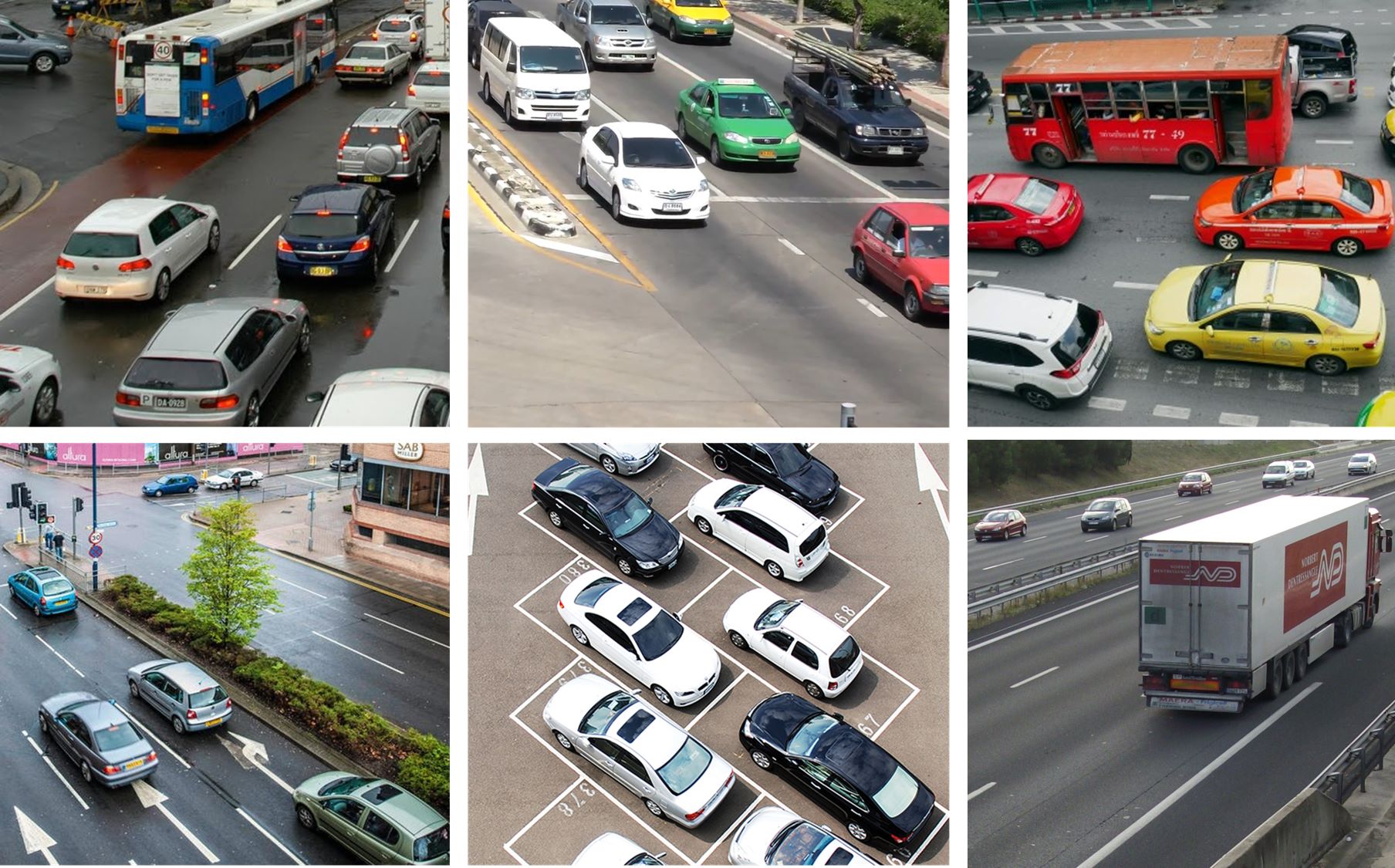}
    \caption{Wheel Detector Use Case}
    \label{fig:fig4}
\end{figure}
\subsection{Model Size Selection}
Based on the published performance of the different YOLOv5 model sizes, the medium size has been chosen for the wheel detector to provide a good combination of accuracy and performance. The medium size model definition file has been copied from the original yolov5m.yaml then modified the number of classes to be 1, since dealing with a single type.

Another important consideration during model selection is the available GPU and the training input image size. The GPU used in this case is Nvidia GeForce RTX 3050 Ti with available dedicated GPU memory of 4 GB. I choose 512x512 image size based on initial testing on what can be fit in the available GPU memory. By default YOLOv5 uses a square size for the input and I decided to keep it that way. The reason behind that is the uncertainty of the image size for the final use case of the detector. In this case, square input size minimizes the amount of pixel padding that needs to be added to a vertical or horizontal rectangular image thus maximizing the image content. Of course, if the final image size is known, it will be optimized to build a detector with input size that matches the aspect ratio of final deployment image size.
\begin{figure}[H]
    \centering
    \includegraphics[width=\linewidth]{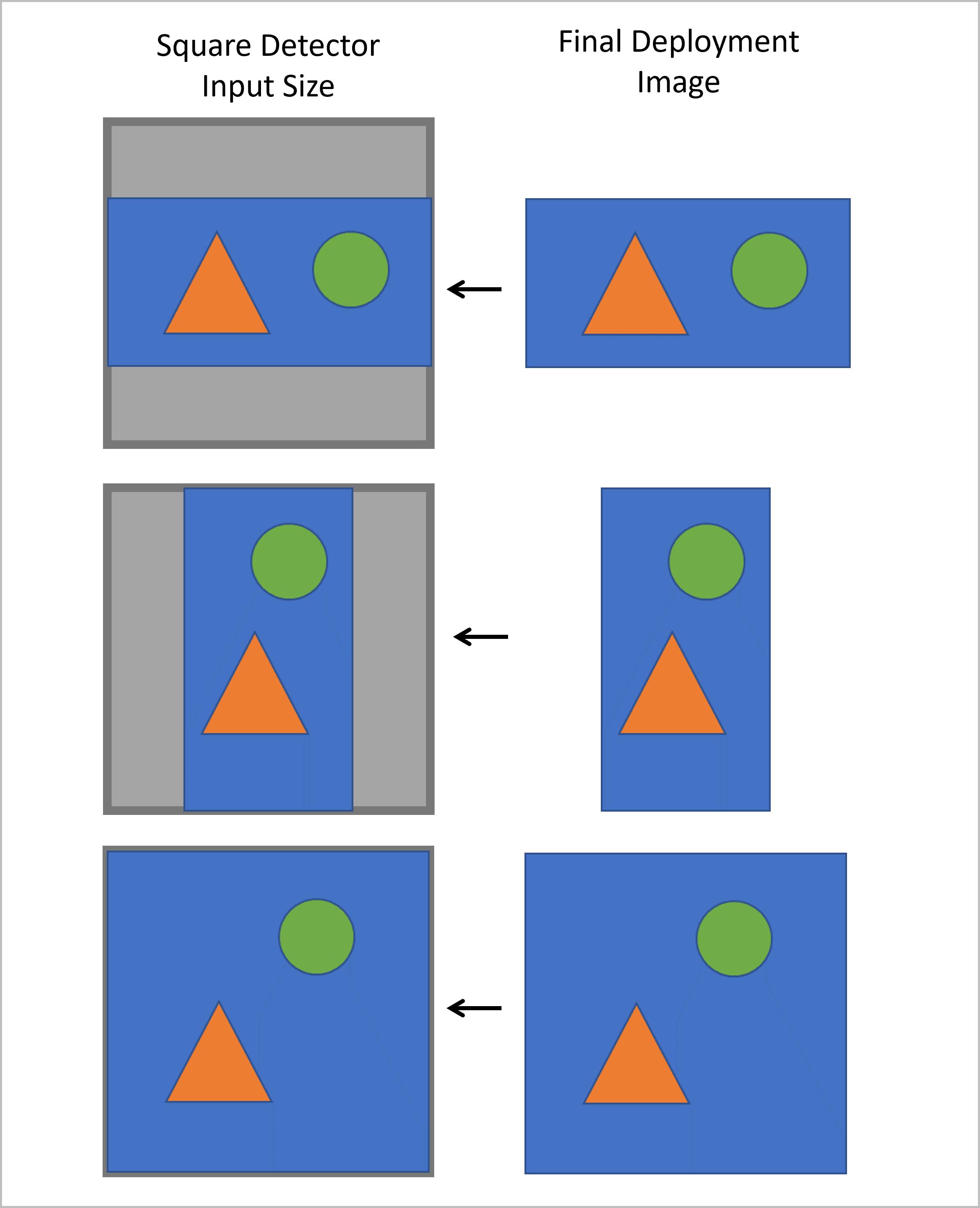}
    \caption{Reason for Using Square Detector Input Size}
    \label{fig:fig5}
\end{figure}
\begin{figure}[H]
    \centering
    \includegraphics[width=\linewidth]{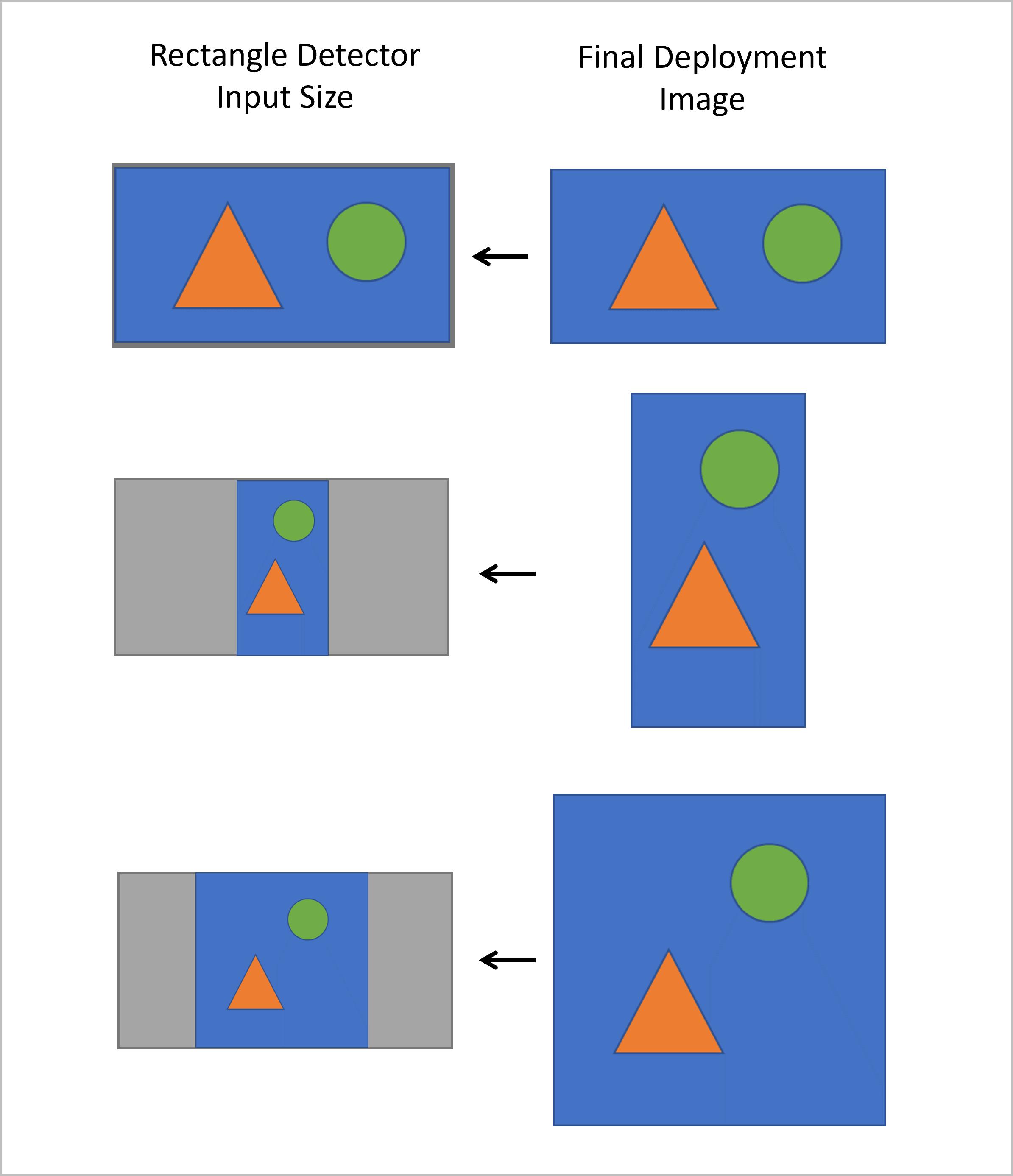}
    \caption{Reason for Not Using Rectangular Detector Input Size}
    \label{fig:fig6}
\end{figure}

\subsection{Understanding Visual Appearance of Object}
The appearance of the wheel can vary based on the camera orientation and different wheel types based on different vehicle types. Also, the positioning of the wheel while the vehicle is turning left or right. Collection of images to be able to cover from simple camera angles to complex scenarios. In addition, images samples should be collected to cover various image capturing quality. For example, samples need to cover low resolution, blurriness, and environmental changes such as fog, harsh sun light, harsh shadows, etc.\\

Considering the simple camera angles of straight on 90 degree, angled 45 degree, and steep angled 20 degrees as shown in the following figure: \\
\begin{figure}[H]
    \centering
    \includegraphics[width=\linewidth]{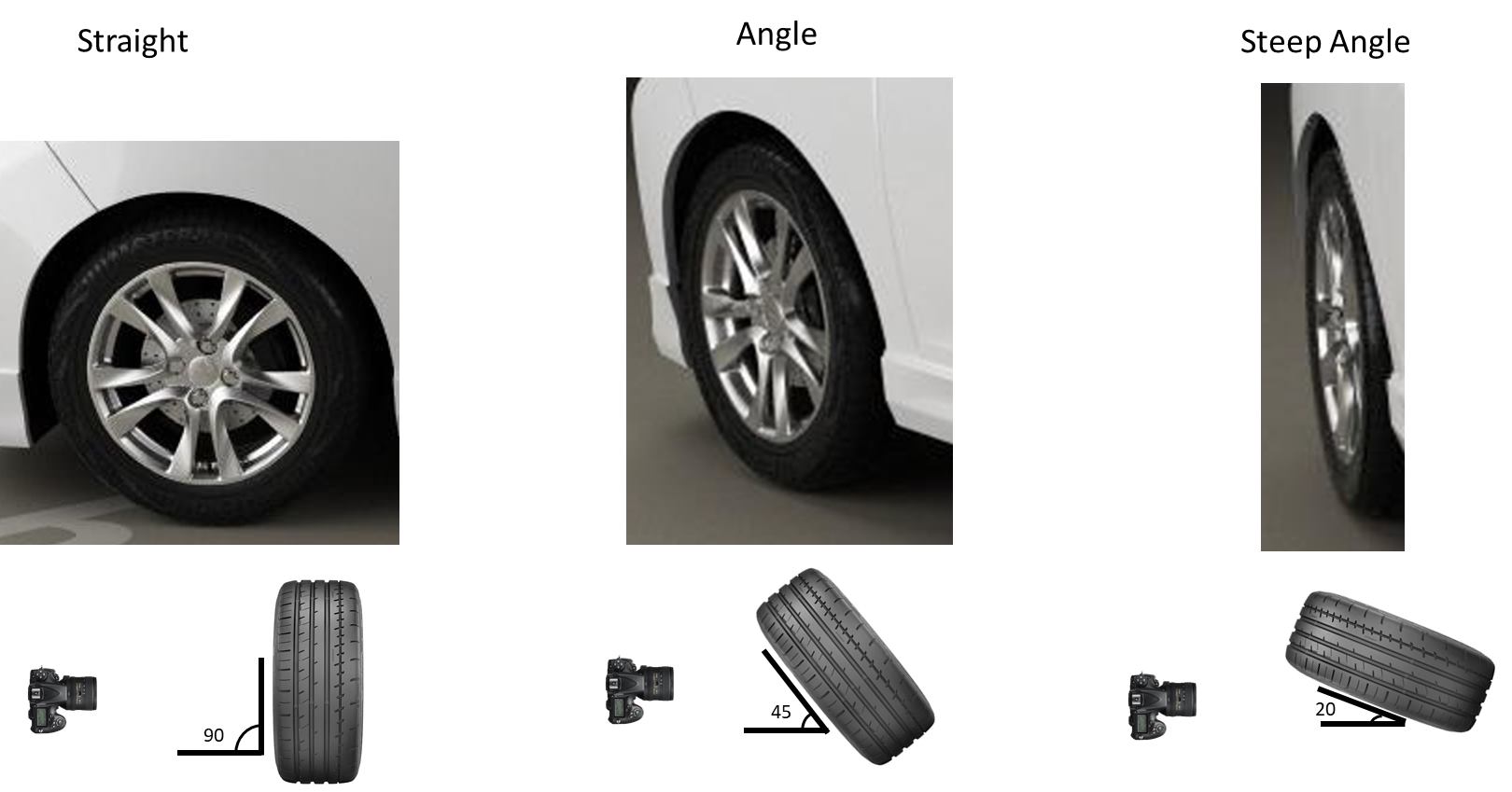}\\
    \caption{Simple Camera Angles}
    \label{fig:fig7}
\end{figure}
Considering the complex scenarios where the wheel could be occluded, partially visible, blurred, or ultra steep angle. Here are some of the examples:
\begin{figure}[H]
    \centering
    \includegraphics[width=\linewidth]{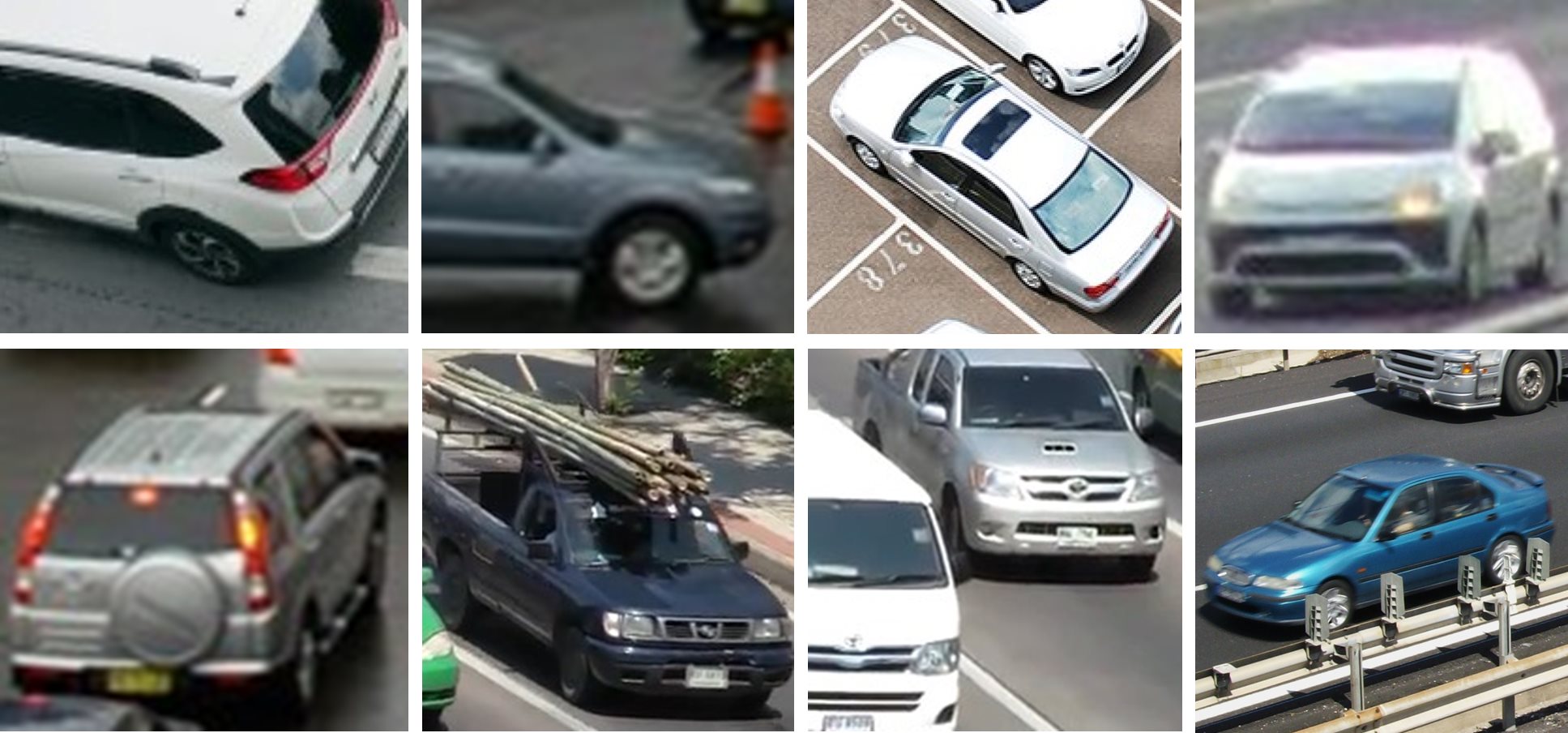}
    \caption{Complex Scenarios to Consider}
    \label{fig:fig8}
\end{figure}
\subsection{Training Initial Model and Evaluation}
Initial image samples have been manually captured to cover various camera orientation as well as slight shadows and sun light conditions. Total initial samples are 72 images, below are some of the images to visualize.
\begin{figure}[H]
    \centering
    \includegraphics[width=\linewidth]{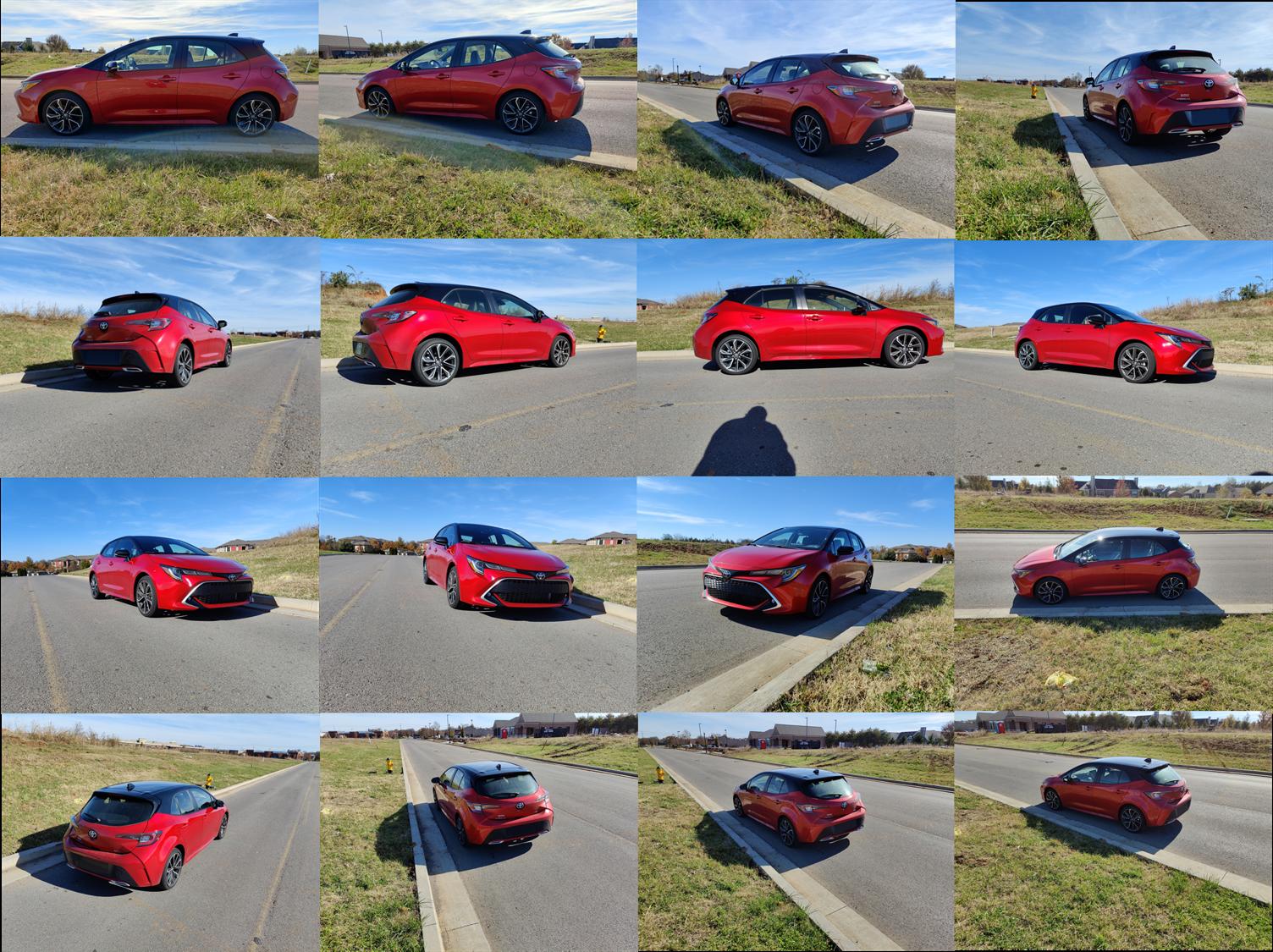}
    \caption{Initial Image Samples Part1 (total of 49 samples)}
    \label{fig:fig9}
\end{figure}
\begin{figure}[H]
    \centering
    \includegraphics[width=\linewidth]{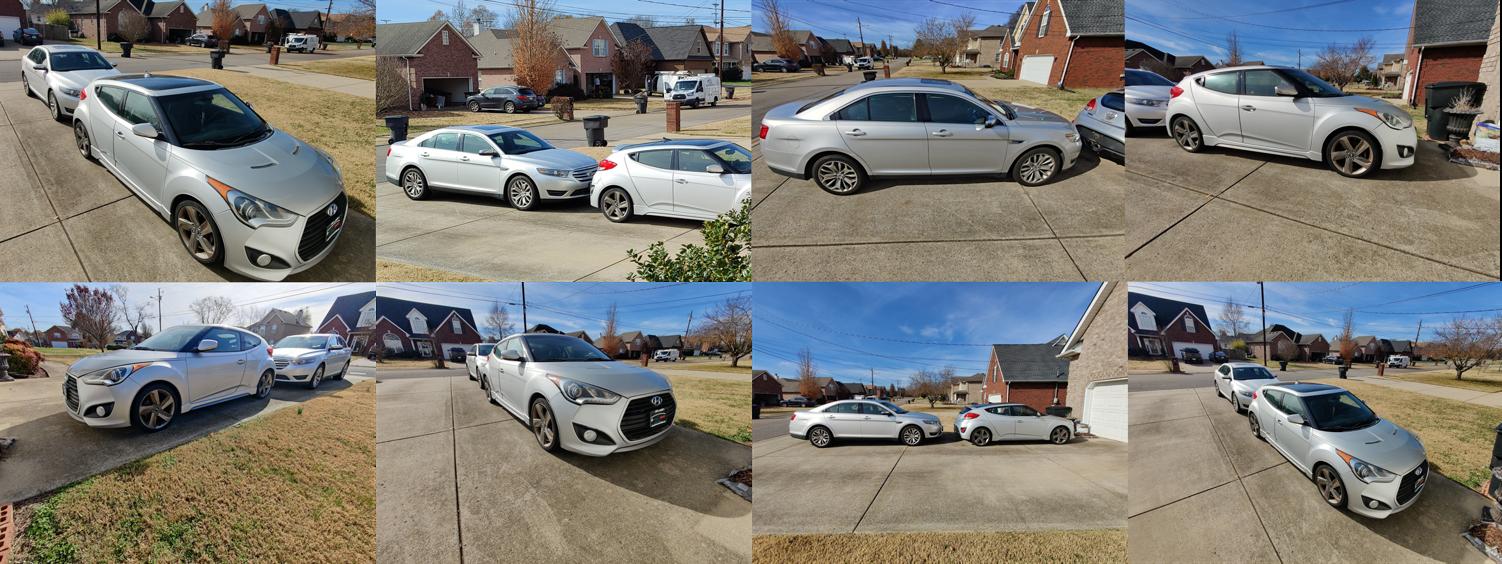}
    \caption{Initial Image Samples Part2 (total of 23 samples)}
    \label{fig:fig10}
\end{figure}
The initial image samples have been manually labeled using LabelImg \cite{b7}, a python tool for labeling images using bounding boxes that supports YOLO bounding box format. YOLO label format for an object is defined as the following:\\
(\textbf{object-class}) (\textbf{x}) (\textbf{y}) (\textbf{width}) (\textbf{height})\\ 
\textbf{object-class} is a numerical index of the object class, which in out case it's 0, for single object.\\
\textbf{x} and \textbf{y} represent the object center and are expressed as relative pixel position.\\
\textbf{width} and \textbf{height} represent the width and height of the object, and expressed as relative size.\\
It uses relative pixel position and relative size to accommodate different scaling of the image without worrying about absolute numbers [1]. 
\begin{figure}[H]
    \centering
    \includegraphics[width=\linewidth]{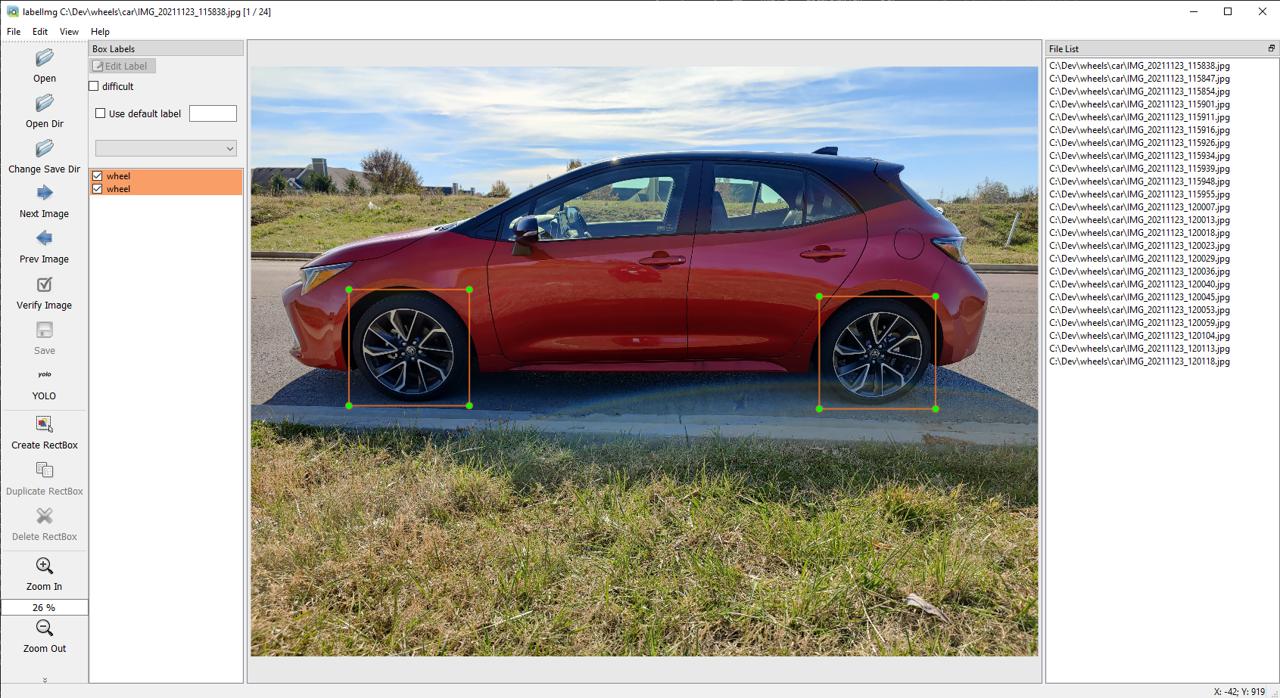}
    \caption{Image Labeling using LabelImg}
    \label{fig:fig11}
\end{figure}
The initial model has been trained with the following parameters:\\
input size = 512\\
batch size = 4\\
model weights = yolov5m.pt\\
validation spit ratio = 0.22 \\

Let's evaluate the initial model starting with the metrics: \\
\begin{figure}[H]
    \centering
    \includegraphics[width=\linewidth]{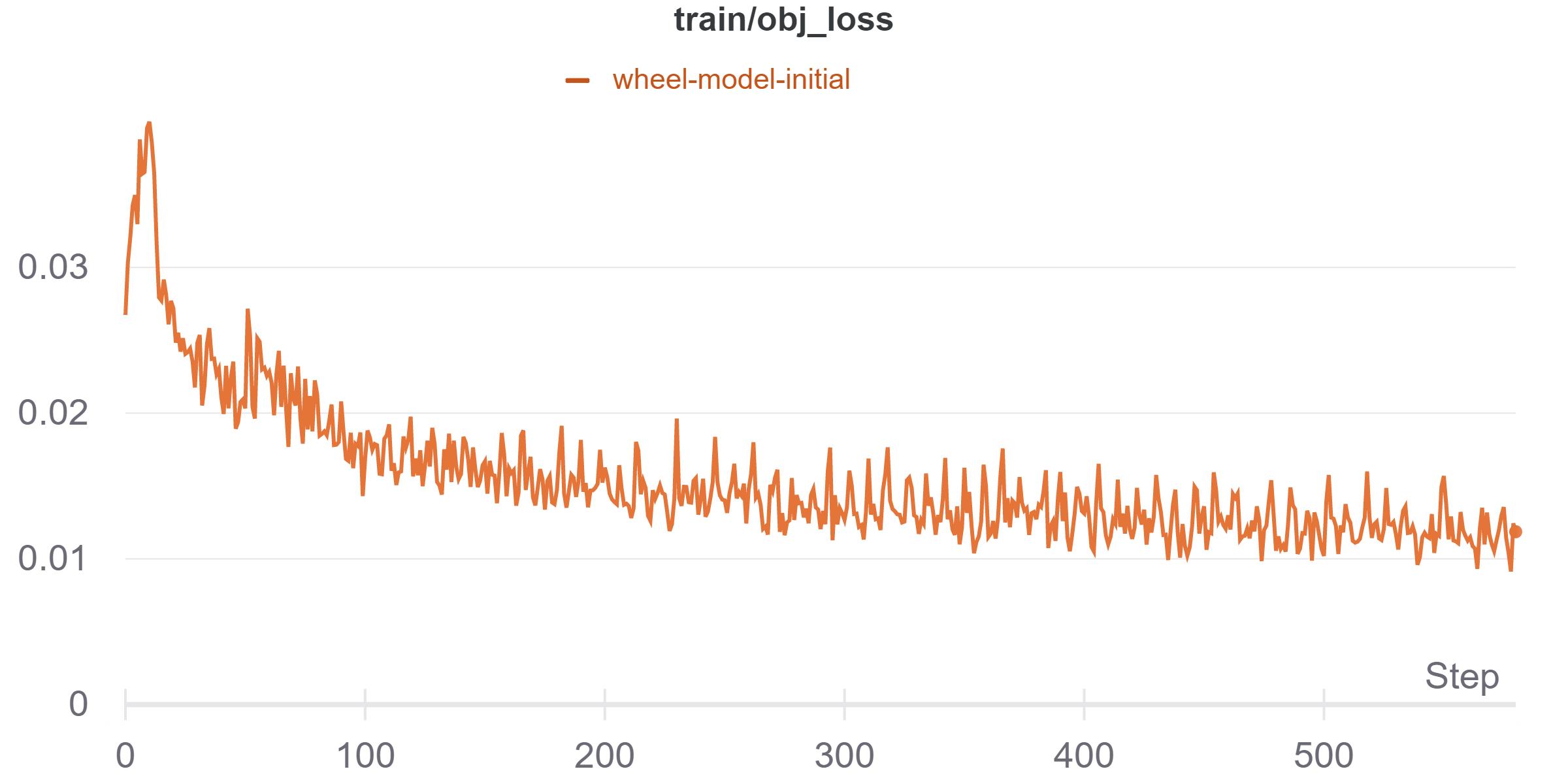}
    \caption{Initial Model Training Object Loss}
    \label{fig:fig12}
\end{figure}
\begin{figure}[H]
    \centering
    \includegraphics[width=\linewidth]{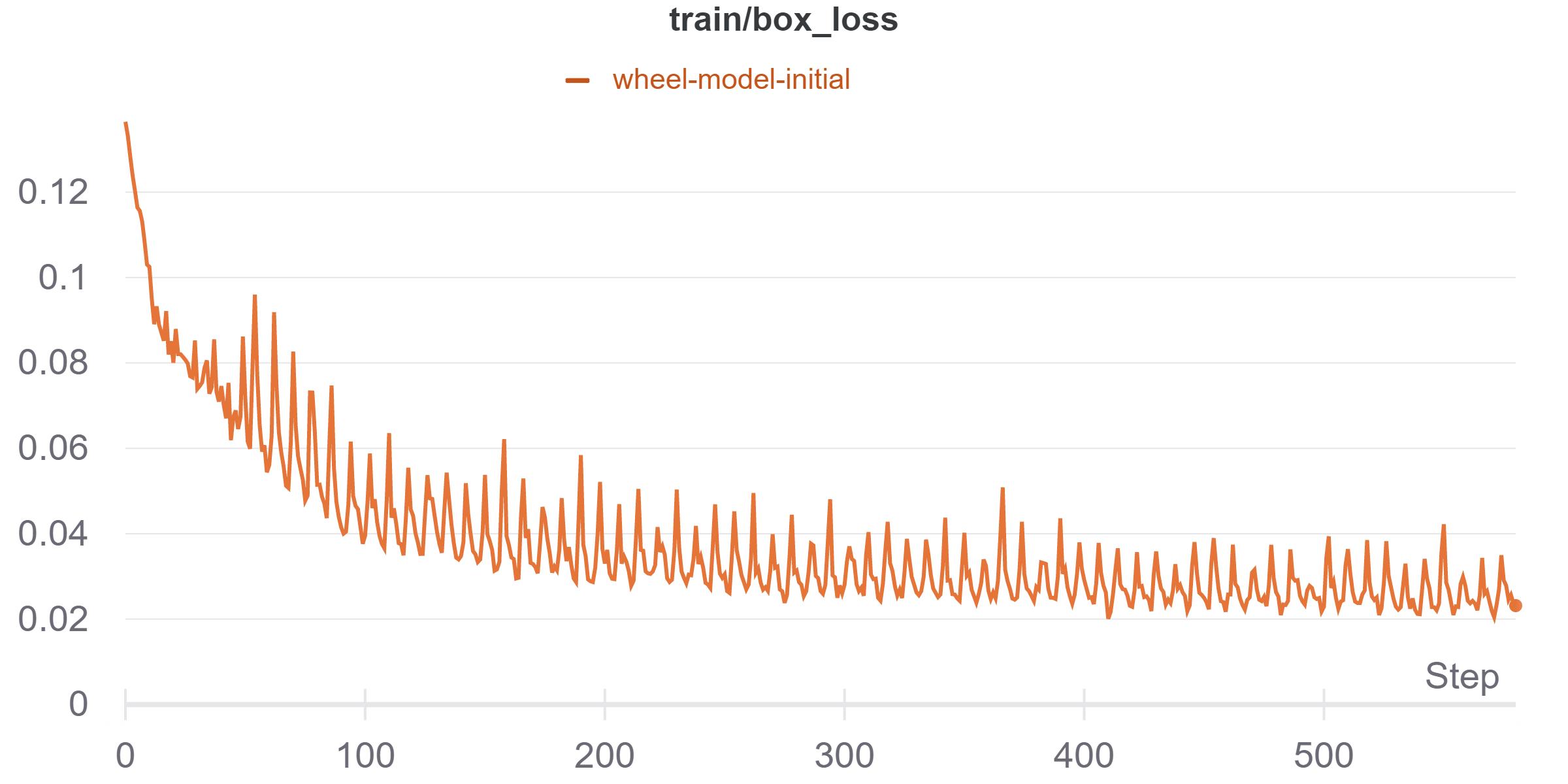}
    \caption{Initial Model Training Box Loss}
    \label{fig:fig12.2}
\end{figure}
\begin{figure}[H]
    \centering
    \includegraphics[width=\linewidth]{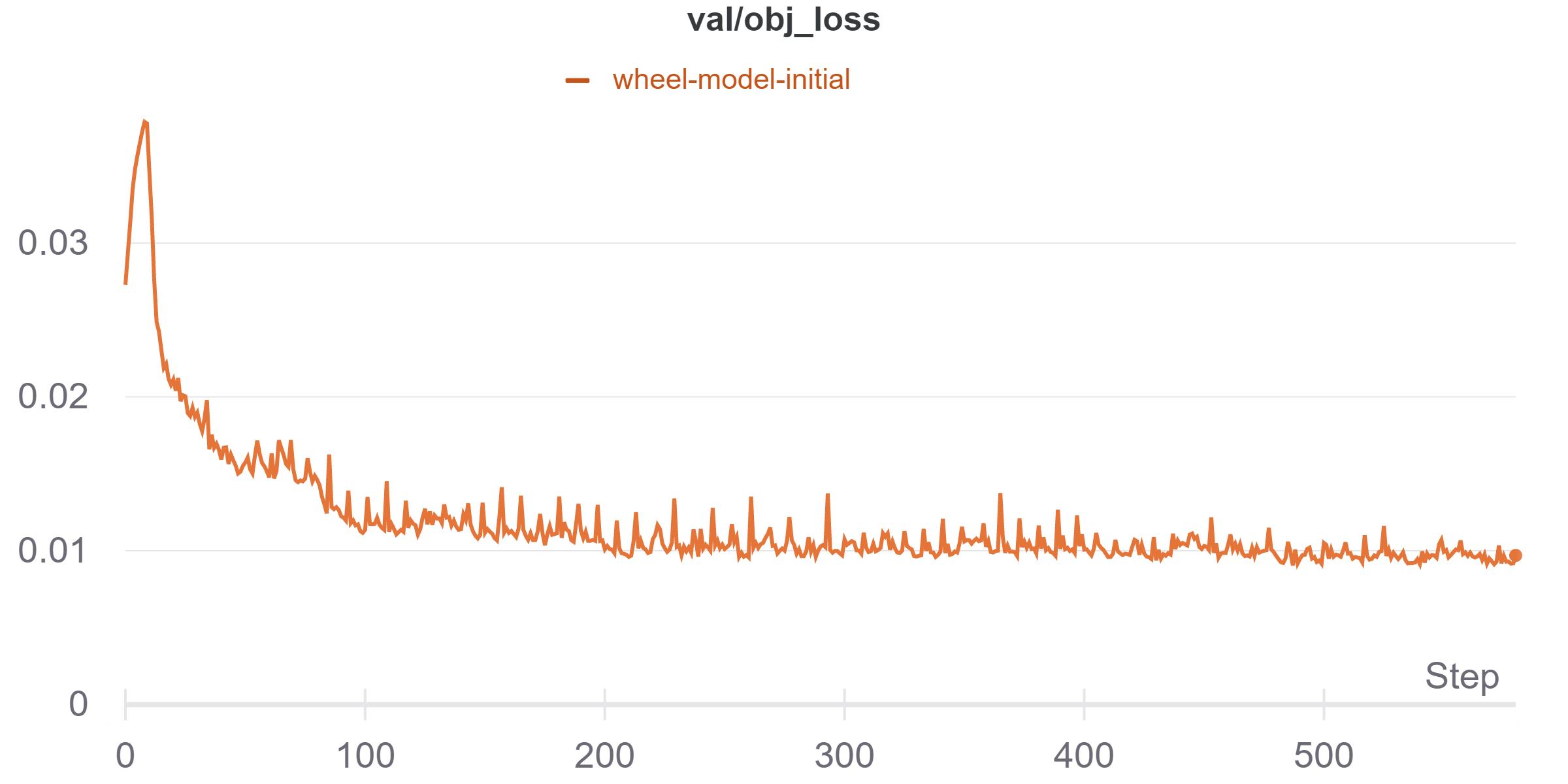}
    \caption{Initial Model Validation Object Loss}
    \label{fig:fig13}
\end{figure}
\begin{figure}[H]
    \centering
    \includegraphics[width=\linewidth]{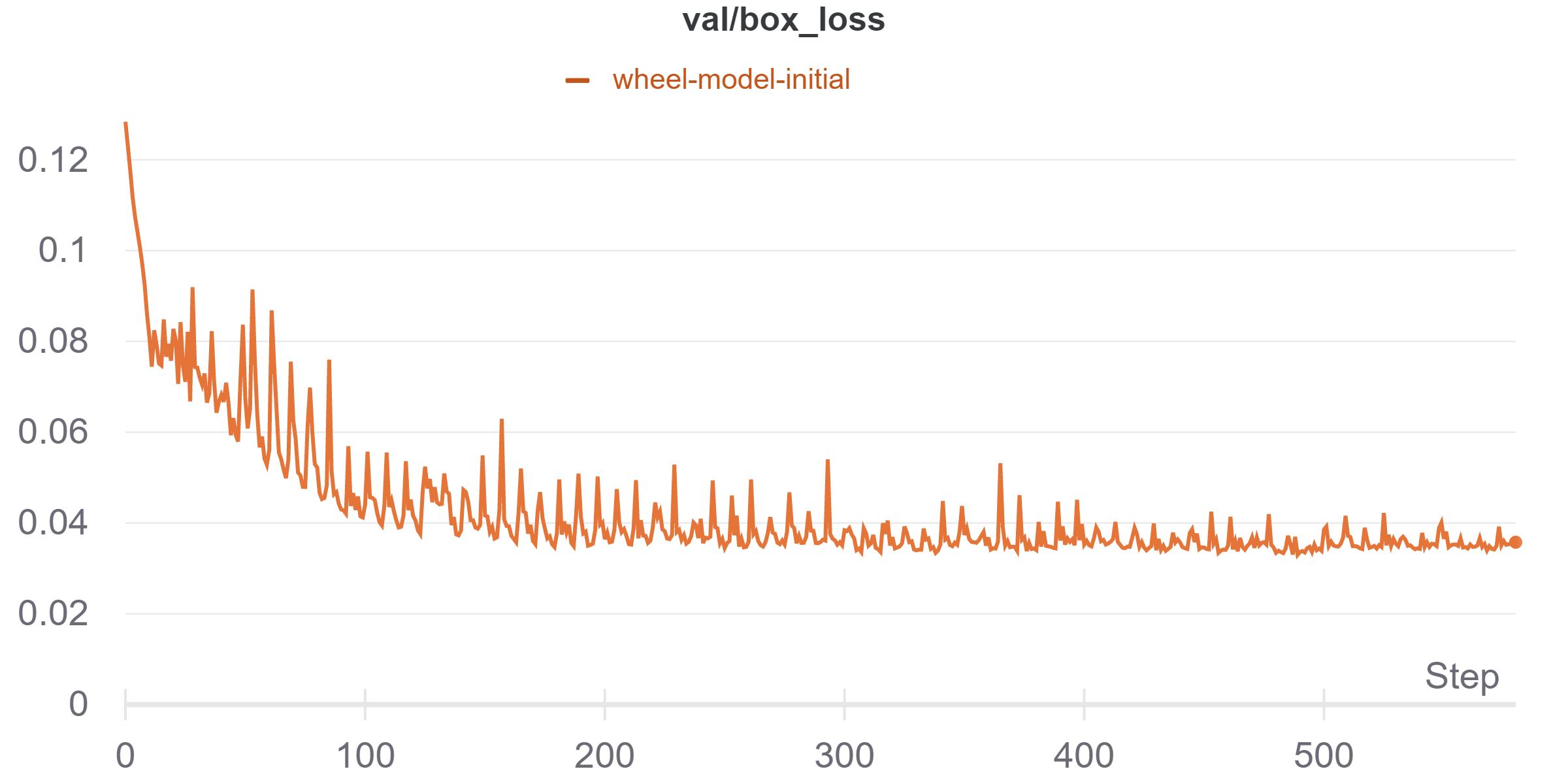}
    \caption{Initial Model Validation Box Loss}
    \label{fig:fig13.2}
\end{figure}
\begin{figure}[H]
    \centering
    \includegraphics[width=\linewidth]{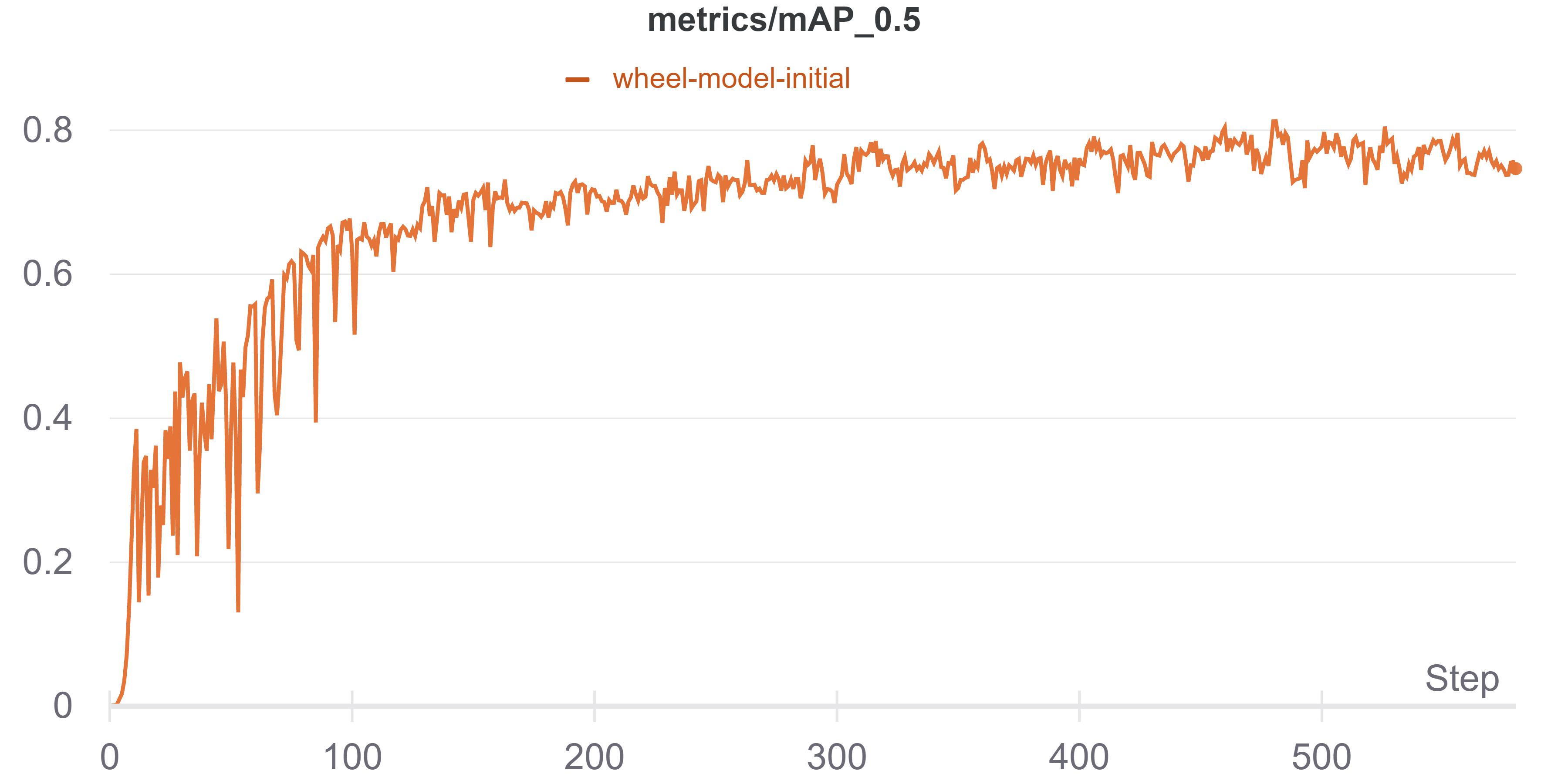}
    \caption{Initial Model Mean Average Precision}
    \label{fig:fig14}
\end{figure}
\begin{figure}[H]
    \centering
    \includegraphics[width=\linewidth]{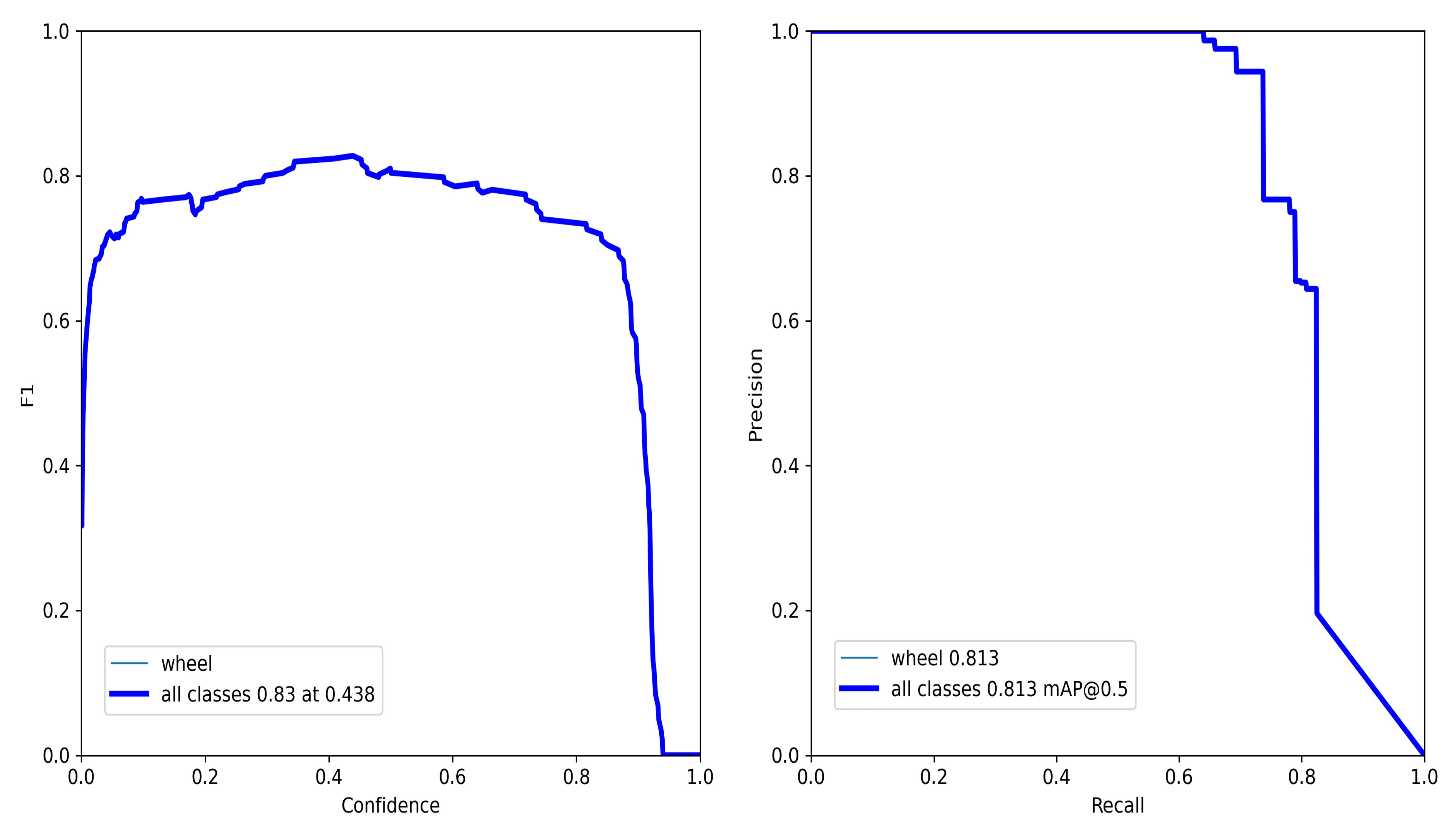}
    \caption{Initial Model F1 and Precision vs Recall Curves}
    \label{fig:fig15}
\end{figure}
The metrics looks reasonable for the initial training. The training and validation loss seems consistent, but the mean average precision is only hovering around 0.8. 
Let's check the detection performance on random samples from validation set.
\begin{figure}[H]
    \centering
    \includegraphics[width=\linewidth]{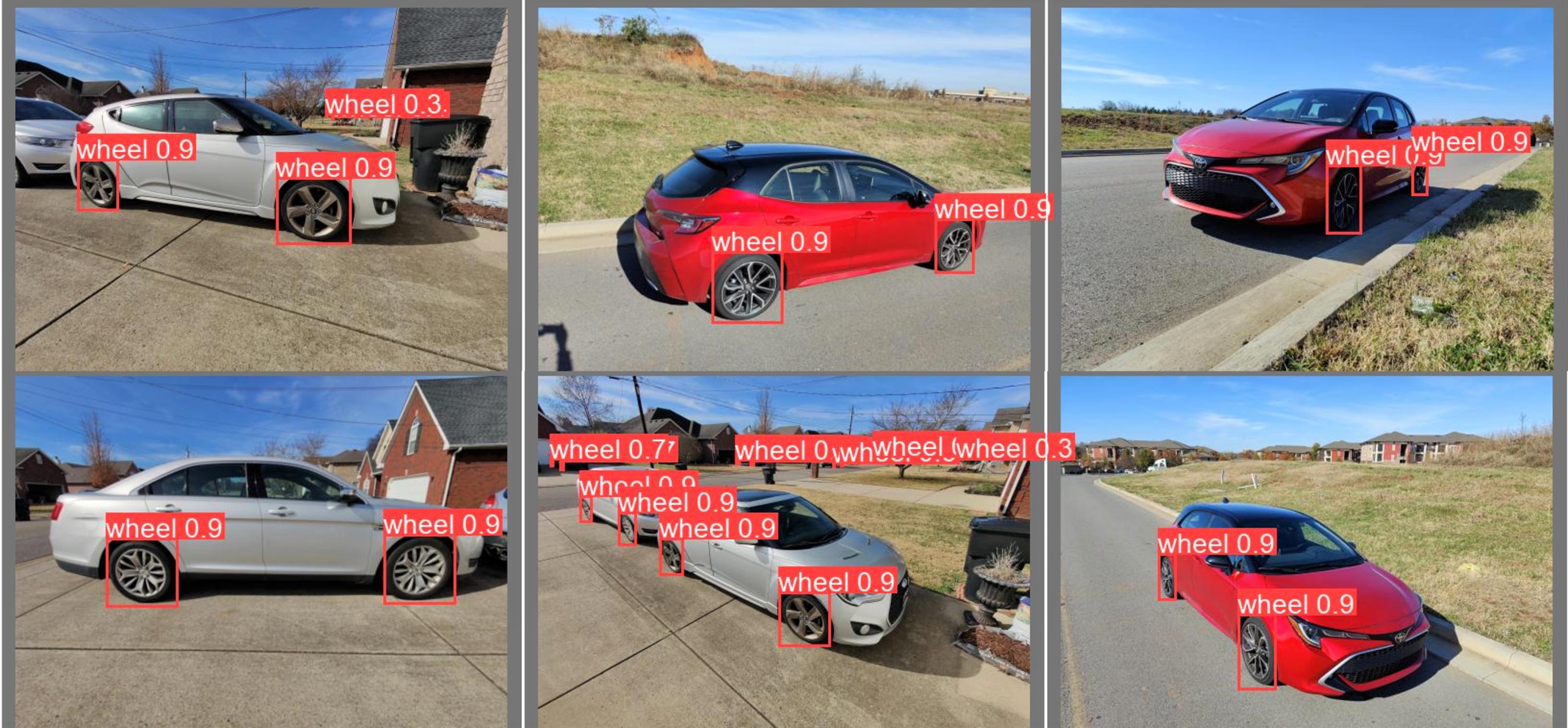}
    \caption{Initial Model Detection on Validation Set}
    \label{fig:fig16}
\end{figure}
The model does good job of detecting the wheels on the validation set. Before getting overexcited, we need to consider that the validation set is not far off from how the training set looks visually. We will need to test how the model would perform on the 3D synthetic images, which weren't part of the training yet.
That should give a better insight of how the model would behave on images that it hasn't seen before during the training.
\begin{figure}[H]
    \centering
    \includegraphics[width=\linewidth]{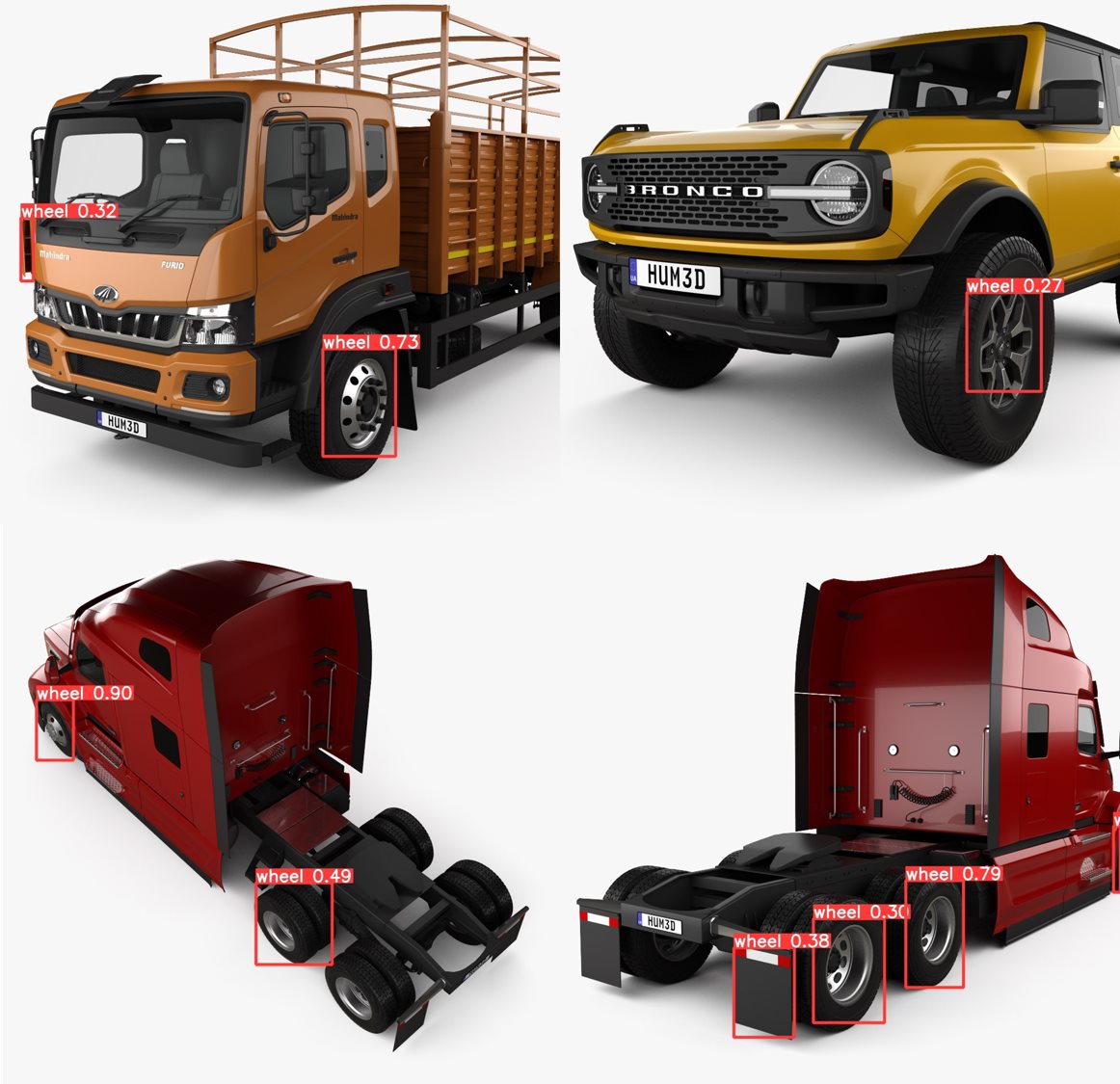}
    \caption{Initial Model Detection on 3D Synthetic Images}
    \label{fig:fig16.2}
\end{figure}
We can clearly see that the model has false positive on other parts of the vehicles. It detects truck mirror as a wheel and also mud flap of the semi truck. Also, some of the bounding boxes are not precisely localizing the wheel.

\subsection{Training with Synthetic Images and Evaluation}
3D synthetic image samples have been collected to cover various camera orientation with the goal of providing different varieties of vehicle wheel shapes. Total synthetic samples collected 165 images. Below are some of the images to visualize. The 3D synthetic images here are just yet another 2D images that were gathered from Hum3D \cite{b8}. Ideally, we would use ground the truth labels wheel labels extracted from the 3D model and unnecessary for labeling those images by machine or human. Unfortunately, due to lack of time for learning a 3D simulation software that provides such capabilities, such as Carla Simulator, I decided to not to pursue that route.  In this case, the wheel have been automatically labeled by machine using the initial model then manually reviewed and corrected using LabelImg \cite{b7} annotation tool.
\begin{figure}[H]
    \centering
    \includegraphics[width=\linewidth]{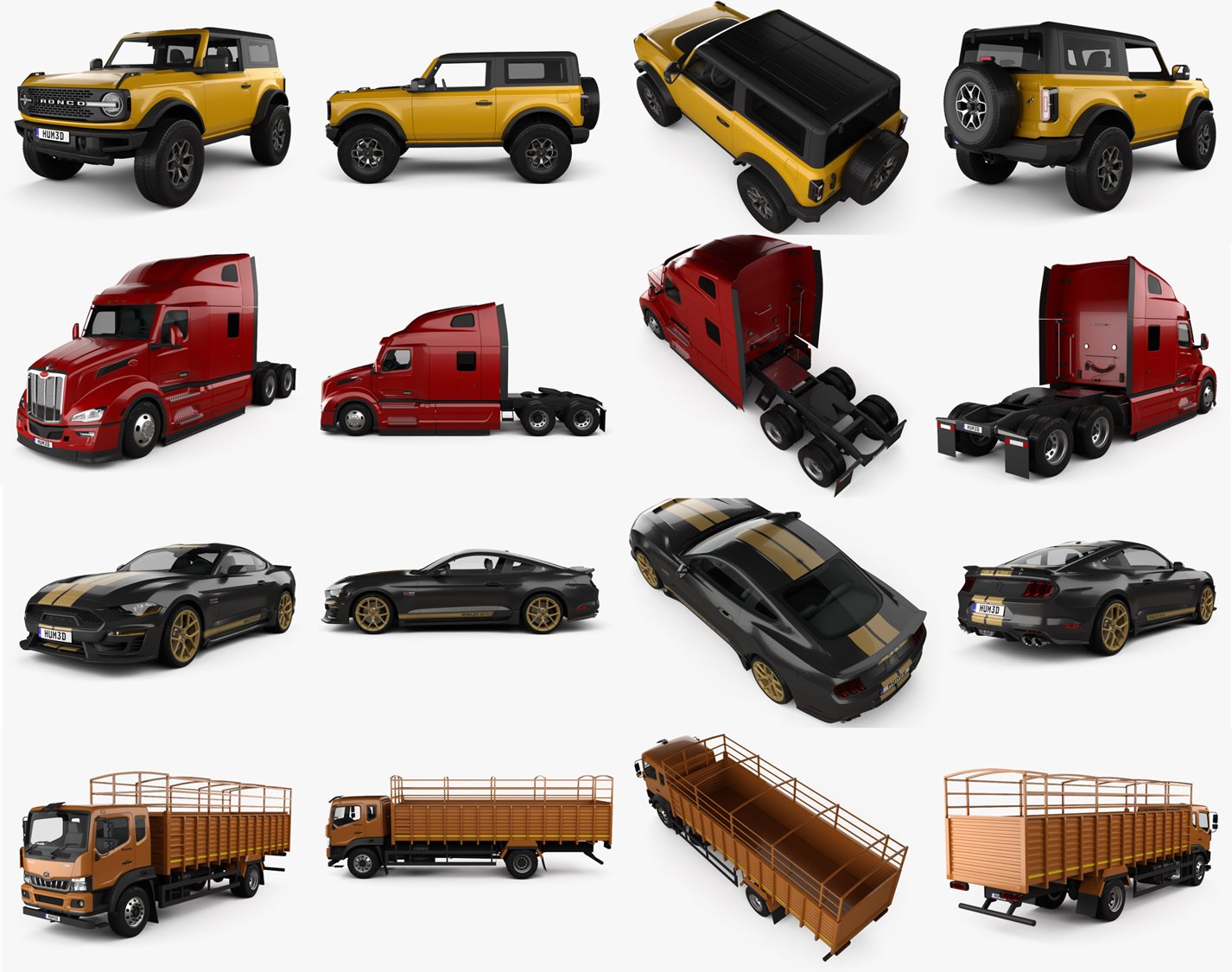}
    \caption{3D Synthetic Images}
    \label{fig:fig17}
\end{figure}
The model has been trained with 3D synthetic images added to the initial sample images with the following parameters:\\
input size = 512\\
batch size = 6\\
model weights = last.pt, from previous training session \\
validation spit ratio = 0.22 \\
Let's evaluate the model starting with the metrics: \\

\begin{figure}[H]
    \centering
    \includegraphics[width=\linewidth]{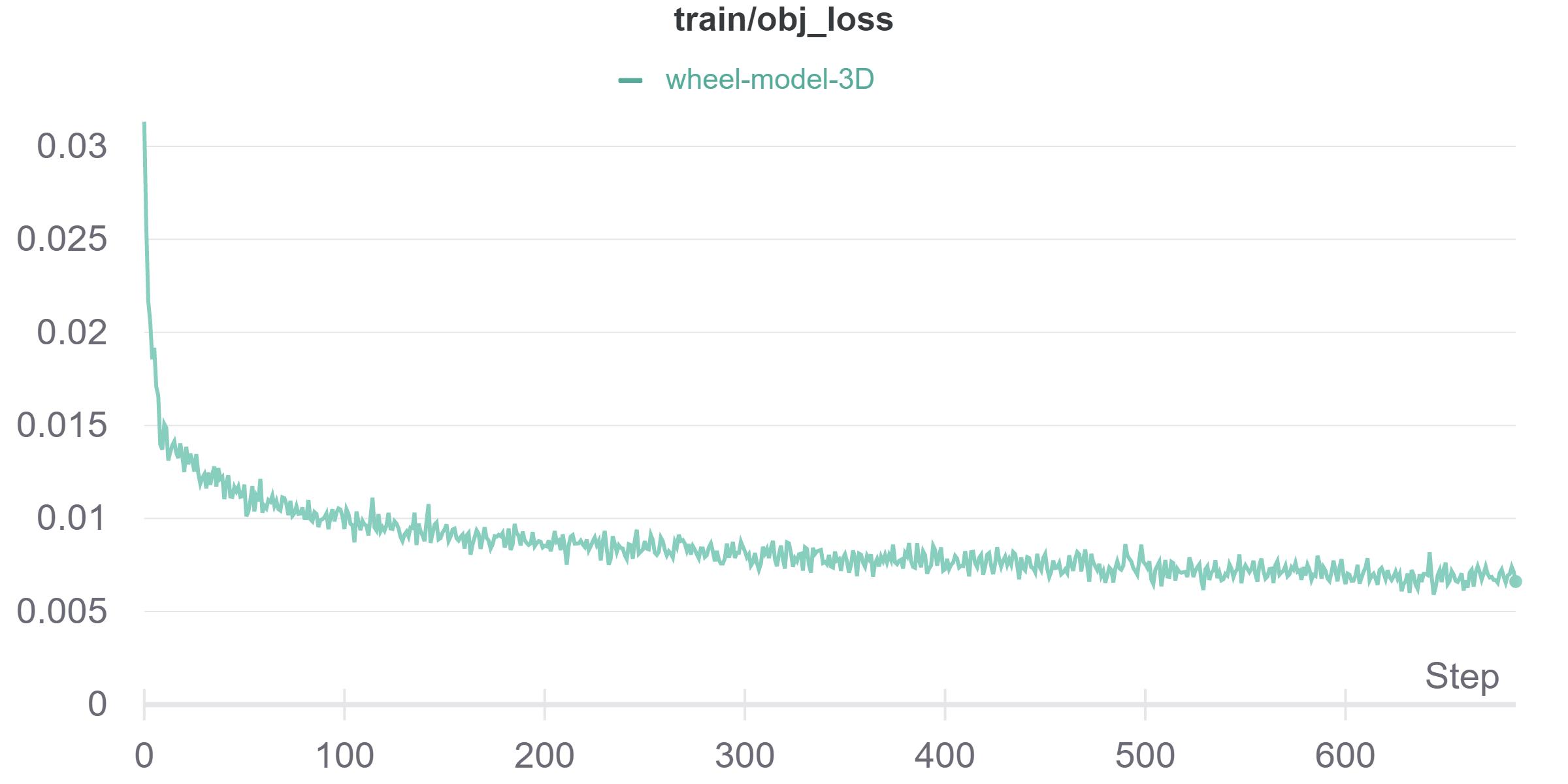}
    \caption{(3D Synthetic Images) Training Object Loss}
    \label{fig:fig18}
\end{figure}
\begin{figure}[H]
    \centering
    \includegraphics[width=\linewidth]{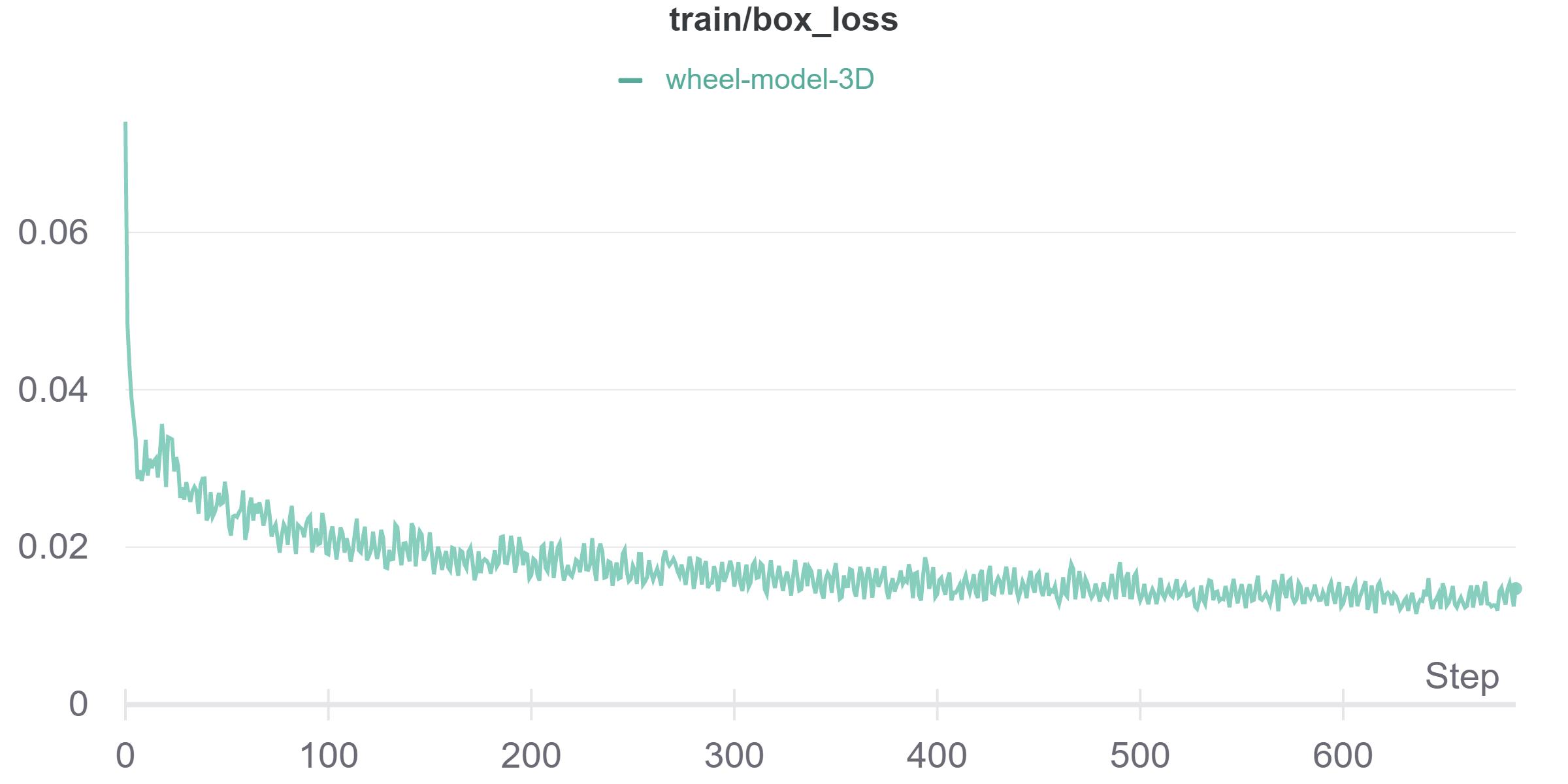}
    \caption{(3D Synthetic Images) Training Box Loss}
    \label{fig:fig18.2}
\end{figure}
\begin{figure}[H]
    \centering
    \includegraphics[width=\linewidth]{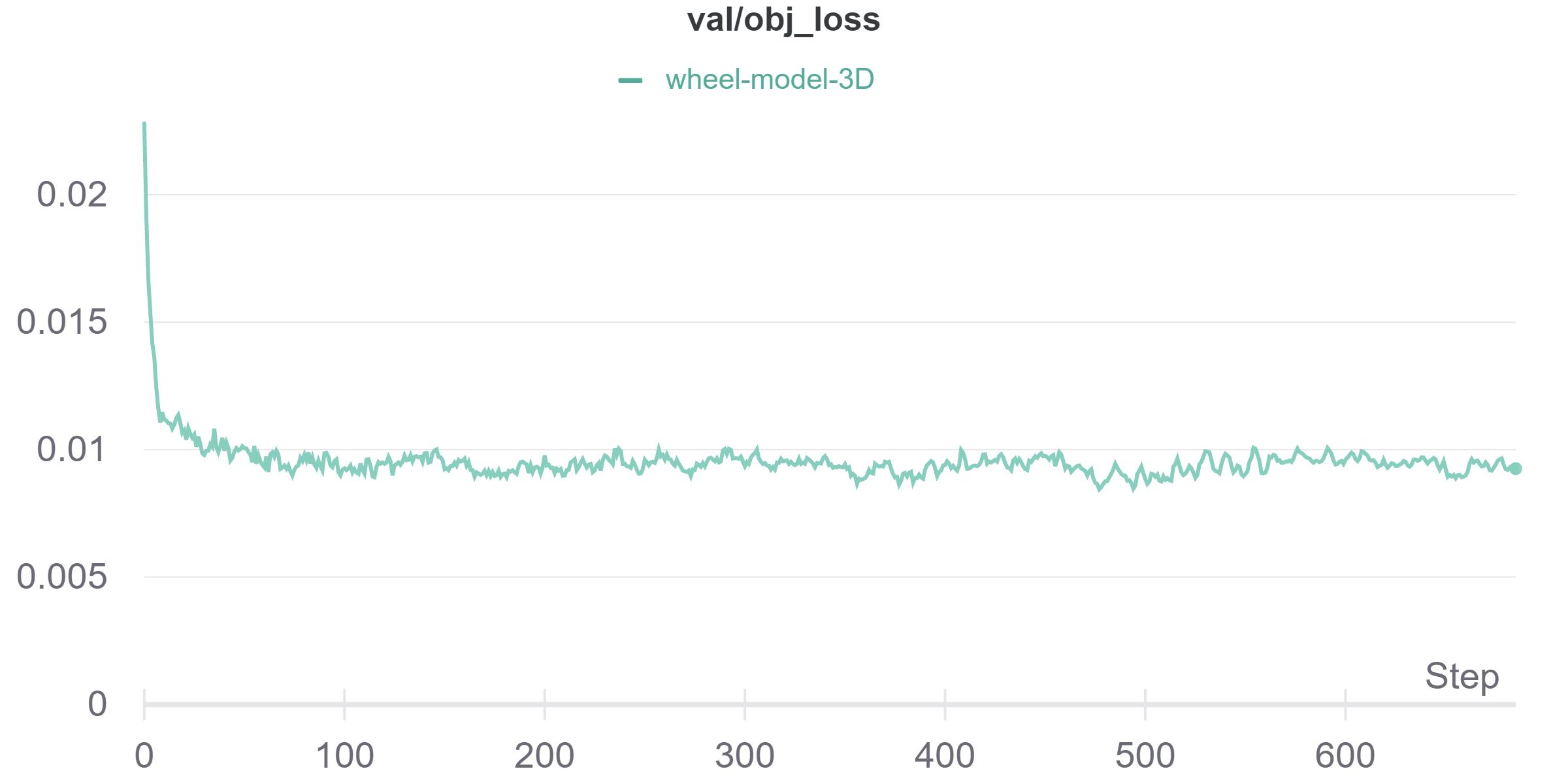}
    \caption{(3D Synthetic Images) Validation Object Loss}
    \label{fig:fig19}
\end{figure}
\begin{figure}[H]
    \centering
    \includegraphics[width=\linewidth]{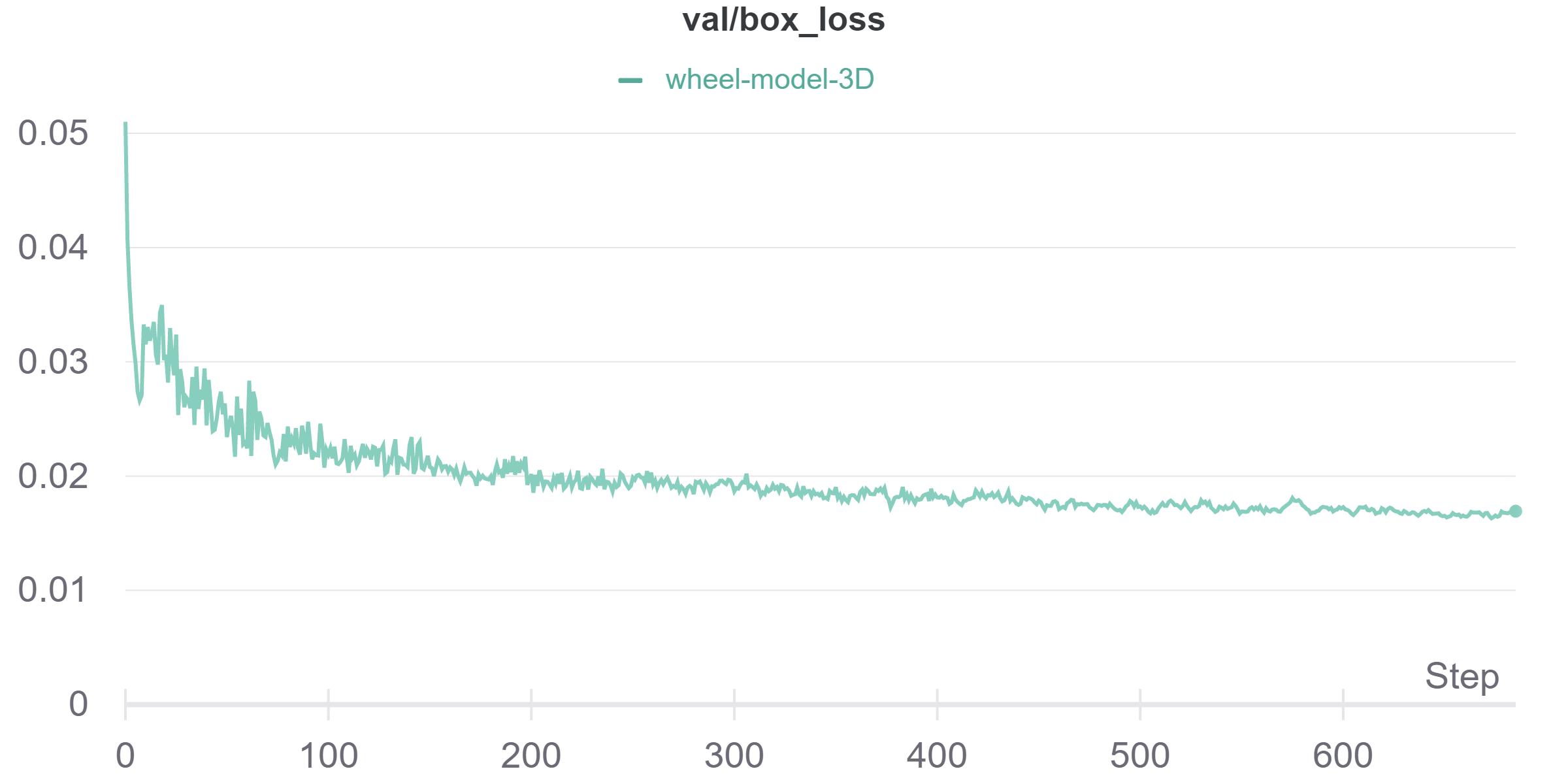}
    \caption{(3D Synthetic Images) Validation Box Loss}
    \label{fig:fig19.2}
\end{figure}
\begin{figure}[H]
    \centering
    \includegraphics[width=\linewidth]{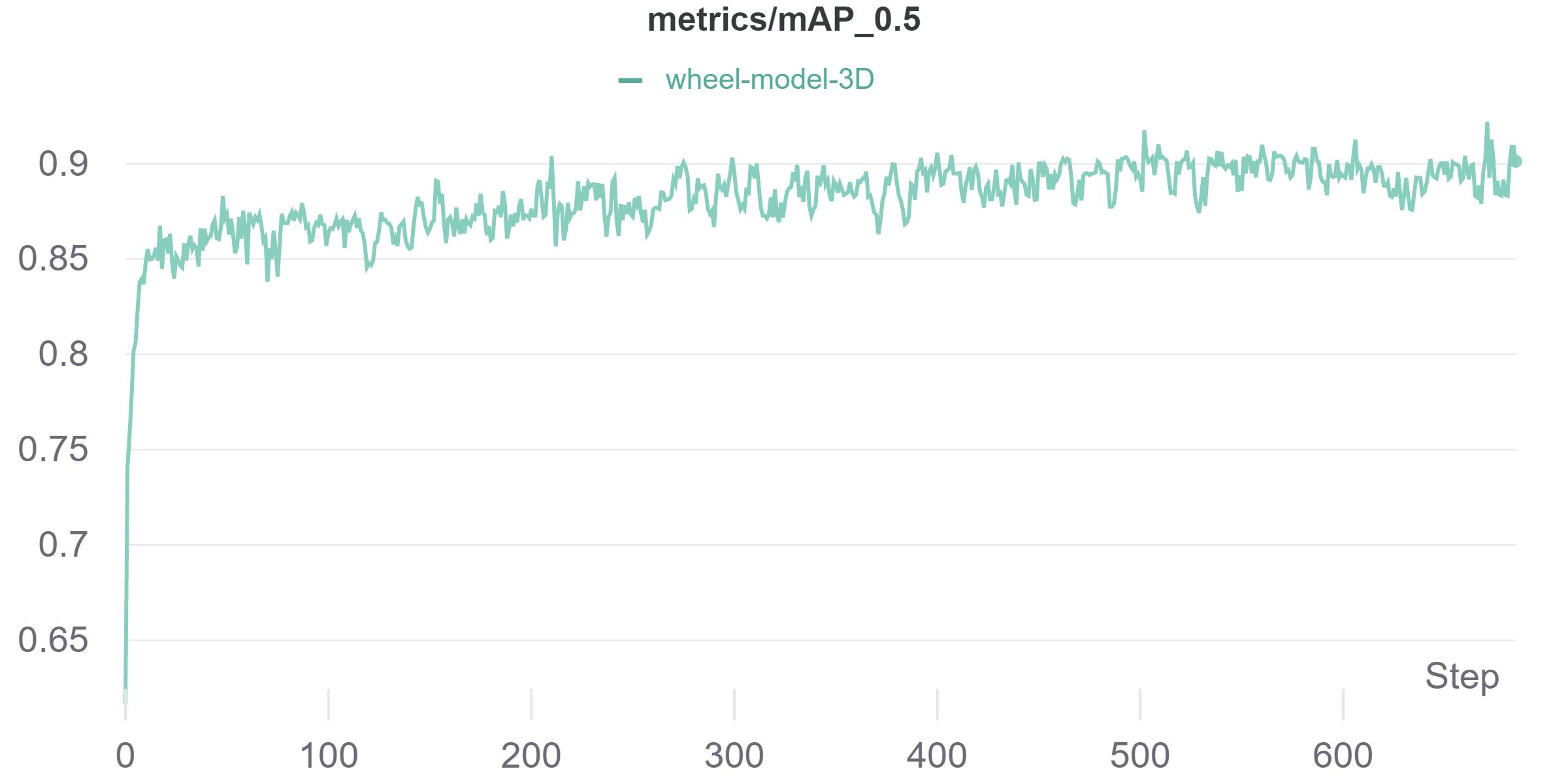}
    \caption{(3D Synthetic Images) Mean Average Precision}
    \label{fig:fig20}
\end{figure}
\begin{figure}[H]
    \centering
    \includegraphics[width=\linewidth]{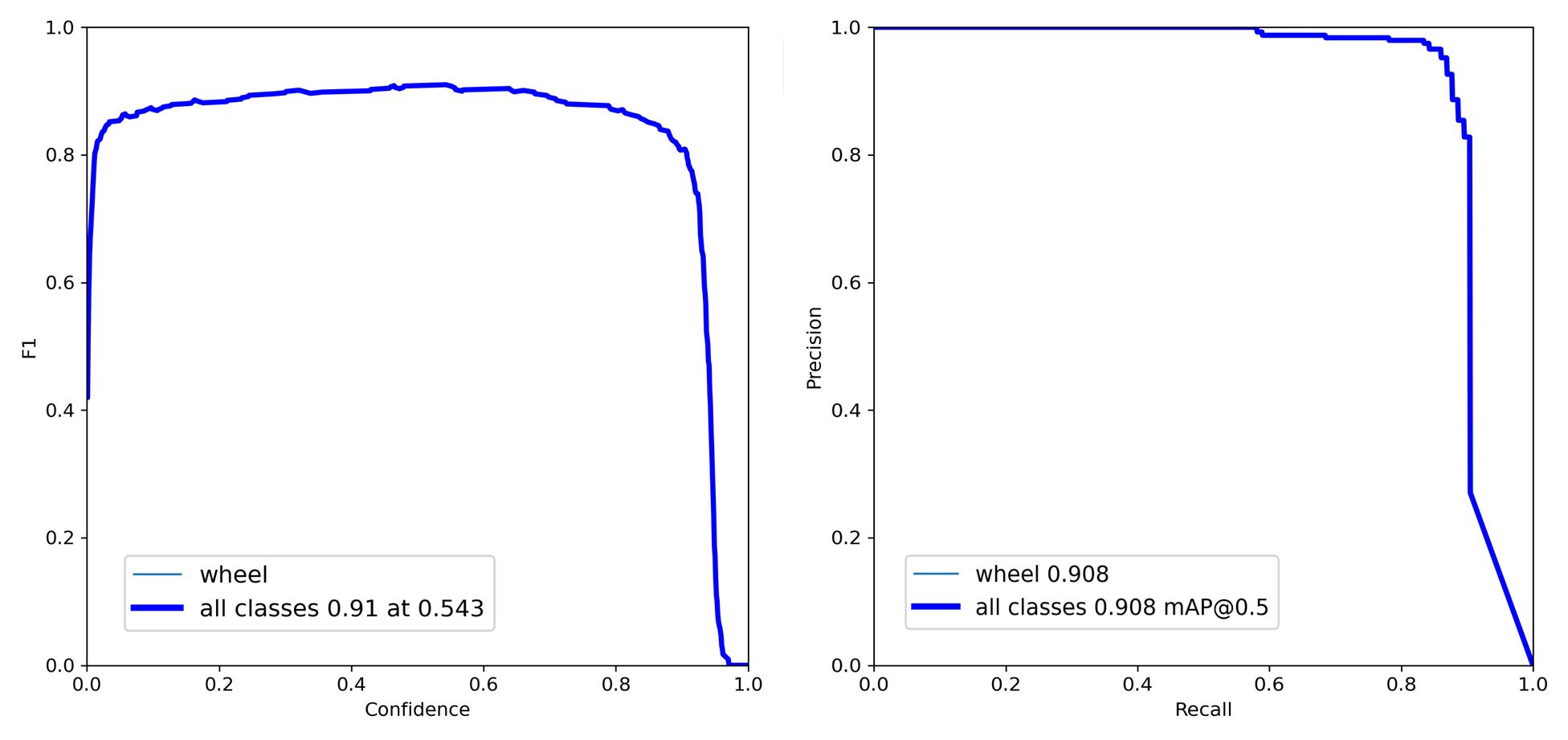}
    \caption{(3D Synthetic Images) F1 vs Confidence and Precision vs Recall}
    \label{fig:fig21}
\end{figure}
The metrics looks better. The training and validation loss seems consistent, and the mean average precision is getting up to the 0.9 ranges\\
Let's check the detection performance on random samples from validation set.
\begin{figure}[H]
    \centering
    \includegraphics[width=\linewidth]{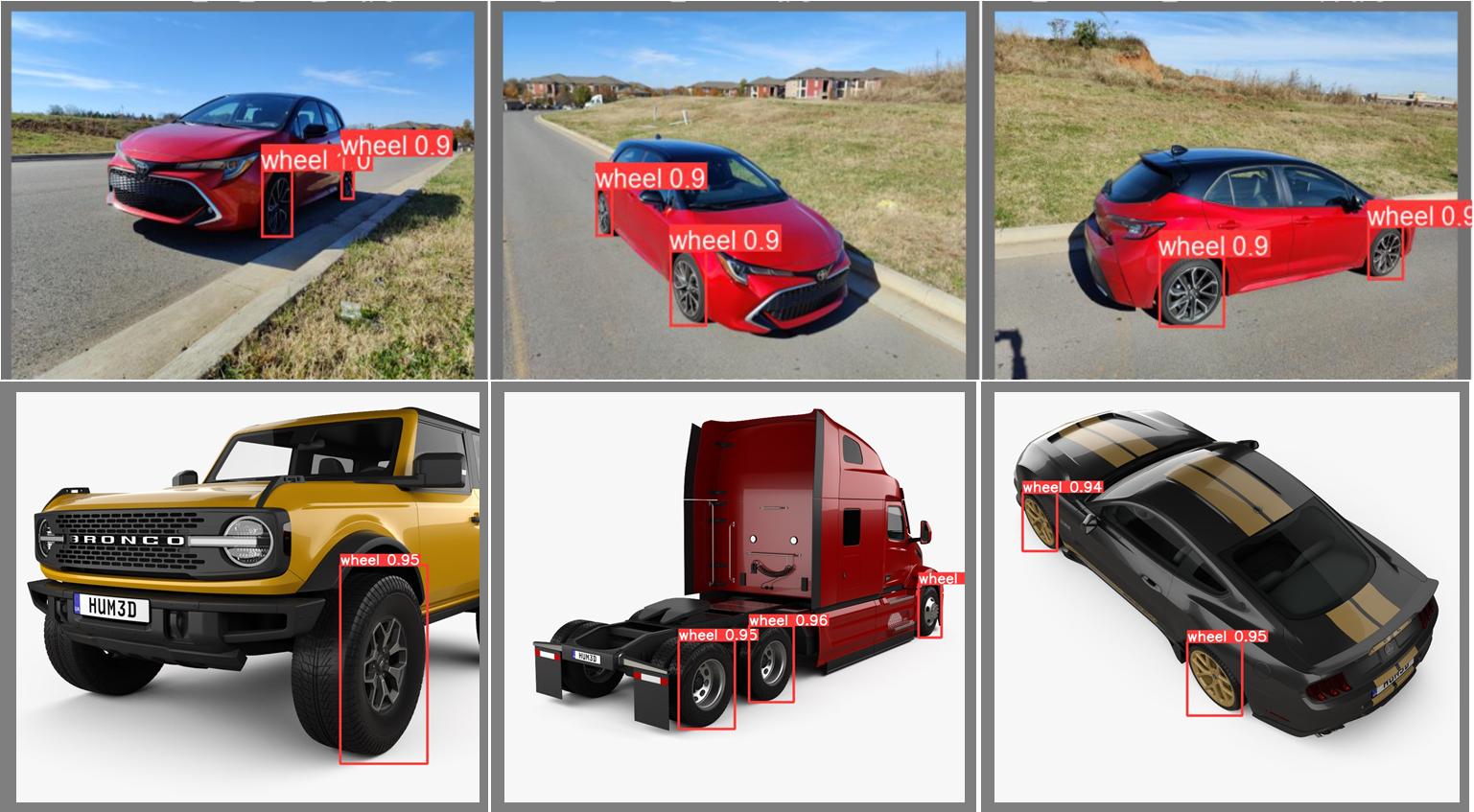}
    \caption{(3D Synthetic Images) Detection Validation Set}
    \label{fig:fig22}
\end{figure}
The model has improved and the validation looks good.
Now, let's test on some of the next image samples that are collected from CompCars \cite{b9} public dataset.

\begin{figure}[H]
    \centering
    \includegraphics[width=\linewidth]{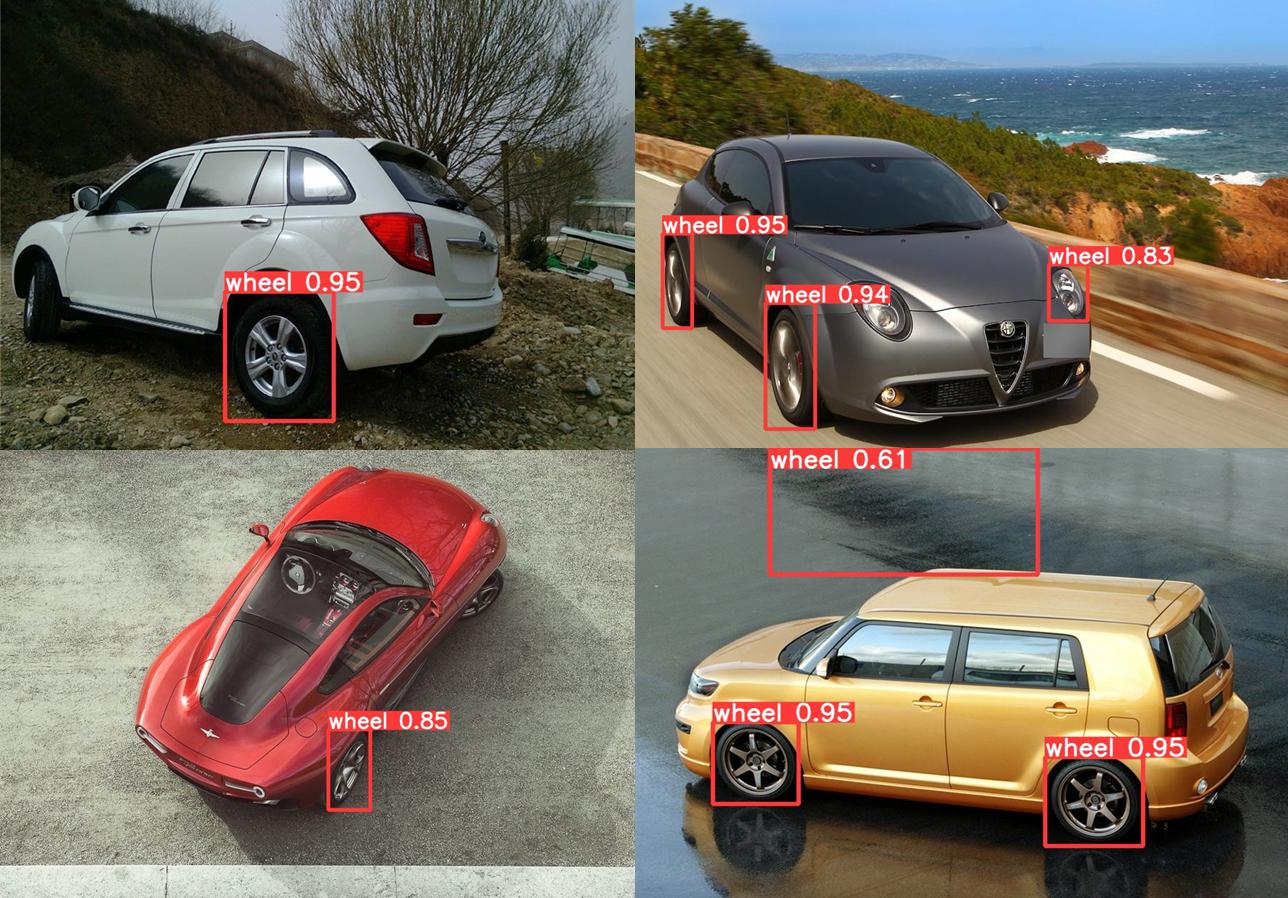}
    \caption{(3D Synthetic Images) Detection on CompCars Images}
    \label{fig:fig23}
\end{figure}
Overall, the detection is looking good in terms of bounding box accuracy, but there are some miss detection and false positives. It seems like the false positives are shapes that looks like wheel and reflective dark gray ground that has same color shade as the vehicle tire! 

\subsection{Training with added Public Images and Evaluation}
The public datasets that have been selected are CompCars and OpenImages. I decided to start with CompCars, since it has a comprehensive dataset that contains 163 car makes with 1,716 car models \cite{b9} as well as combination of different roads with nature and buildings. The ability to get comprehensive set of vehicles is ideal to get variety of vehicle wheel shapes. Also, CompCars provide a close up view of the cars which makes it ideal for the model to learn more details about the different wheels introduced.\\
Total of 827 images have been manually selected from CompCars to machine label them using the previous trained model then manually verified by LabelImg. \\
Below are some of the image examples from CompCars: \\
\begin{figure}[H]
    \centering
    \includegraphics[width=\linewidth]{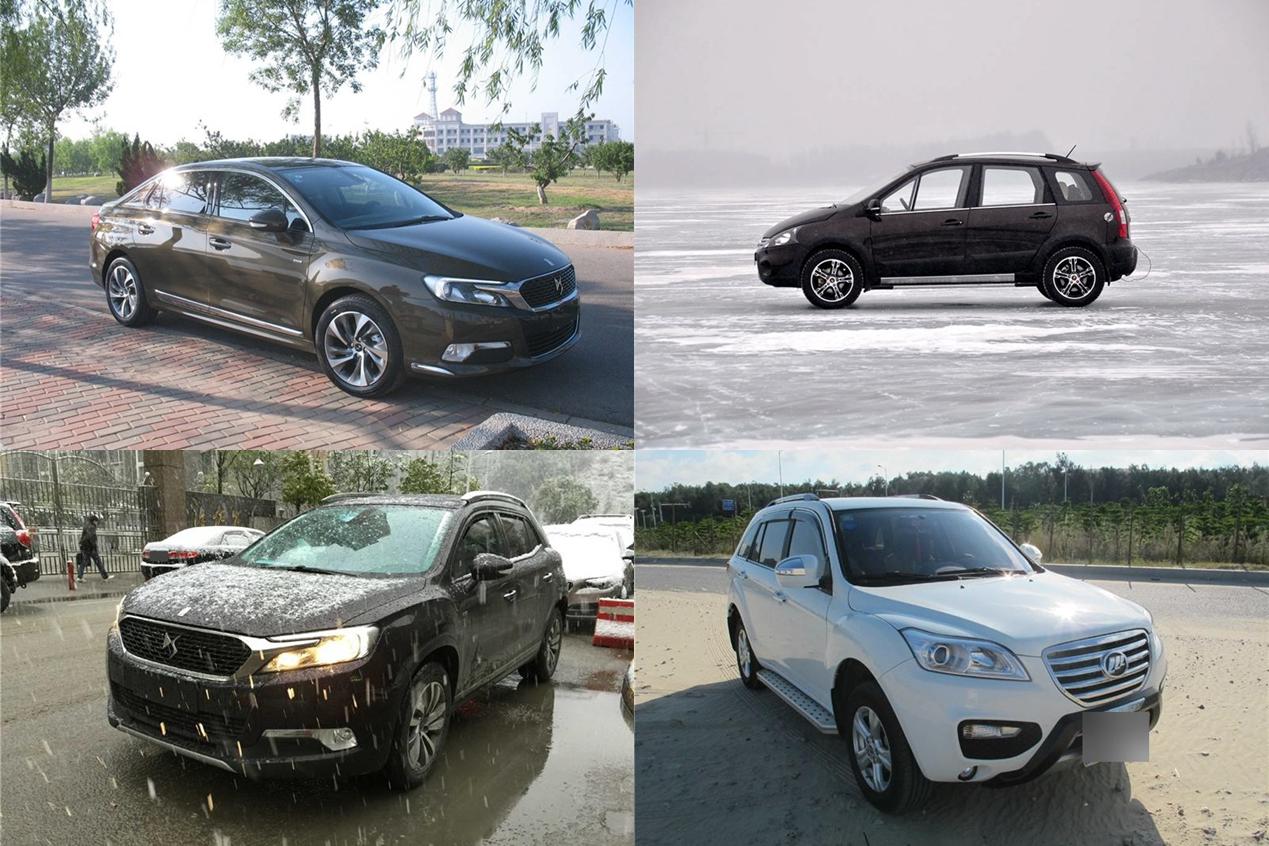}
    \caption{CompCars Image Samples}
    \label{fig:fig24}
\end{figure}
The model has been trained with CompCars images added to the previous sample images with the following parameters:\\
input size = 512\\
batch size = 6\\
model weights = last.pt, from previous training session \\
validation spit ratio = 0.22 \\\\
Let's evaluate the model starting with the metrics: \\
\begin{figure}[H]
    \centering
    \includegraphics[width=\linewidth]{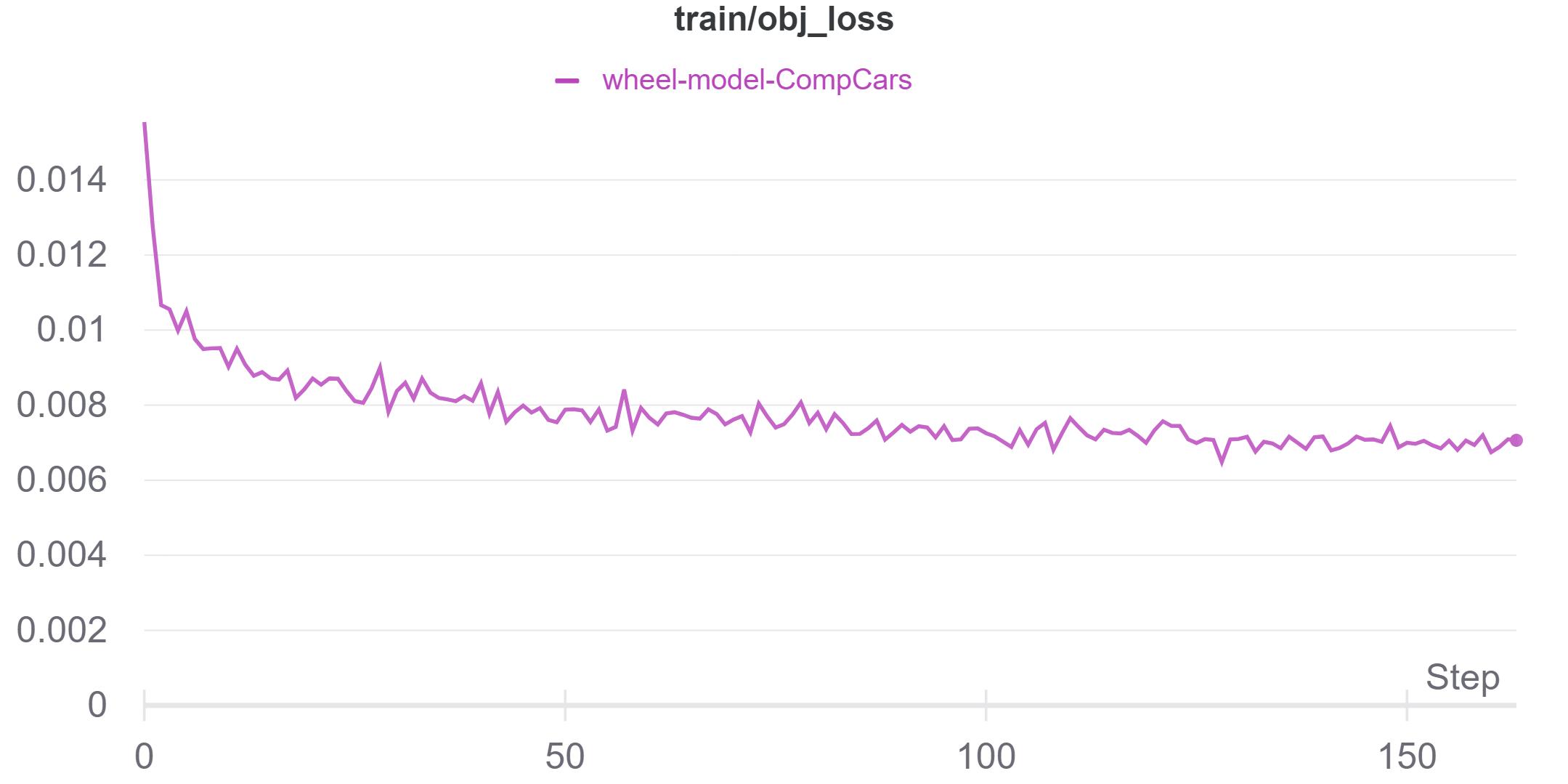}
    \caption{CompCars Model Train Object Loss}
    \label{fig:fig25}
\end{figure}
\begin{figure}[H]
    \centering
    \includegraphics[width=\linewidth]{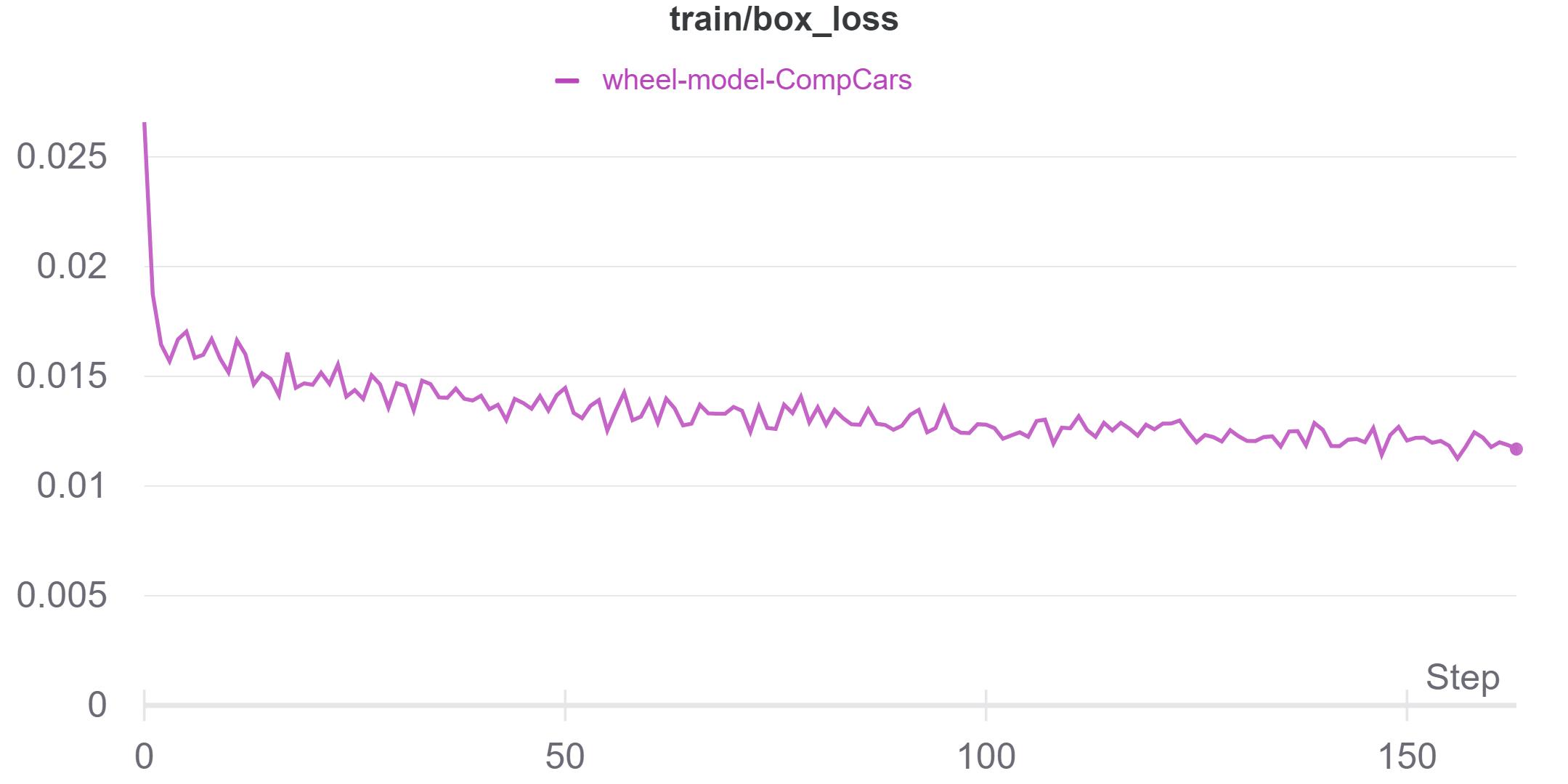}
    \caption{CompCars Model Train Box Loss}
    \label{fig:fig26}
\end{figure}
\begin{figure}[H]
    \centering
    \includegraphics[width=\linewidth]{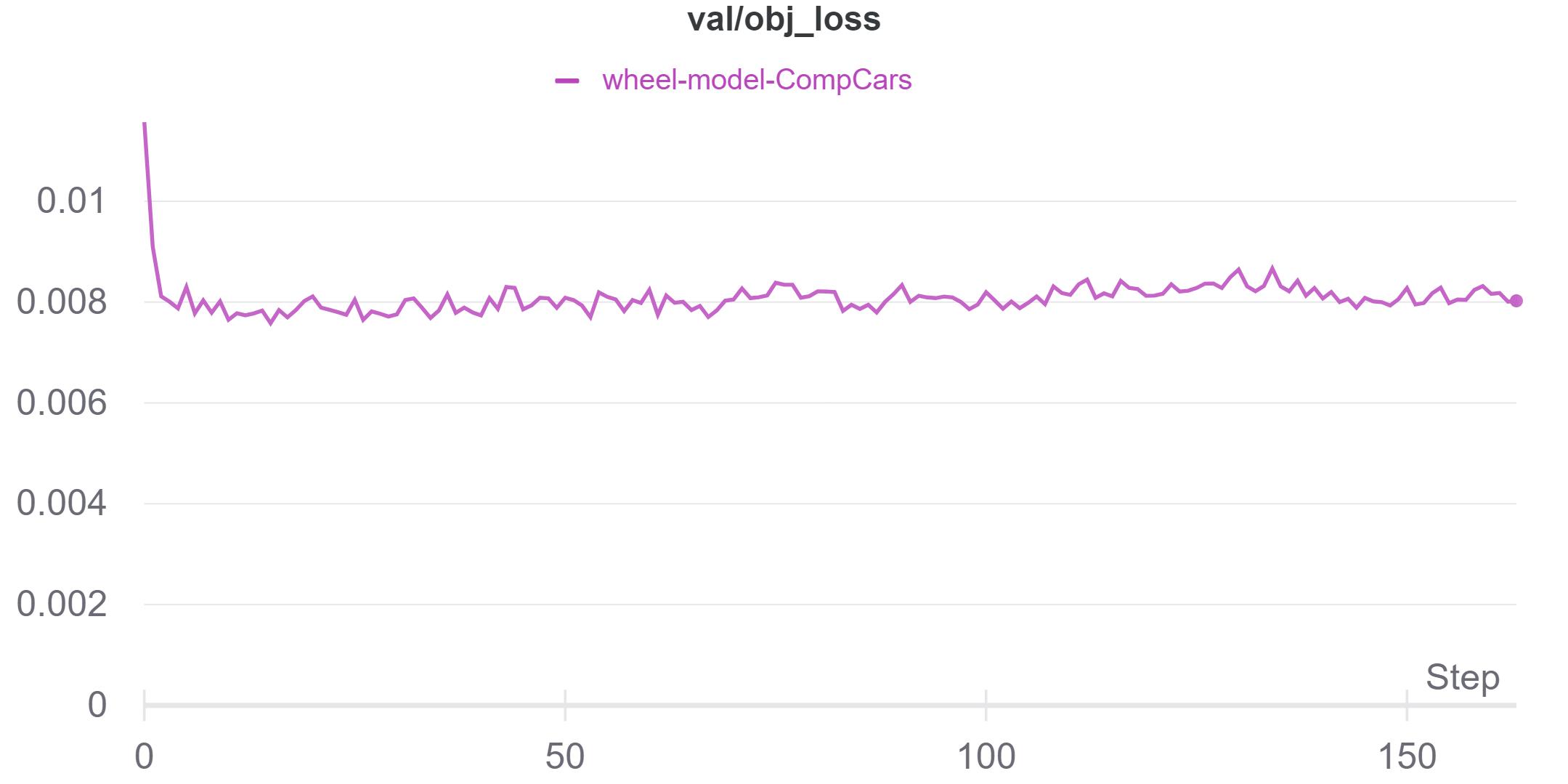}
    \caption{CompCars Model Validation Object Loss}
    \label{fig:fig27}
\end{figure}
\begin{figure}[H]
    \centering
    \includegraphics[width=\linewidth]{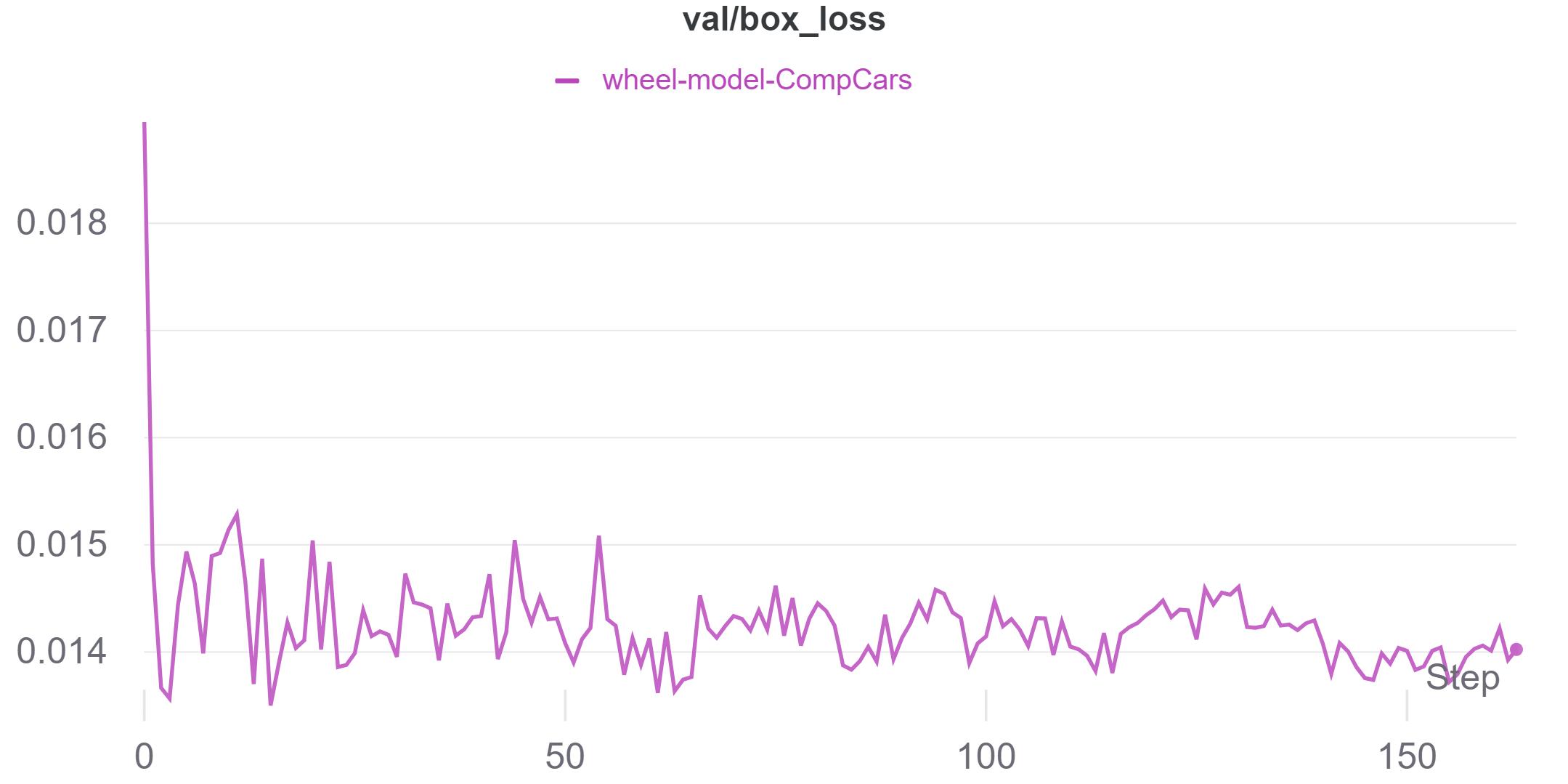}
    \caption{CompCars Model Validation Box Loss}
    \label{fig:fig28}
\end{figure}
\begin{figure}[H]
    \centering
    \includegraphics[width=\linewidth]{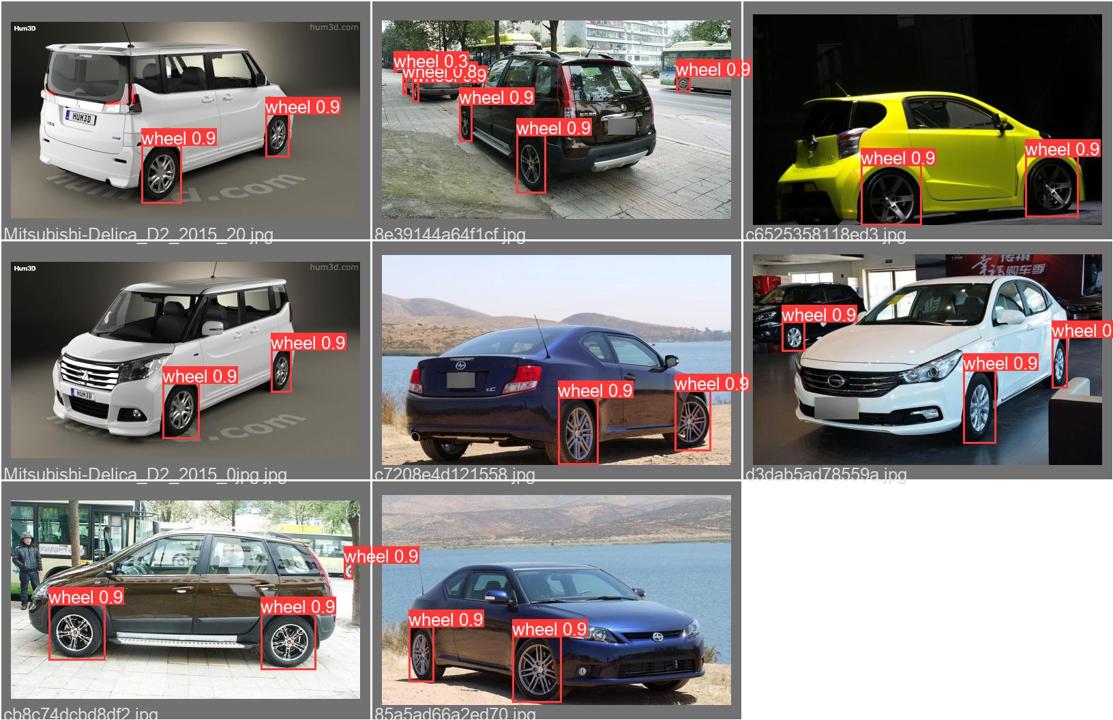}
    \caption{CompCars Detection on Random Validation Set}
    \label{fig:fig29}
\end{figure}
The model has performed well on the validation set. \\
Then tested the model with CompCars on some of the samples from OpenImages to check the detection performance.
\begin{figure}[H]
    \centering
    \includegraphics[width=\linewidth]{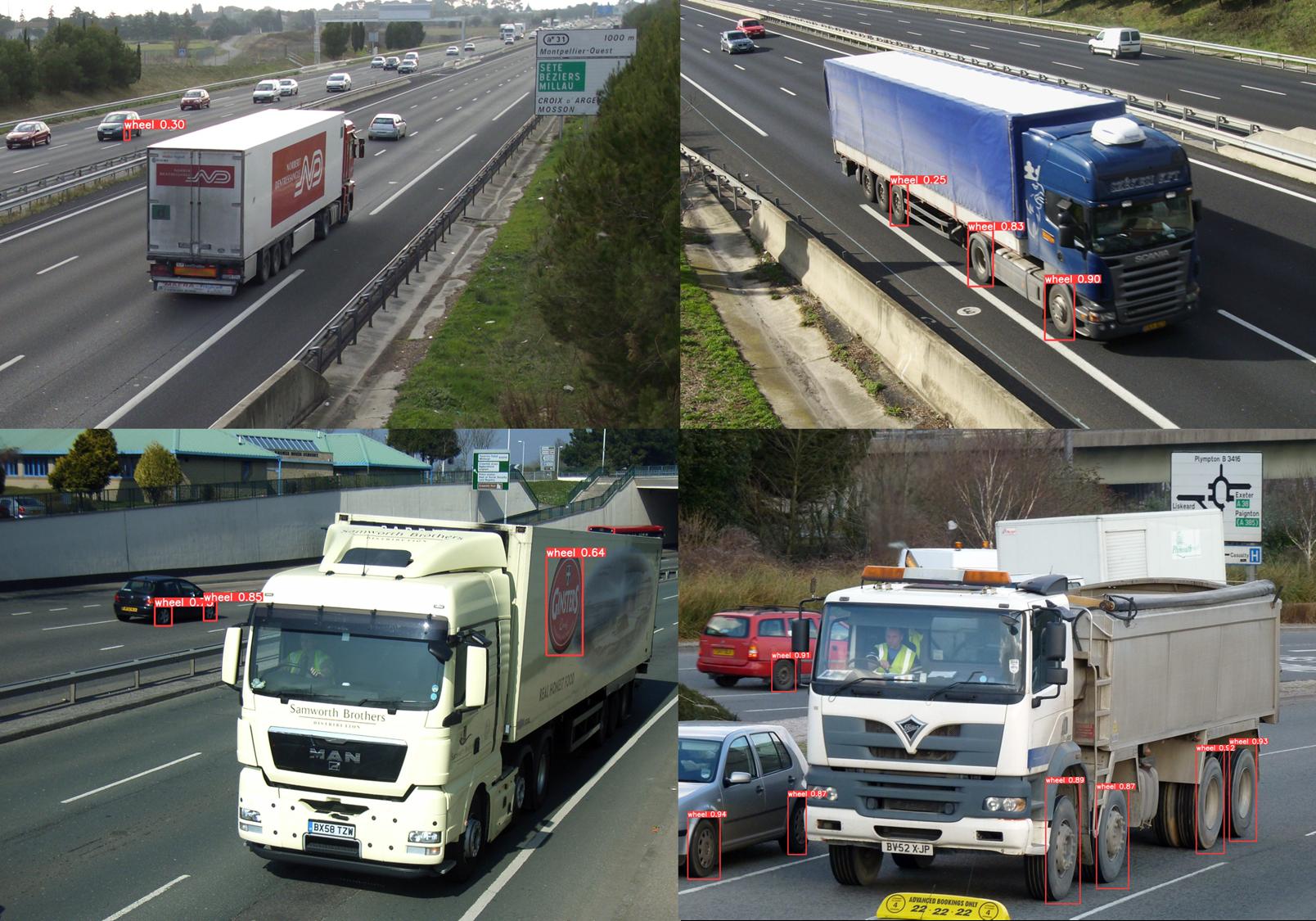}
    \caption{CompCars Model Testing on OpenImages Samples}
    \label{fig:fig30}
\end{figure}
It seems that the detection results are inconsistent. I decided to choose OpenImages as the next image set for training, since it contains some real examples that are visually close to the final use case of the detector. The total number of samples from OpenImages are 179 images that have been machine labeled using the previous trained model. Manual review has been done on the automatically labeled images. It seems that the model gets better every time with less manual edits needed to be made.
\begin{figure}[H]
    \centering
     \includegraphics[width=\linewidth]{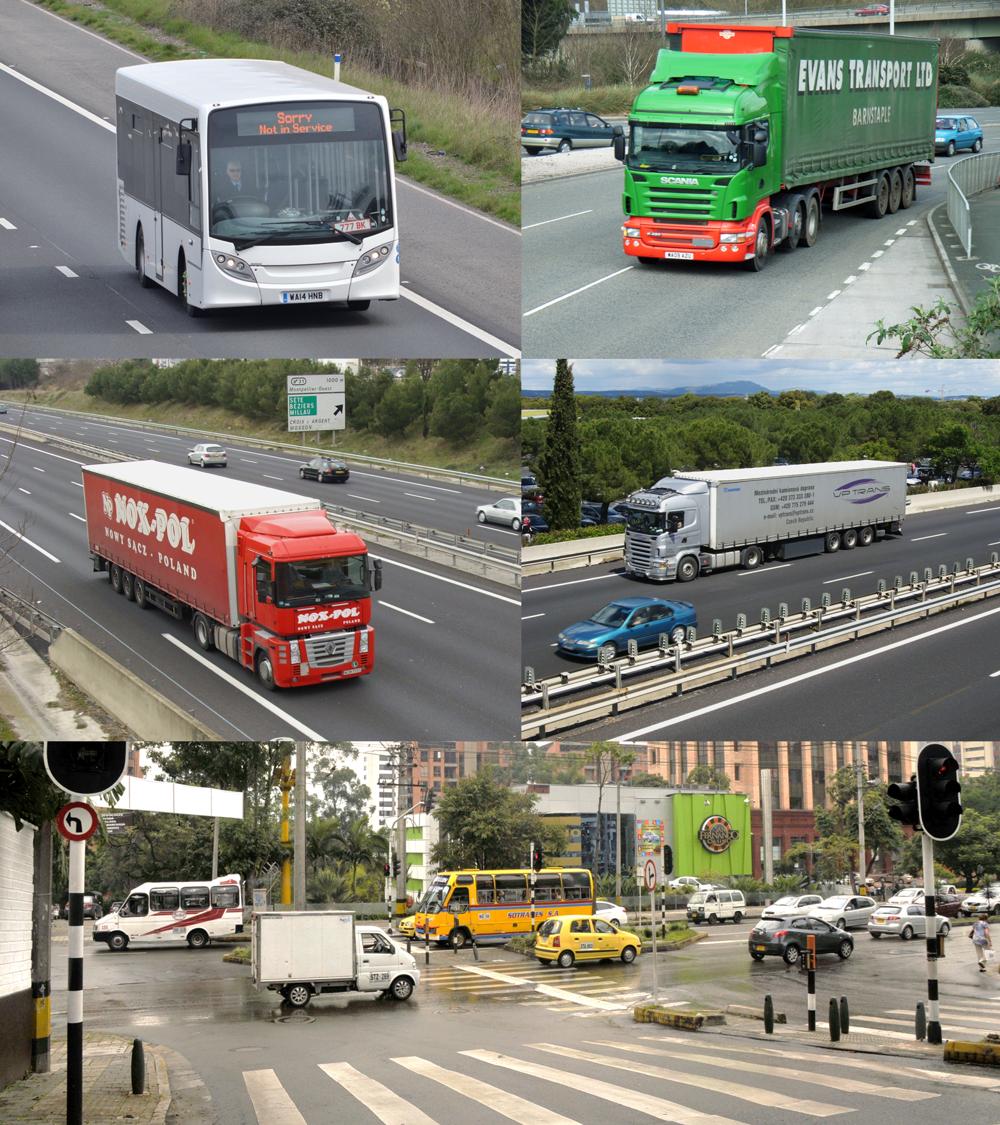}
    \caption{OpenImages Samples}
    \label{fig:fig31}
\end{figure}
The model has been trained with the samples from OpenImages added to the previous sample images with the following parameters:\\
input size = 512\\
batch size = 6\\
model weights = last.pt, from previous training session \\
validation spit ratio = 0.22 \\\\

Let's evaluate the model starting with the metrics: \\
This time will be comparing the metrics from OpenImages and CompCars in same plots so we can get better insight since they are getting very close to each other. 
\begin{figure}[H]
    \centering
     \includegraphics[width=\linewidth]{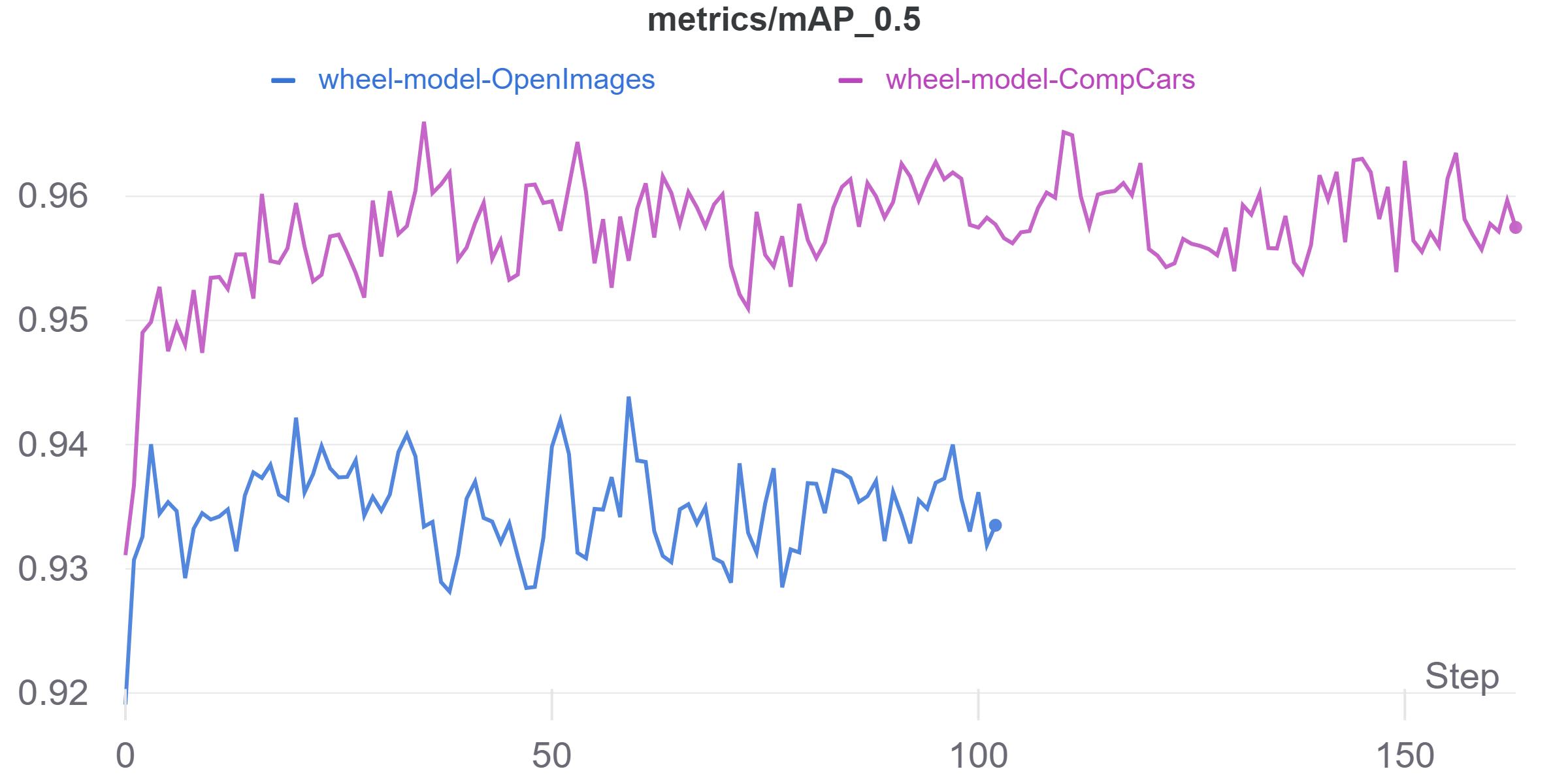}
    \caption{OpenImages and CompCars mAP}
    \label{fig:fig32}
\end{figure}
\begin{figure}[H]
    \centering
     \includegraphics[width=\linewidth]{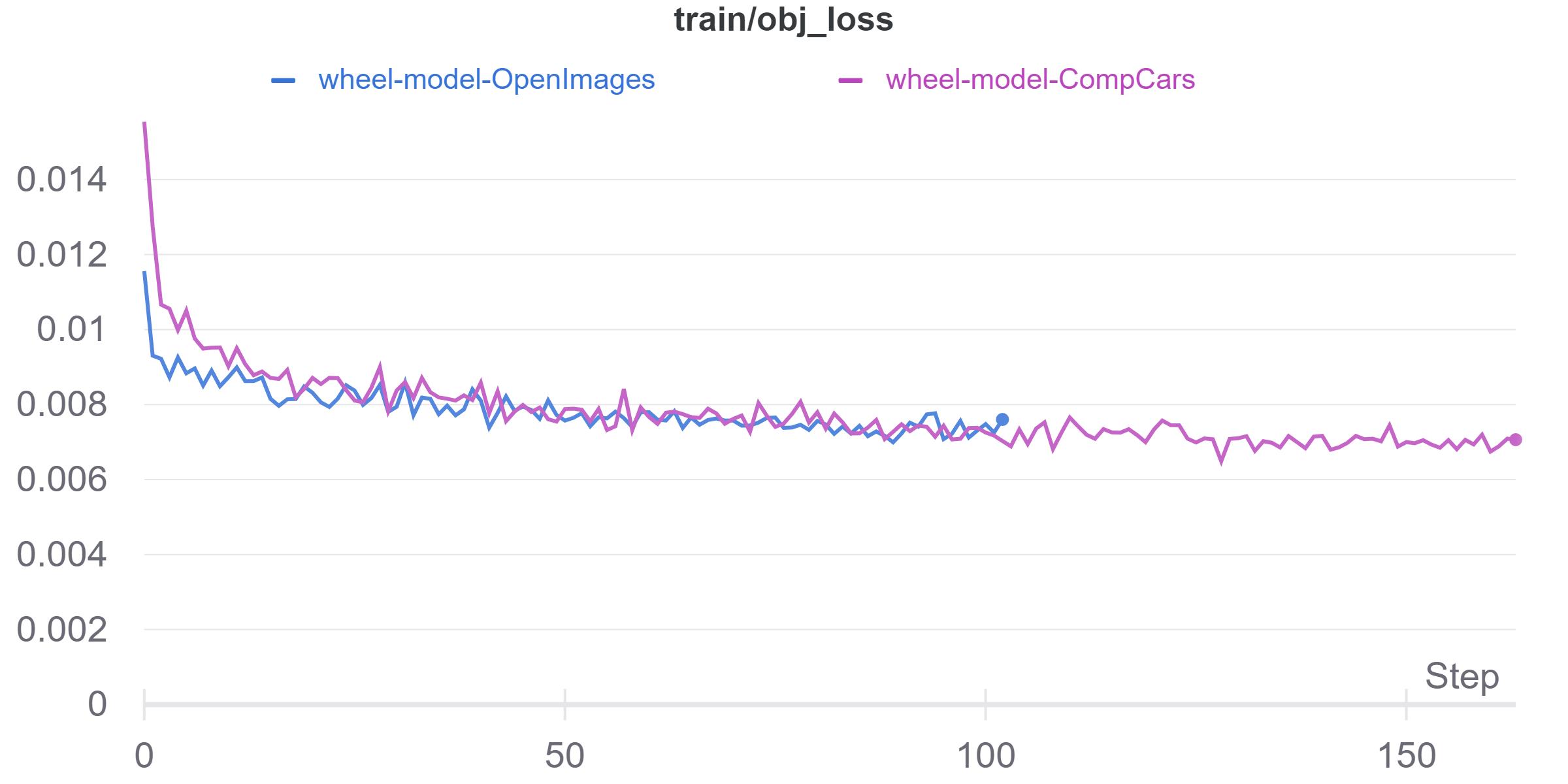}
    \caption{OpenImages and CompCars Train Object Loss}
    \label{fig:fig33}
\end{figure}
\begin{figure}[H]
    \centering
     \includegraphics[width=\linewidth]{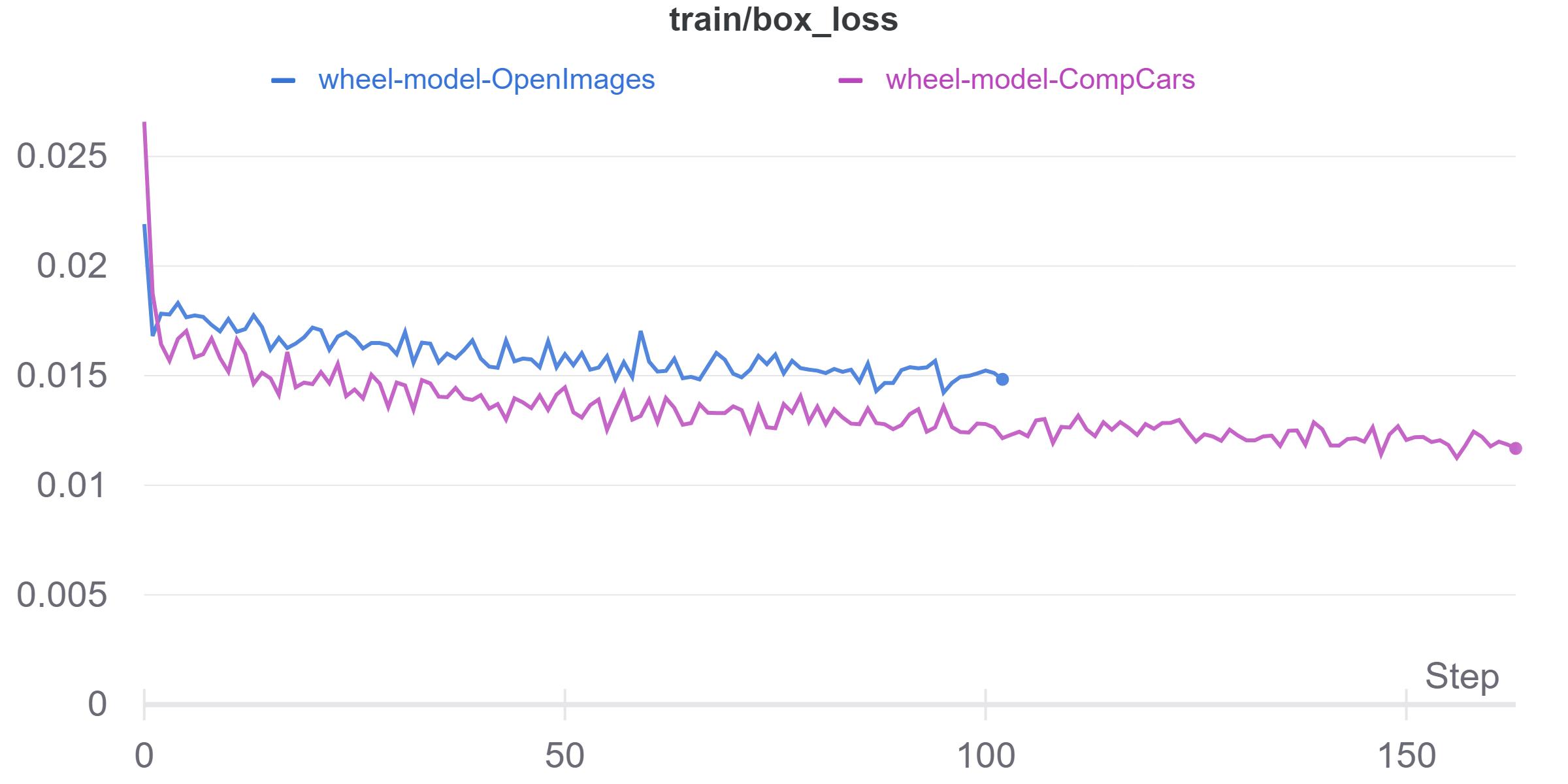}
    \caption{OpenImages and CompCars Train Box Loss}
    \label{fig:fig34}
\end{figure}
\begin{figure}[H]
    \centering
     \includegraphics[width=\linewidth]{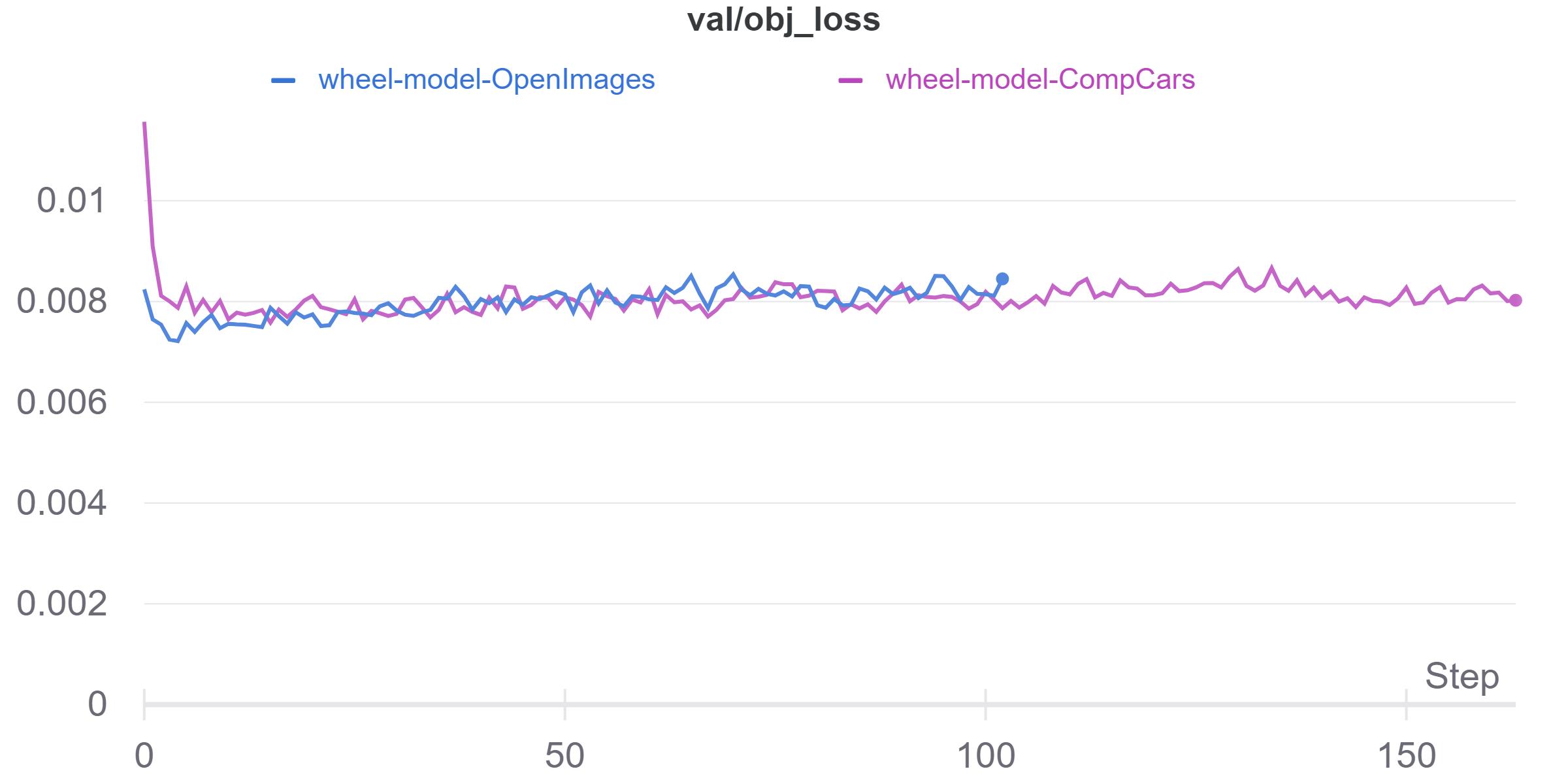}
    \caption{OpenImages and CompCars Validation Object Loss}
    \label{fig:fig35}
\end{figure}
\begin{figure}[H]
    \centering
     \includegraphics[width=\linewidth]{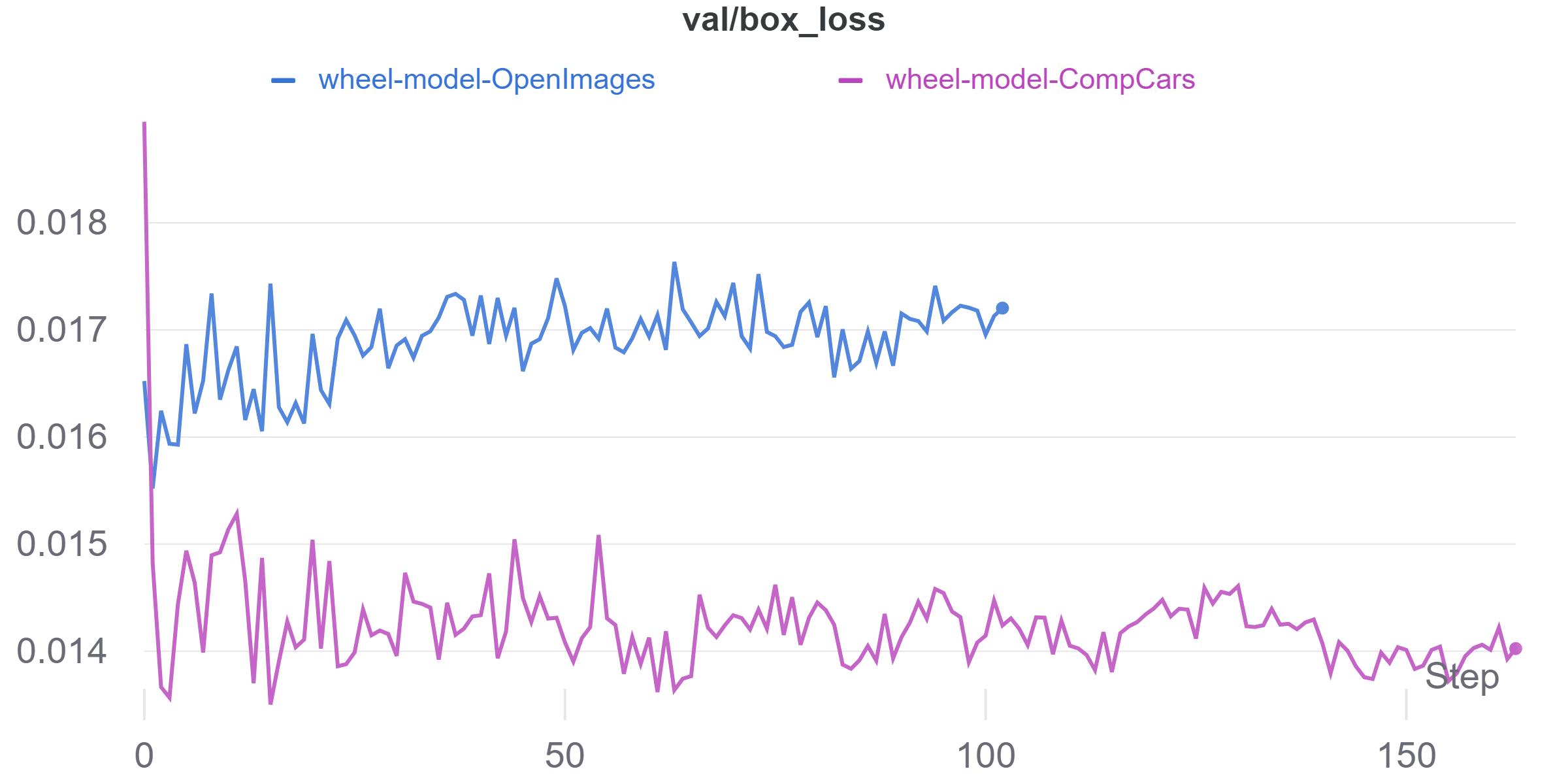}
    \caption{OpenImages and CompCars Validation Box Loss}
    \label{fig:fig36}
\end{figure}
Looking at those plots and trying to understand what just happened, it was an aha moment.
The metrics all of the sudden started to look bad. Mean average precision has dropped.
The interesting insight here is the bounding box loss. We can clearly see OpenImages has gotten worse than CompCars. At the same time, the actual object loss has remained very similar to each other. That's where I realized that my samples from OpenImages contains very small vehicle sizes and their wheels are obviously too small after scaling down the image to 512 pixels during the training. Those small bounding boxes can encounter rounding issues at that small scale of 512 input size. This especially magnified with rectangular image sizes that are far from the aspect ratio of a square. That's when I came to realization that I need to crop the OpenImages to vehicle size region of interest in order to maximize the wheel size in the image. I utilized the existing YOLOv5 model to do the detection and save the cropped images based on the bounding boxes of the detection with classification of bus, car, truck, and others. Every other type of object has been used as negative samples with empty label files.

Before I proceeded with my theory, I decided to test the detector on completely different image found to mimic the final ideal use case for the detector. 

\begin{figure}[H]
    \centering
    \includegraphics[width=\linewidth]{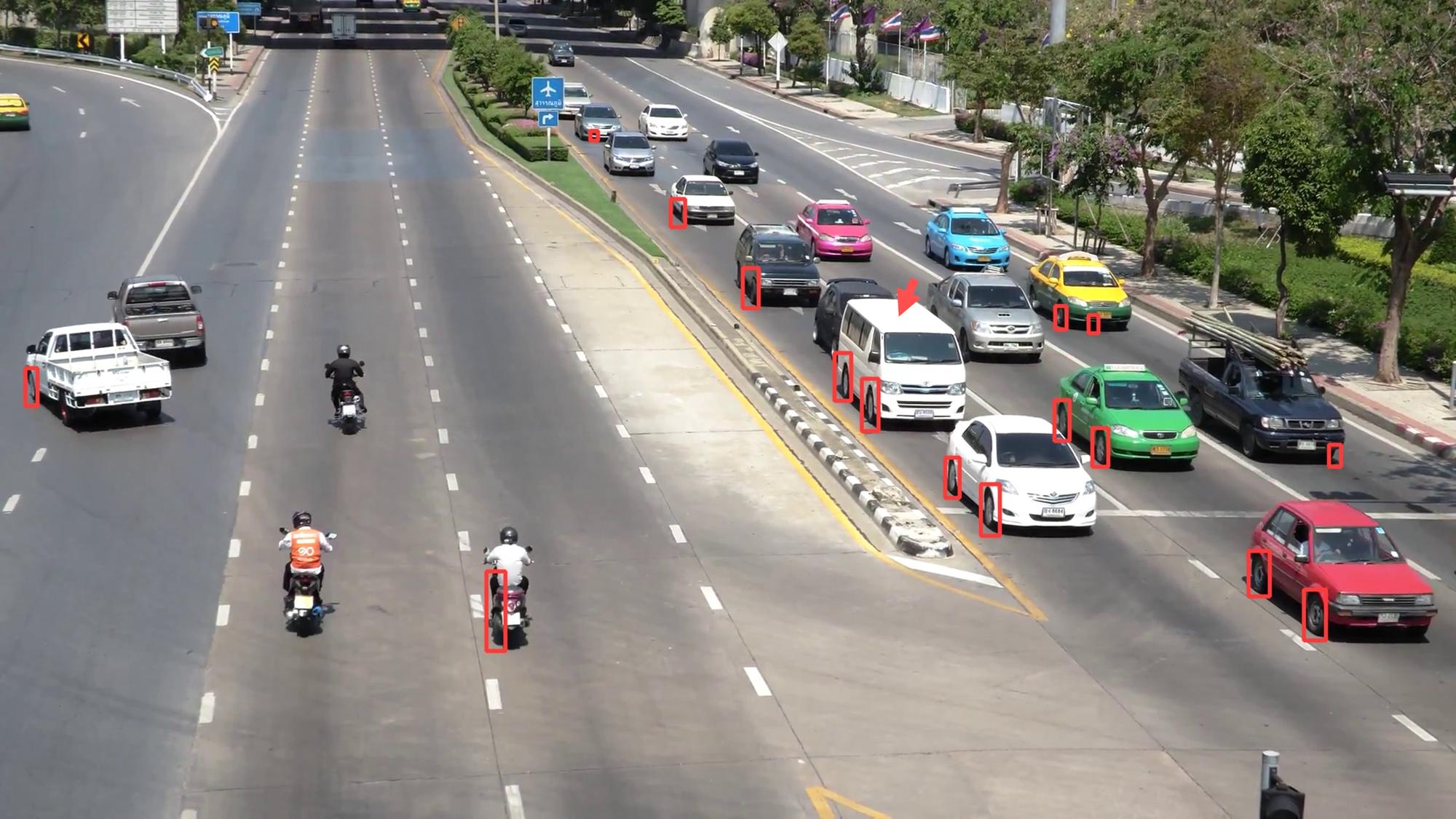}
    \caption{Model with OpenImages Model Test on Rectangular Full Image}
    \label{fig:fig37}
\end{figure}

A good example here if we look at the white van, highlighted with red arrow on the full view rectangular image, we can see how the localization has loose bounding box around the wheels. 

\begin{figure}[H]
    \centering
    \includegraphics[width=\linewidth]{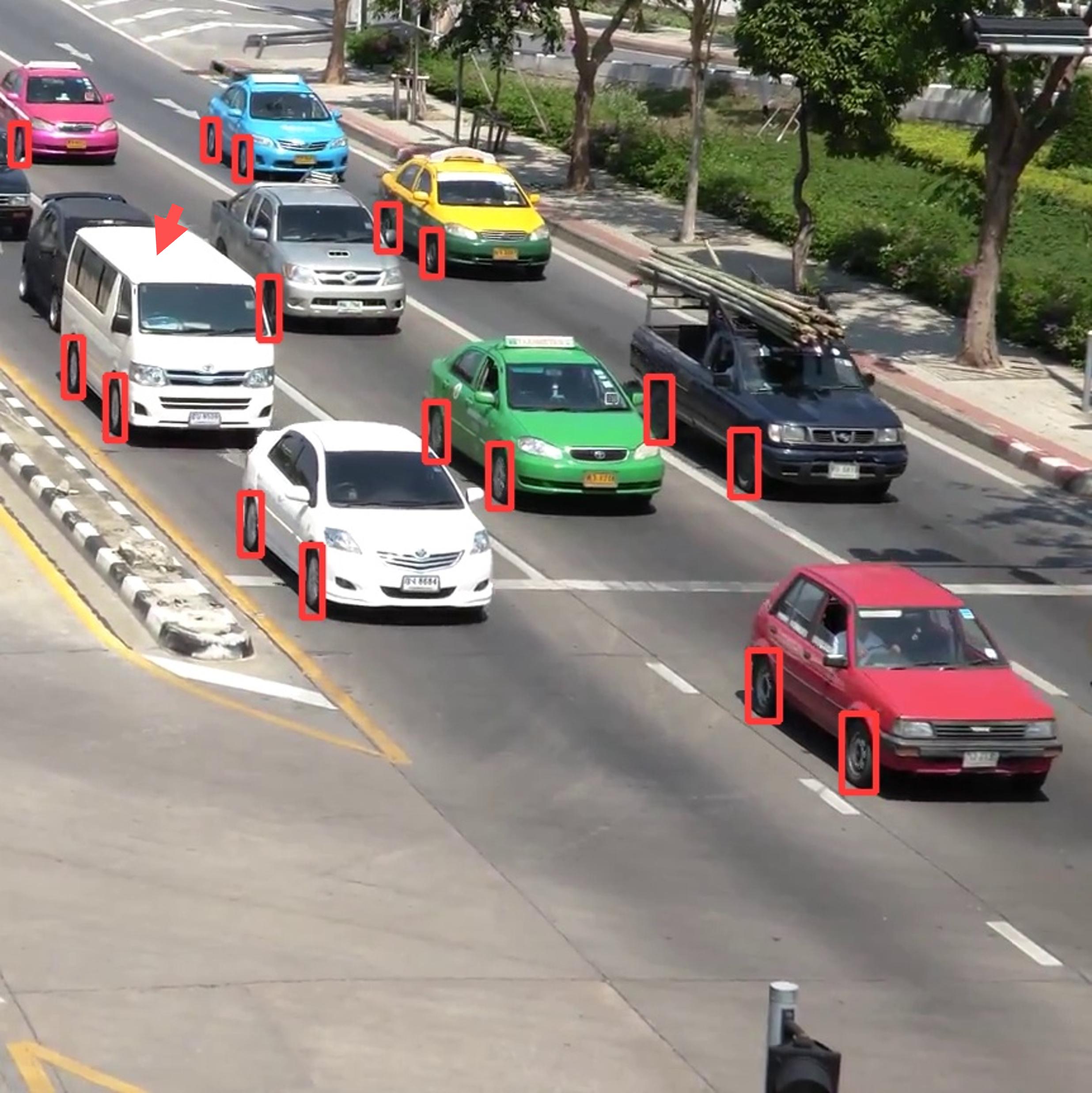}
    \caption{Model with OpenImages Test on Cropped Image}
    \label{fig:fig38}
\end{figure}

After comparing the same vehicle on the cropped view, we can clearly see that the localization of the wheels have tight bounding boxes. This clearly explains why the object loss plot looked better than the box loss plot. \\

The next approach is to remove the full resolution sample images then train the model with cropped objects views extracted from the full images. Total cropped images are 1157 that got machine labeled and manually reviewed. 

Finally, the model has been trained with the samples from OpenImages cropped added to the previous sample images, with the full view removed. The following parameters used for the training:\\
input size = 512\\
batch size = 6\\
model weights = last.pt, from previous training session \\
validation spit ratio = 0.22 \\\\

Let's evaluate the model and check the metrics: \\
Again, we will be comparing the metrics from OpenImages, CompCars, and OpenImage-crop in same plots so we can get better insight since they are getting really close to each other. 

\begin{figure}[H]
    \centering
    \includegraphics[width=\linewidth]{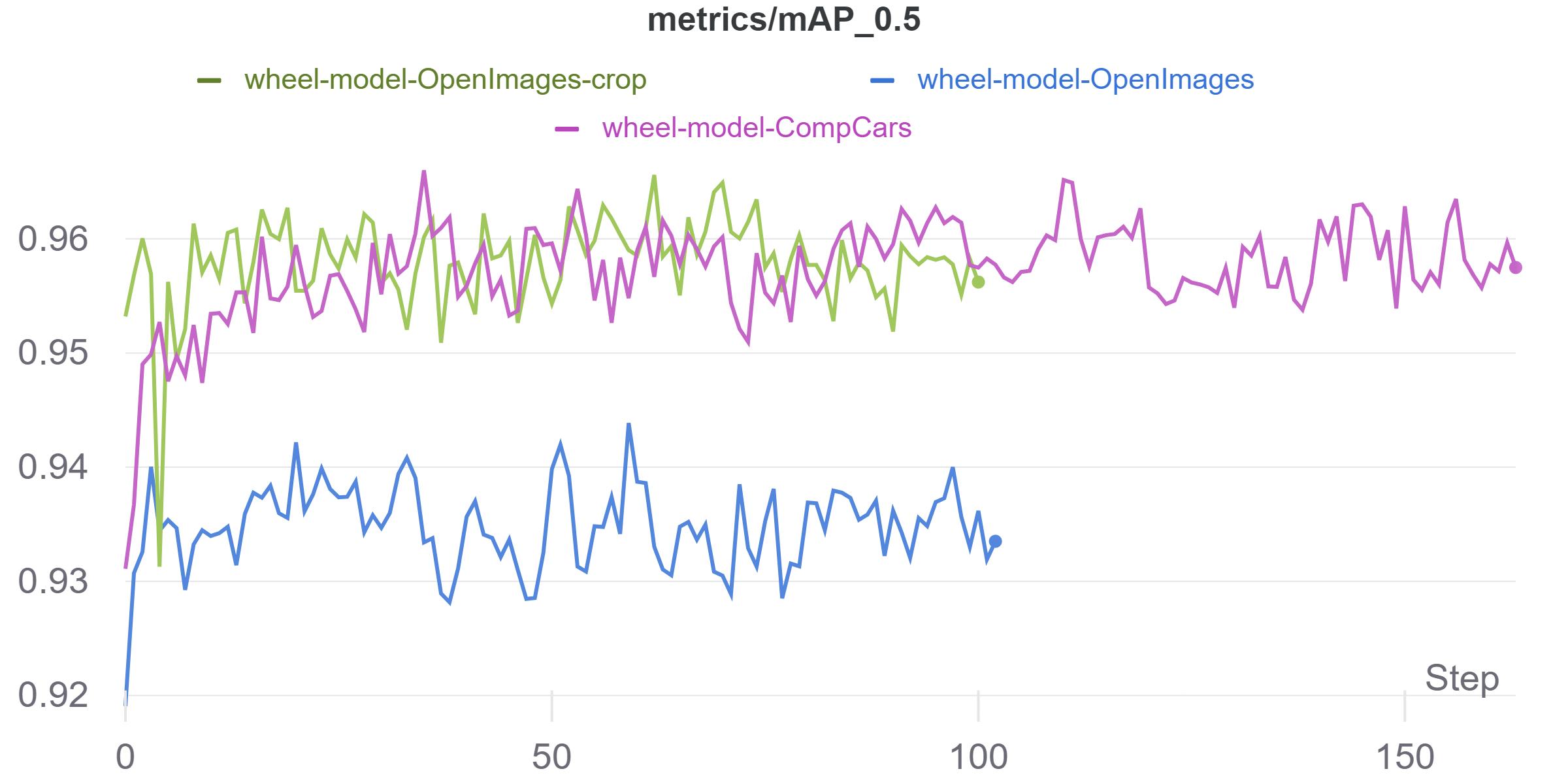}
    \caption{OpenImages-Crop Model mAP}
    \label{fig:fig39}
\end{figure}

\begin{figure}[H]
    \centering
    \includegraphics[width=\linewidth]{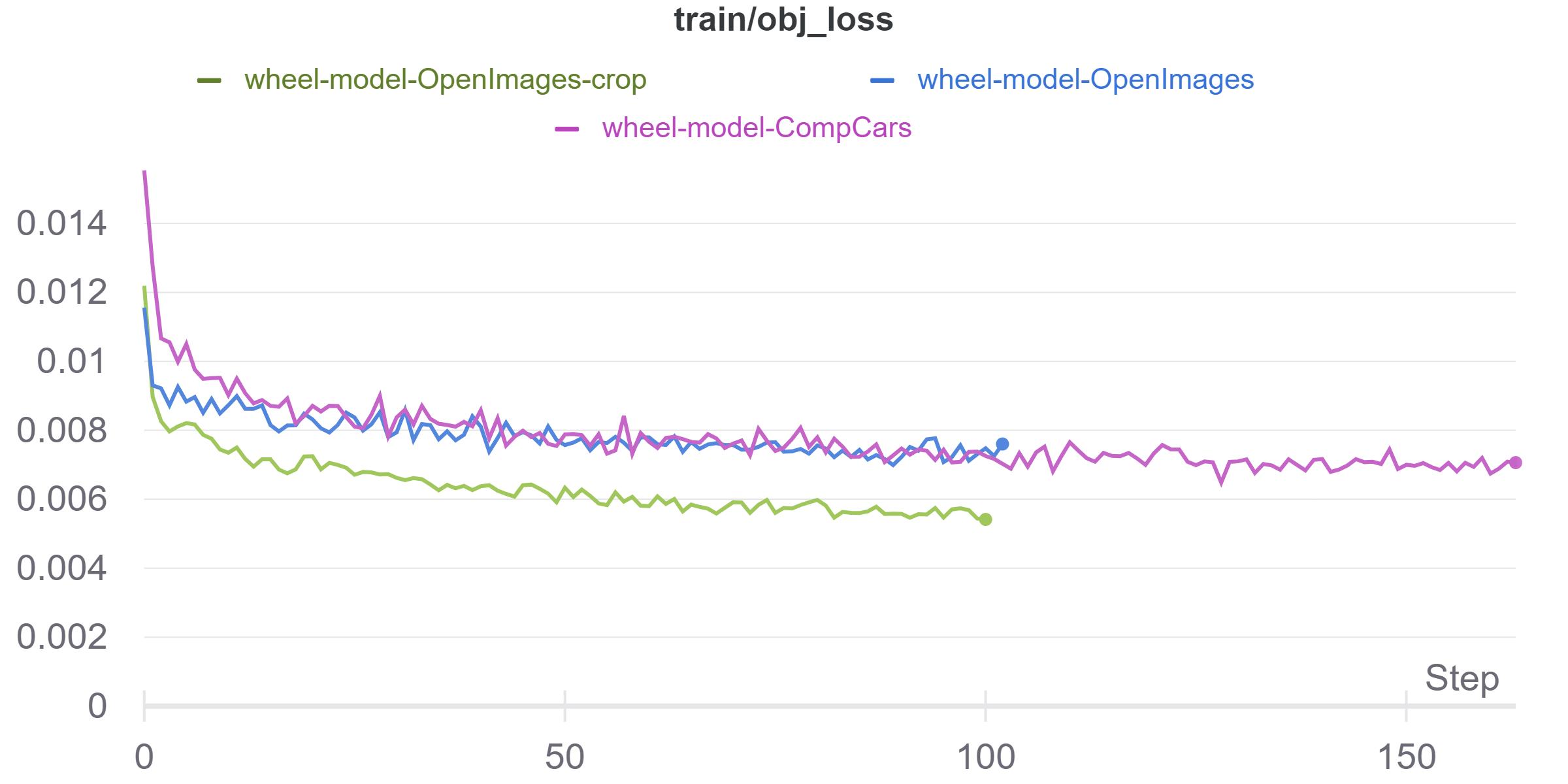}
    \caption{OpenImages-Crop Model Train Object Loss}
    \label{fig:fig40}
\end{figure}

\begin{figure}[H]
    \centering
    \includegraphics[width=\linewidth]{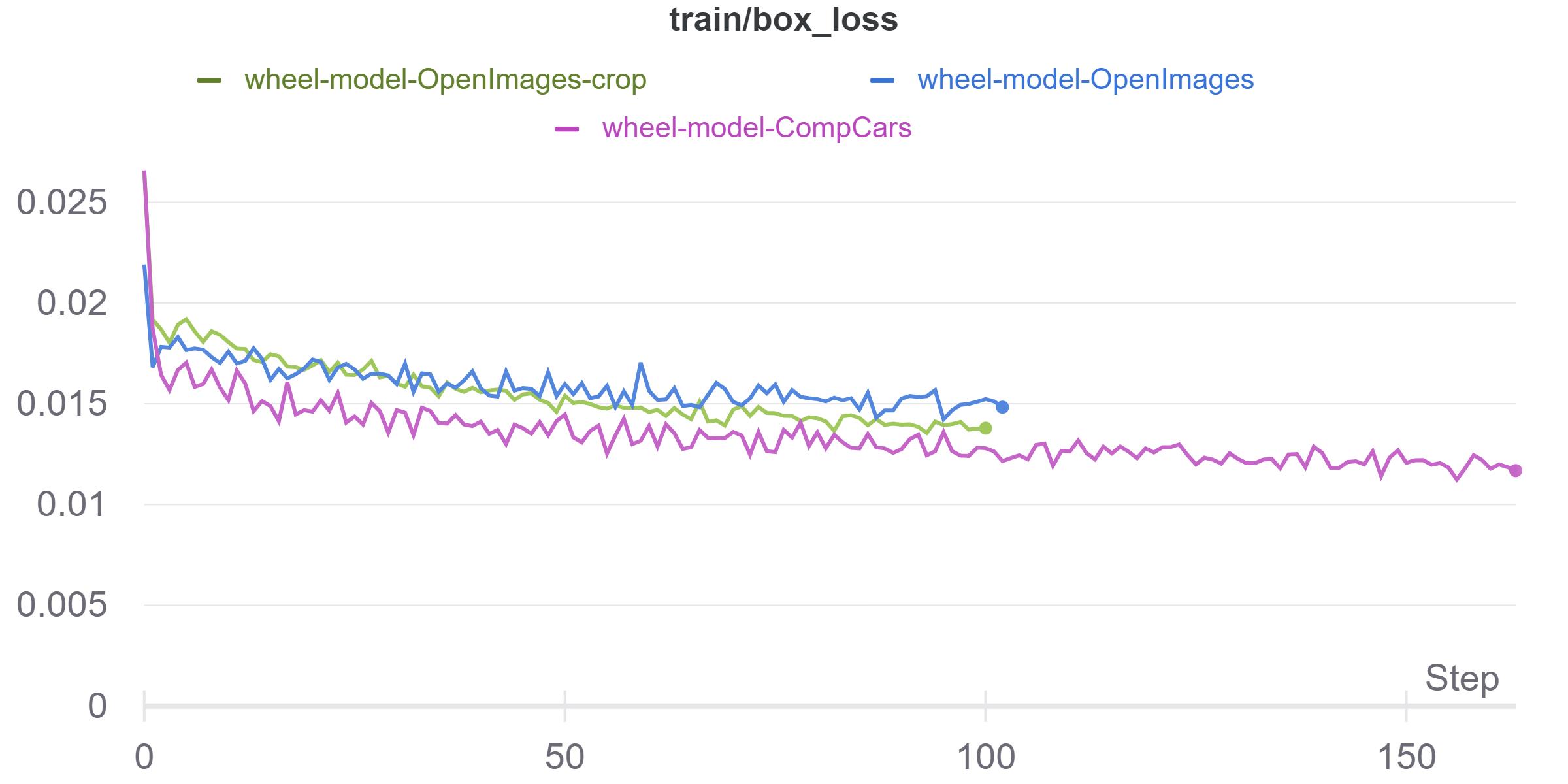}
    \caption{OpenImages-Crop Model Train Box Loss}
    \label{fig:fig41}
\end{figure}

\begin{figure}[H]
    \centering
    \includegraphics[width=\linewidth]{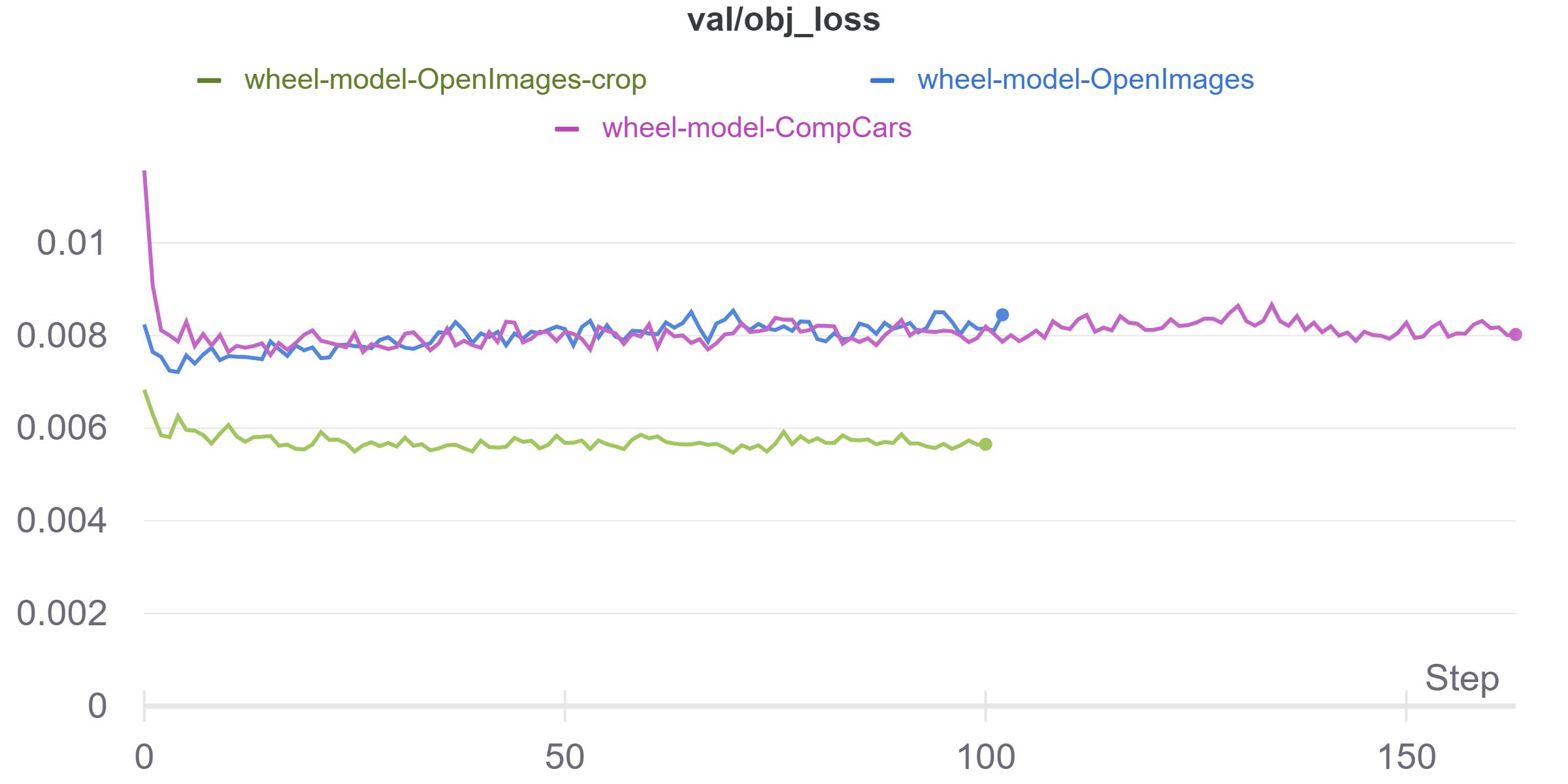}
    \caption{OpenImages-Crop Model Train Object Loss}
    \label{fig:fig42}
\end{figure}

\begin{figure}[H]
    \centering
    \includegraphics[width=\linewidth]{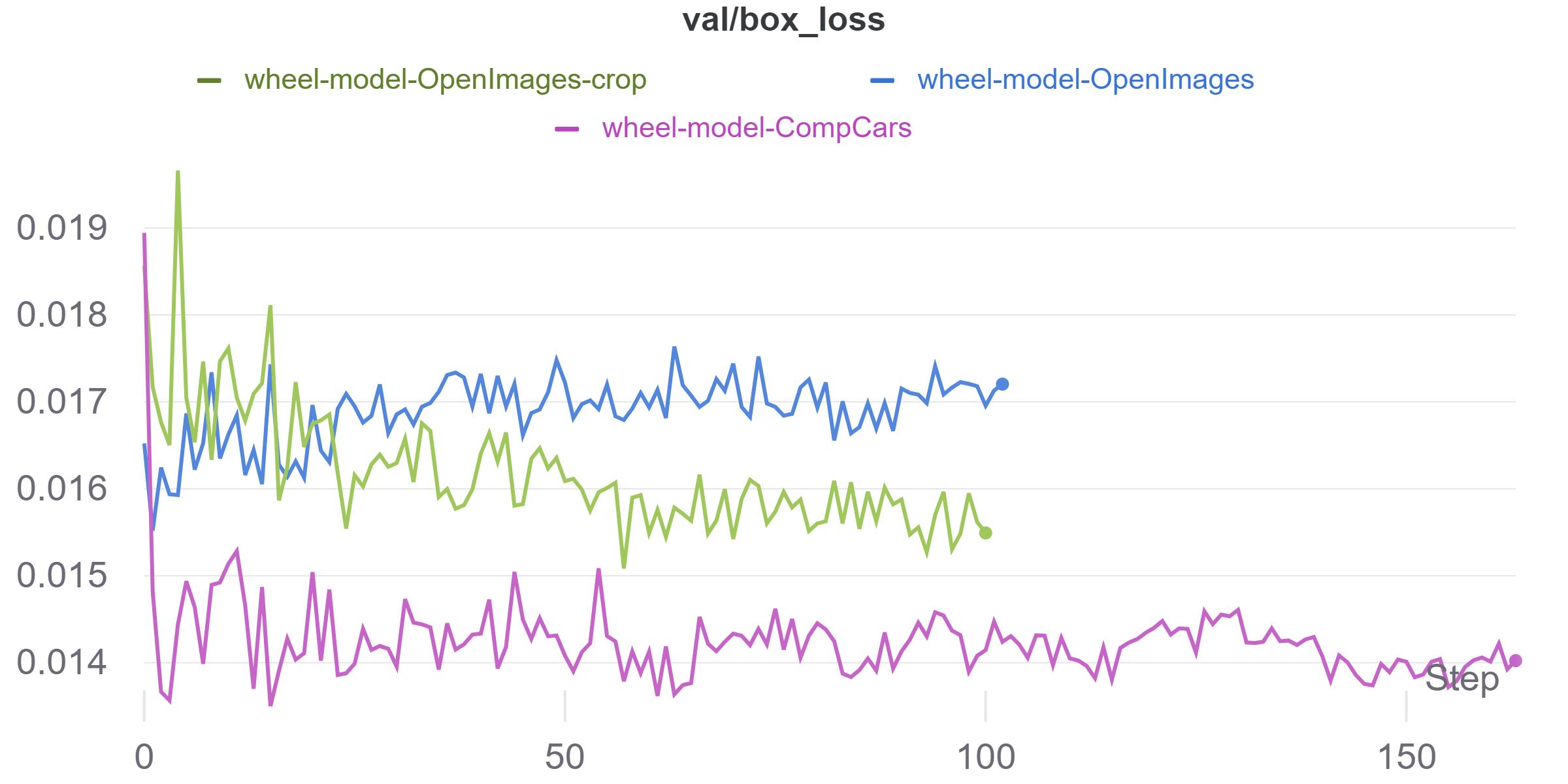}
    \caption{OpenImages-Crop Model Train Box Loss}
    \label{fig:fig43}
\end{figure}
We can clearly see that the model has improved with the cropped images from OpemImages dataset. This confirms the theory once more with evidence based on the metrics. 

\subsection{Final Model Evaluation}
Let's put everything into prospective and look the all metrics from all iterations.
\begin{figure}[H]
    \centering
    \includegraphics[width=\linewidth]{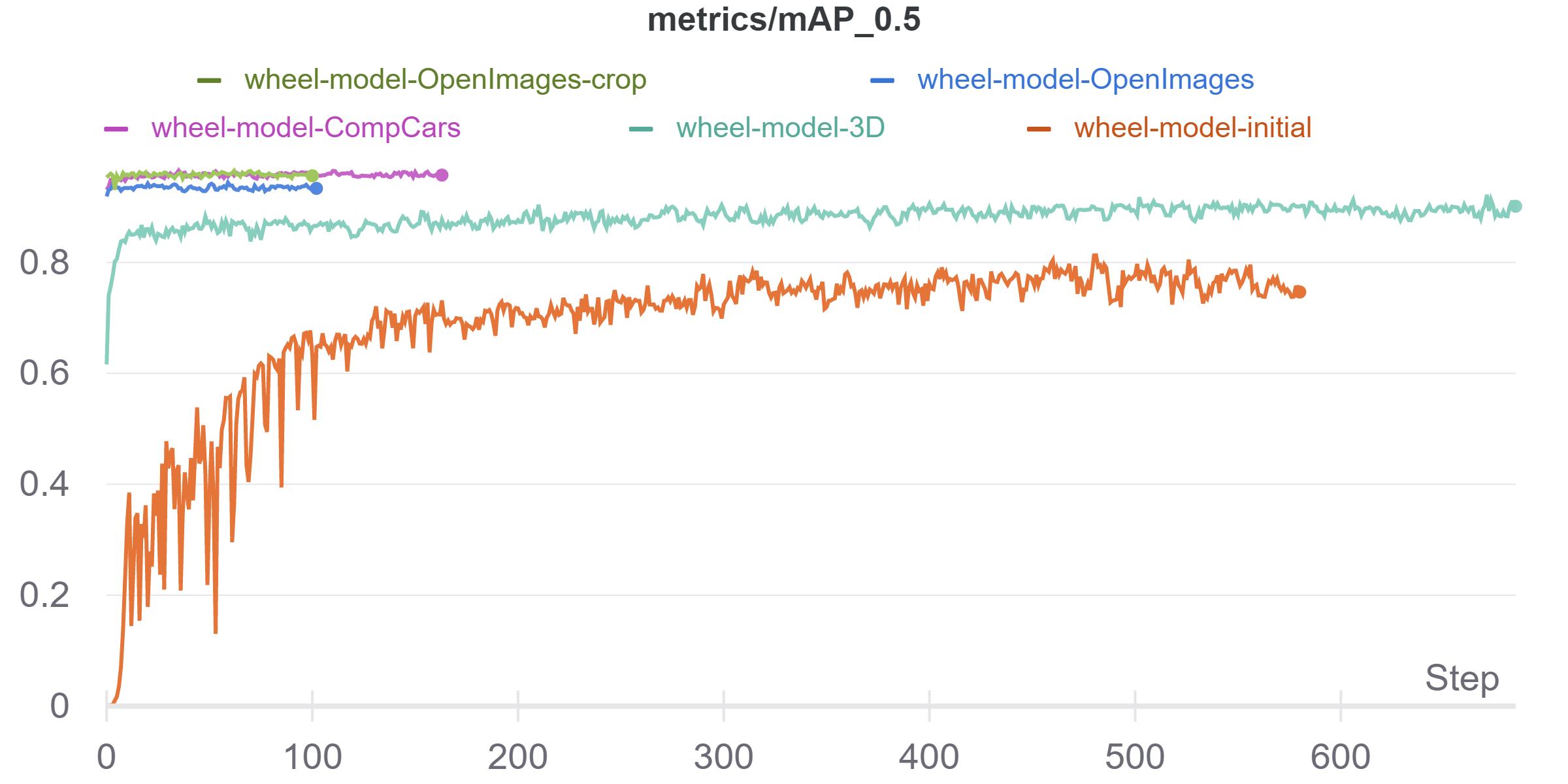}
    \caption{Wheel Model All mAP}
    \label{fig:fig44}
\end{figure}
\begin{figure}[H]
    \centering
    \includegraphics[width=\linewidth]{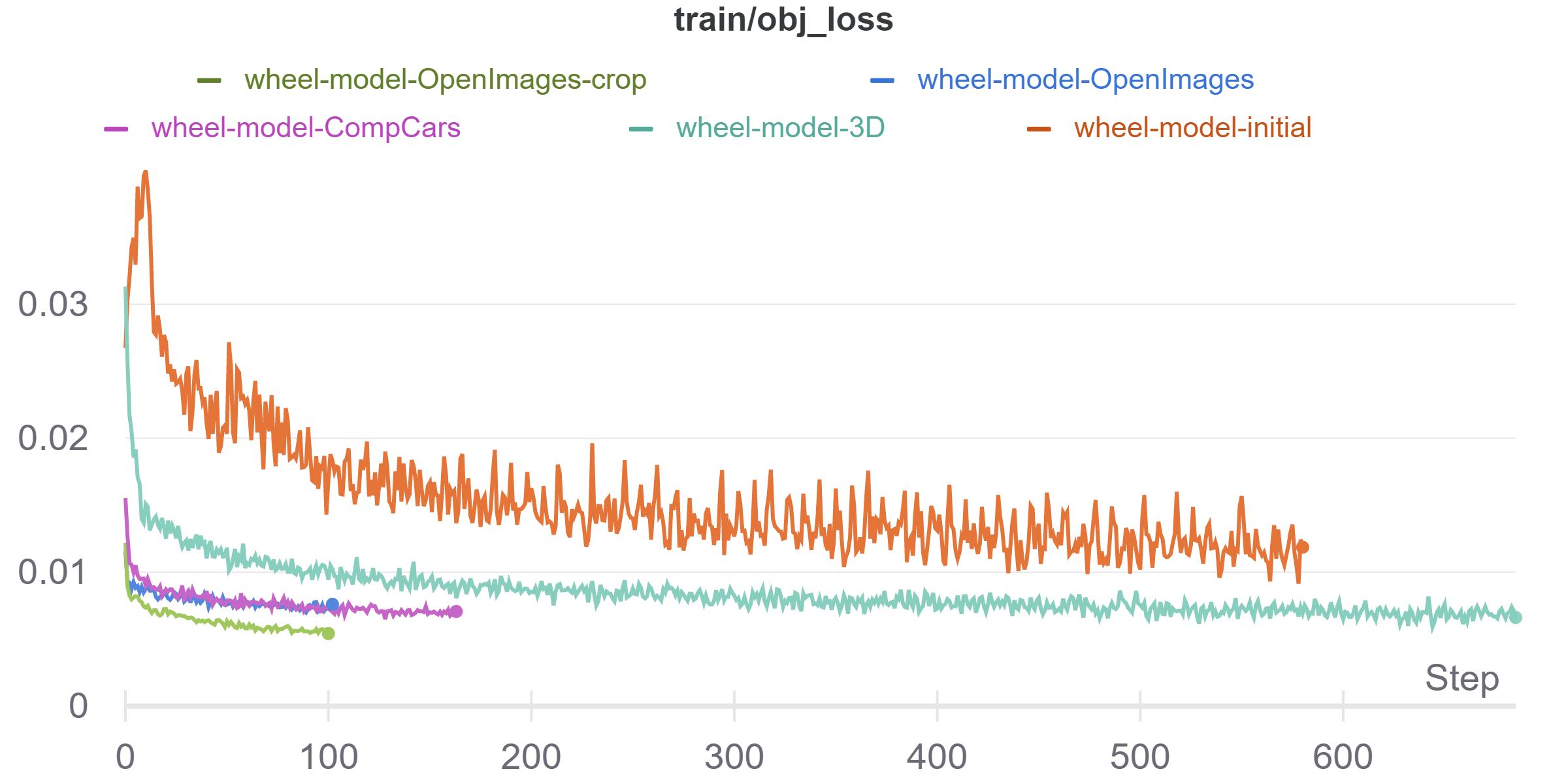}
    \caption{Wheel Model Train Object Loss}
    \label{fig:fig45}
\end{figure}
\begin{figure}[H]
    \centering
    \includegraphics[width=\linewidth]{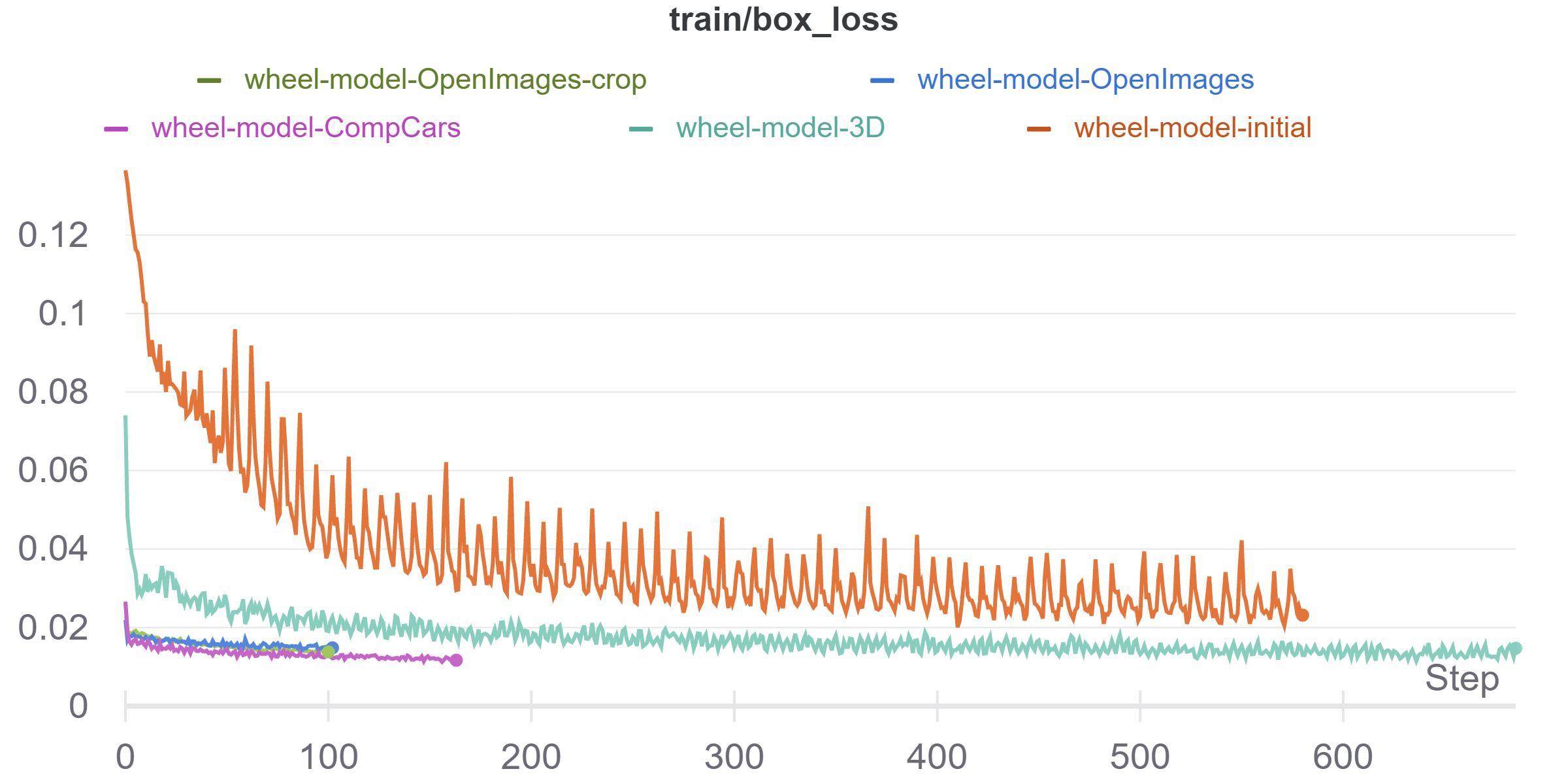}
    \caption{Wheel Model Train Box Loss}
    \label{fig:fig46}
\end{figure}
\begin{figure}[H]
    \centering
    \includegraphics[width=\linewidth]{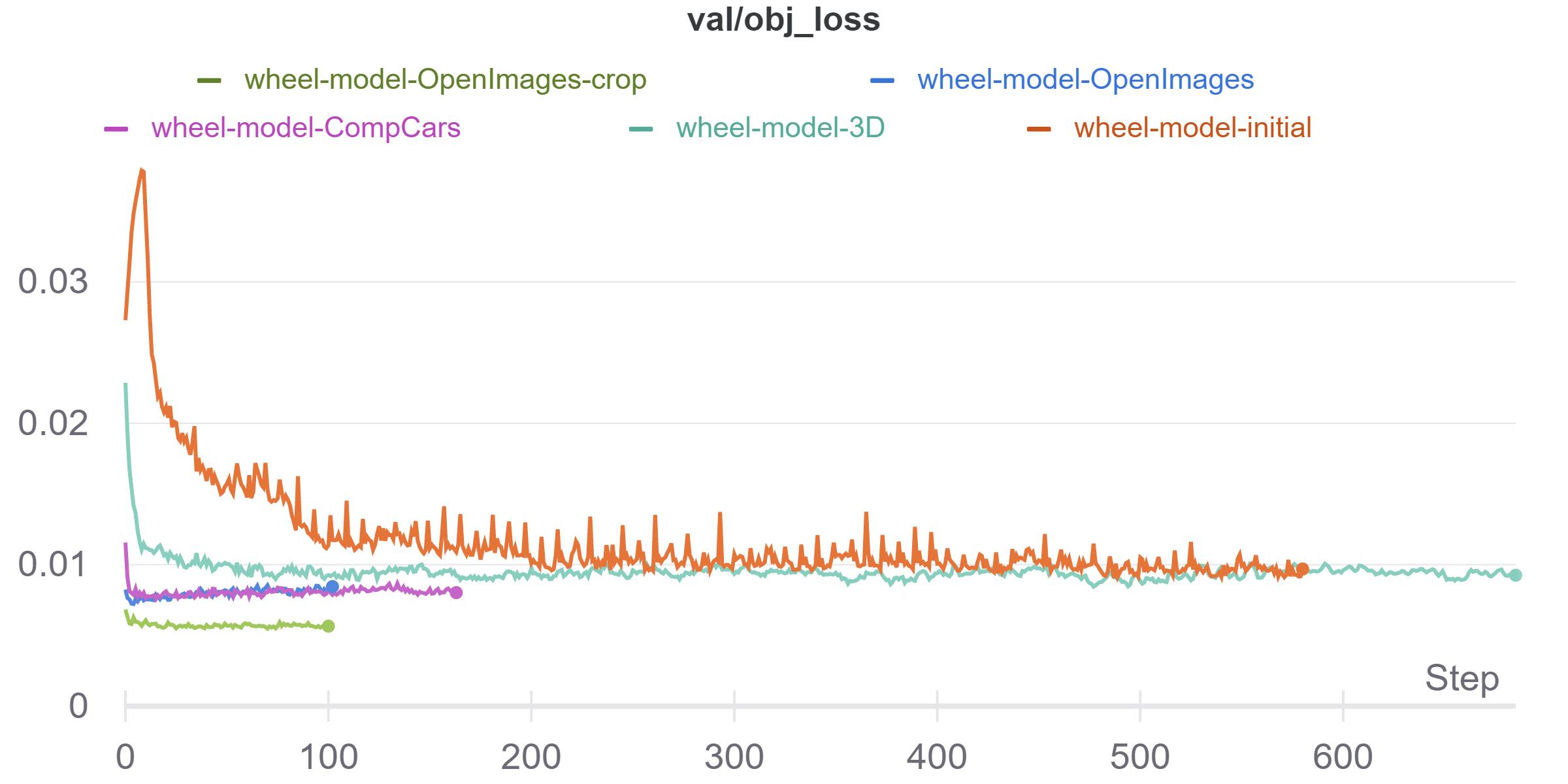}
    \caption{Wheel Model Validation Object Loss}
    \label{fig:fig47}
\end{figure}
\begin{figure}[H]
    \centering
    \includegraphics[width=\linewidth]{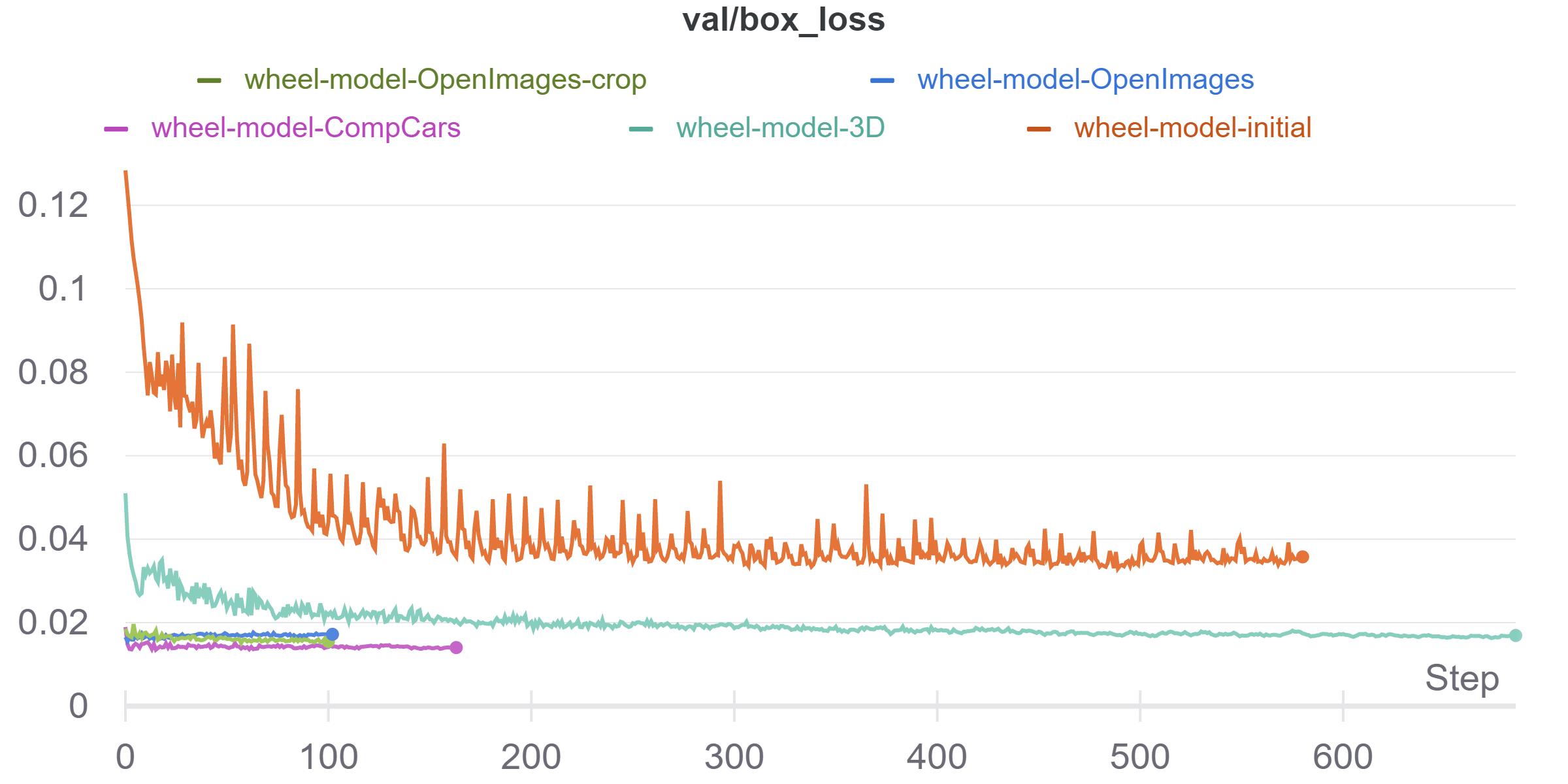}
    \caption{Wheel Model Validation Box Loss}
    \label{fig:fig48}
\end{figure}

Based on the metrics evaluation, the model seem to be in a good stage. It's time to test with image samples that resemble the final use case of the model. Based on the detection testing results, the model has performed very well and the images are showing in the Final Model Testing Results section.

\section{Limitations and Future Improvements}
The main limitation of this model is the input square size. For the final inference, you should try to stay close to a square image as possible. The advice here is to build a model that is optimized for the final deployment image size and transfer the weights from this model to your final optimized model. Future work would include the ground truth wheel labels extracted from 3D simulation such as CARLA \cite{b11} or other type of 3D simulator based on the type of objects of interest. Future improvement or addition to this work is the ability to provide semantic segmentation for the detector.

\section{Conclusion}
In conclusion, the vehicle wheel detector model has reached a reliable phase that is a good starting point for further improvement for a final deployment use case. We can clearly see the benefit of an iterative analytical approach of building a deep neural network model. Crafting the model in an iterative approach can give the insights needed to correct mistakes early on in the model development. It's important to try to understand how the model behavior changes with addition of samples and how metrics and visual validation can be a key to create a reliable model. The guideline proposed to create the vehicle wheel detector is intended to be used as a future reference for creating any kind of detector.

\section{Final Model Testing Results}
\begin{figure}[H]
    \centering
    \includegraphics[width=\linewidth]{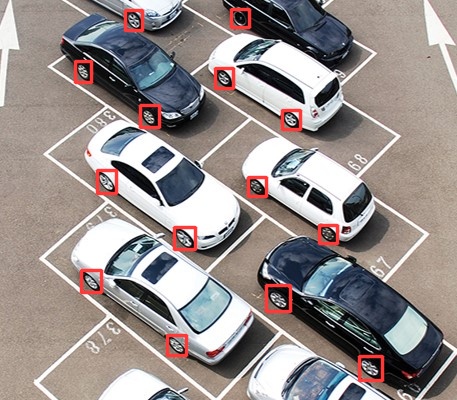}
    \caption{Final Model Test 1}
    \label{fig:test1}
\end{figure}
\begin{figure}[H]
    \centering
    \includegraphics[width=\linewidth]{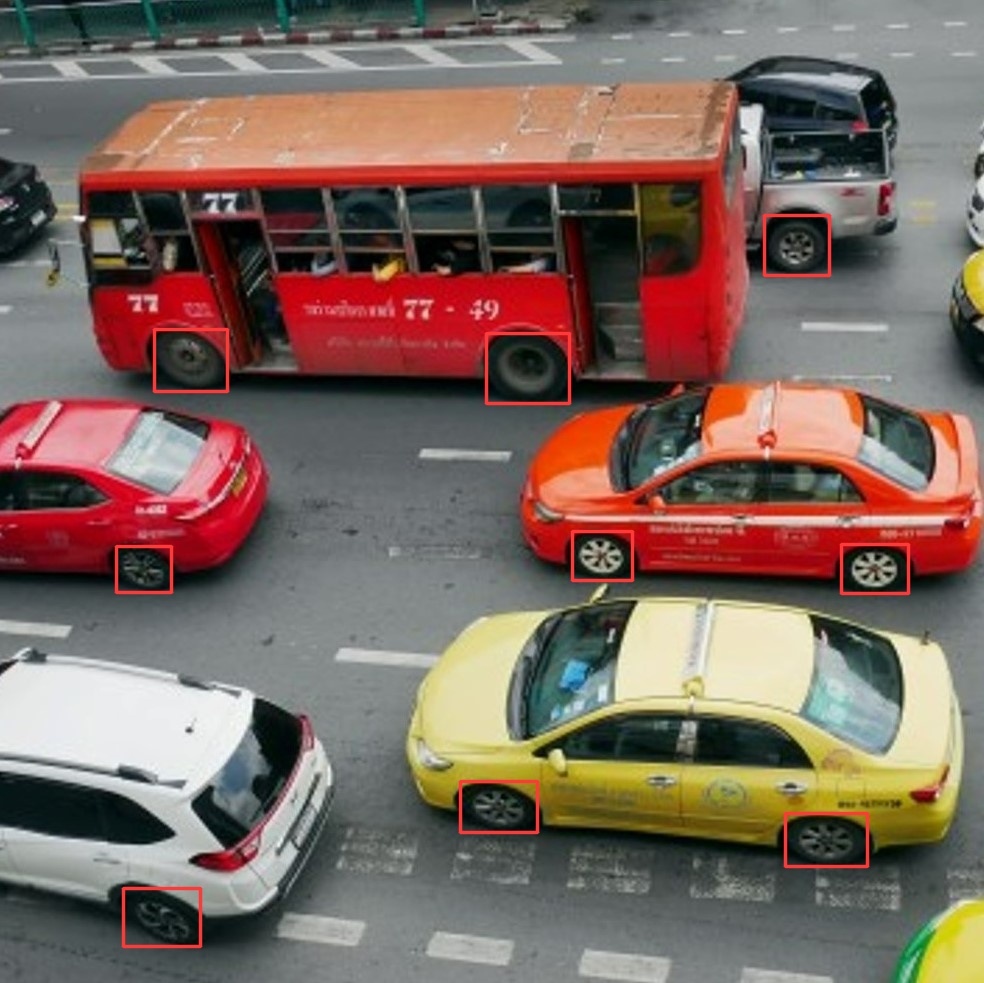}
    \caption{Final Model Test 2}
    \label{fig:test2}
\end{figure}
\begin{figure}[H]
    \centering
    \includegraphics[width=\linewidth]{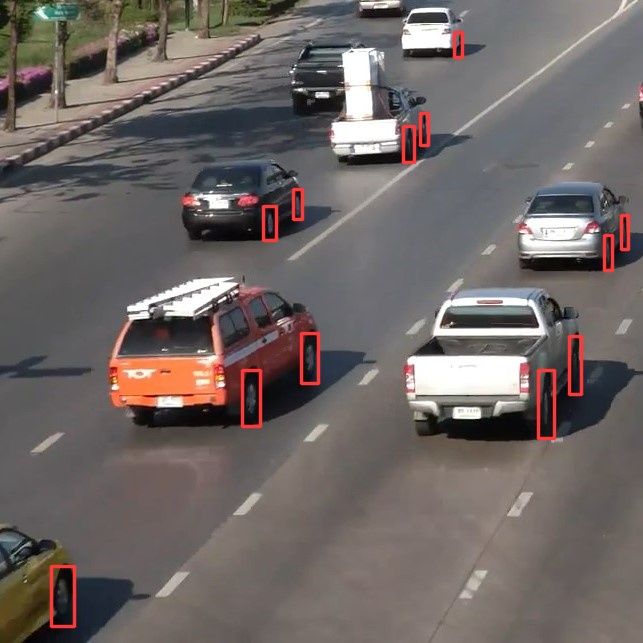}
    \caption{Final Model Test 3}
    \label{fig:test3}
\end{figure}
\begin{figure}[H]
    \centering
    \includegraphics[width=\linewidth]{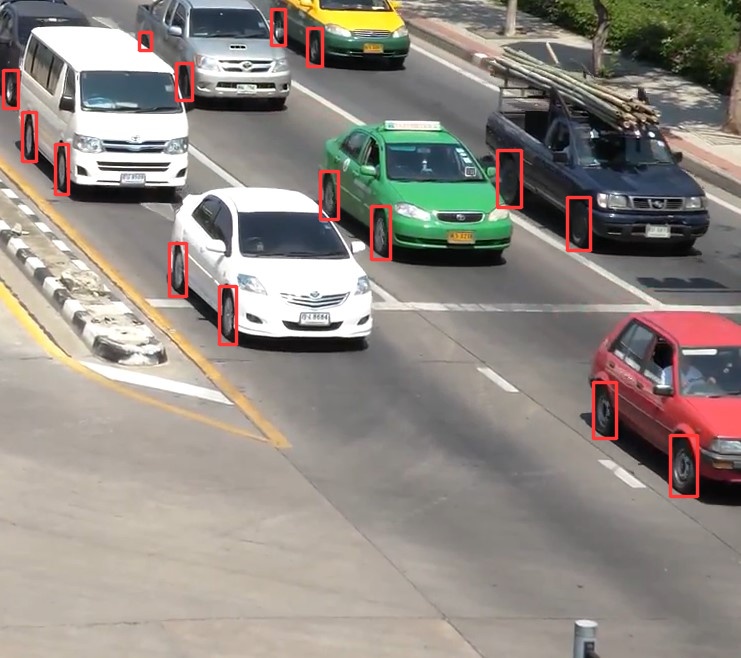}
    \caption{Final Model Test 4}
    \label{fig:test4}
\end{figure}
\begin{figure}[H]
    \centering
    \includegraphics[width=\linewidth]{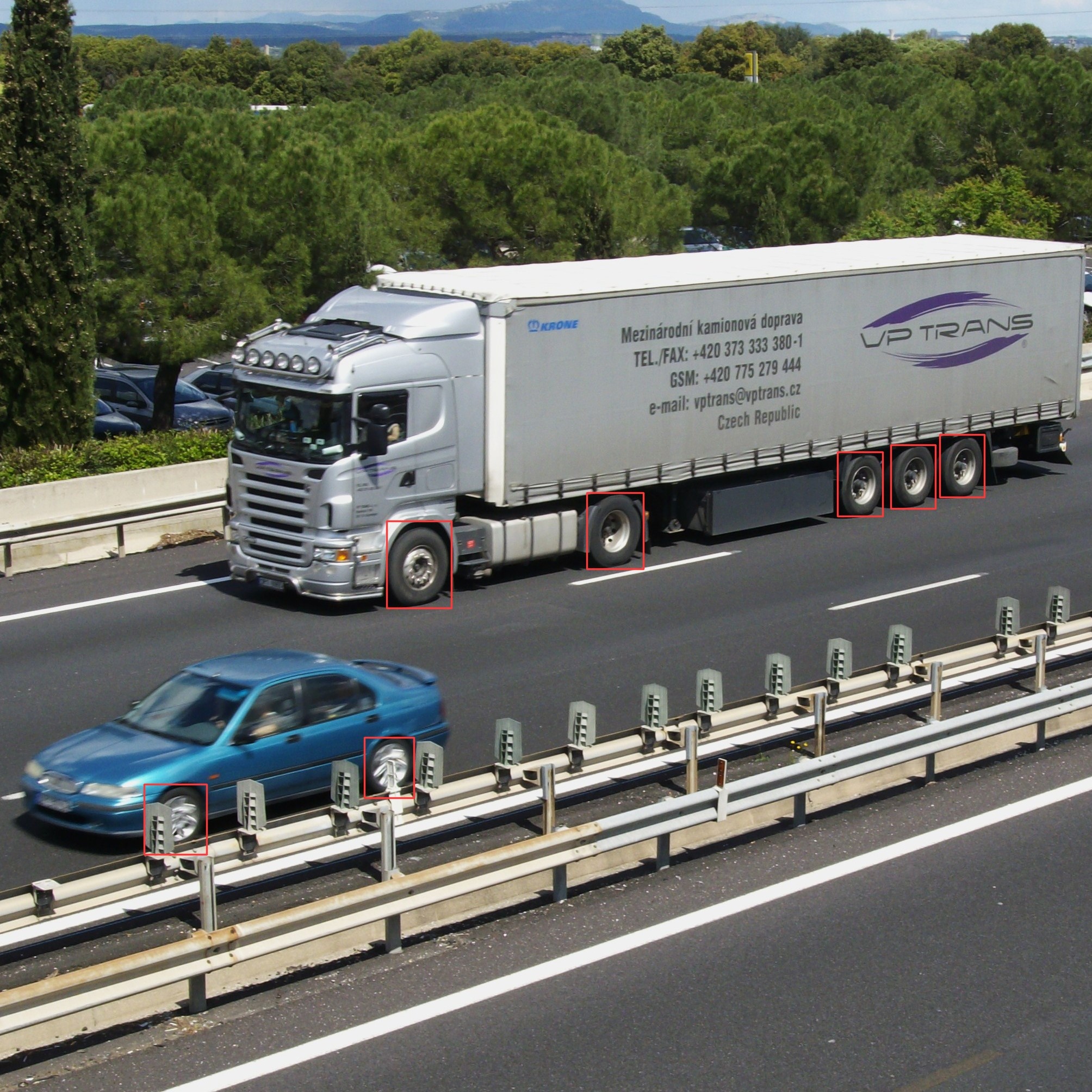}
    \caption{Final Model Test 5}
    \label{fig:test5}
\end{figure}
\begin{figure}[H]
    \centering
    \includegraphics[width=\linewidth]{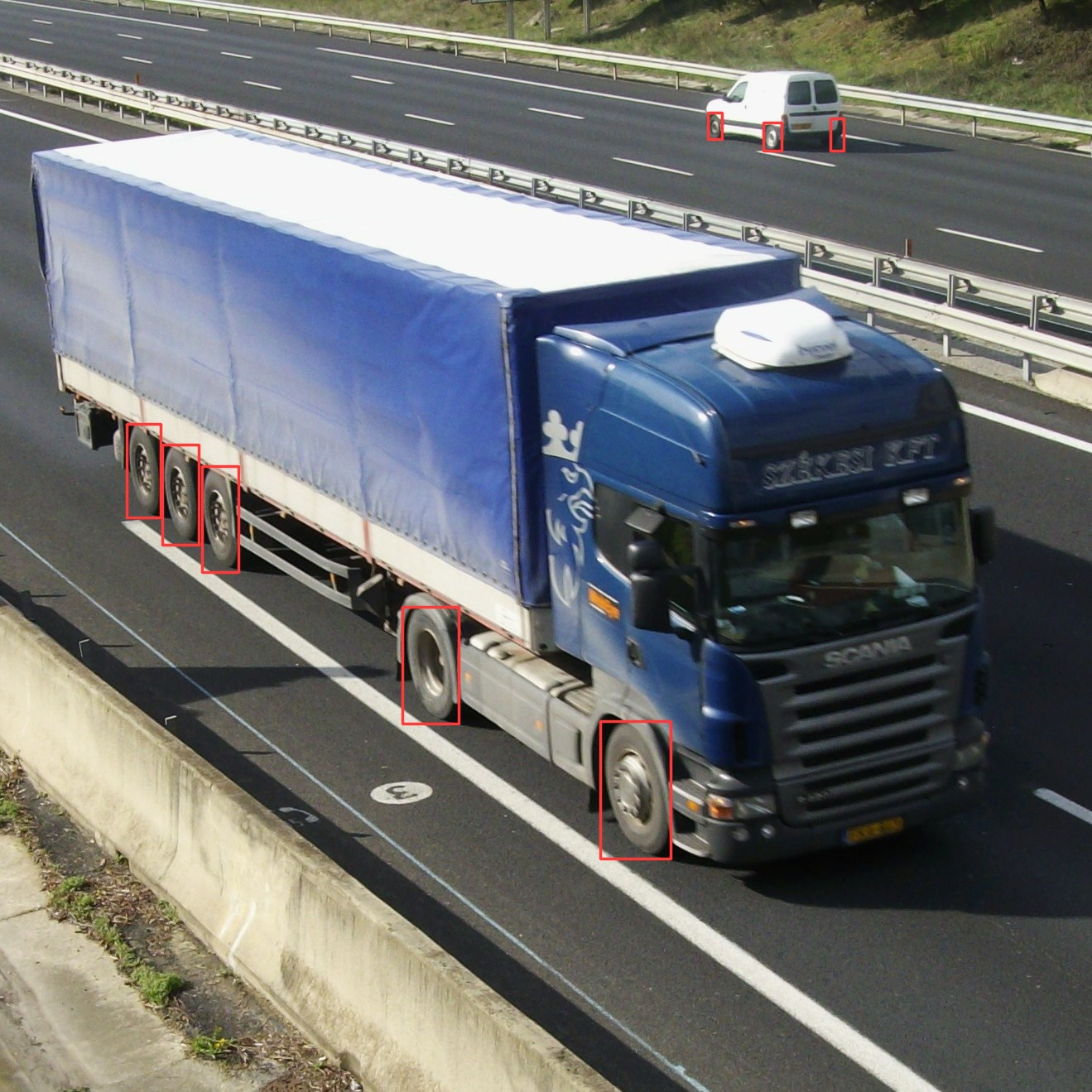}
    \caption{Final Model Test 6}
    \label{fig:test6}
\end{figure}
\begin{figure}[H]
    \centering
    \includegraphics[width=\linewidth]{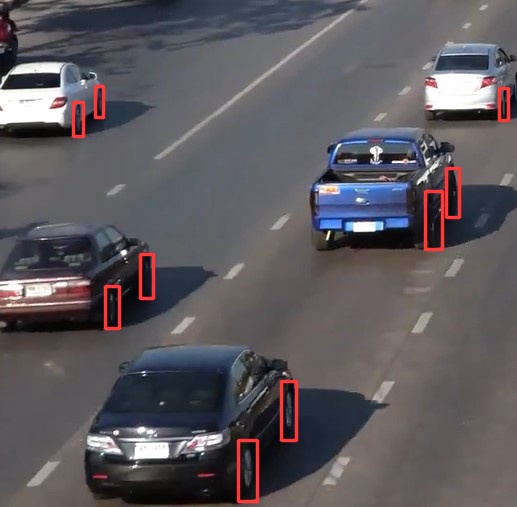}
    \caption{Final Model Test 7}
    \label{fig:test7}
\end{figure}
\begin{figure}[H]
    \centering
    \includegraphics[width=\linewidth]{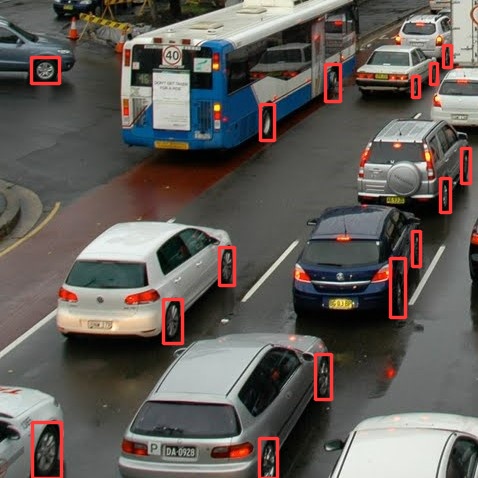}
    \caption{Final Model Test 8}
    \label{fig:test8}
\end{figure}
\begin{figure}[H]
    \centering
    \includegraphics[width=\linewidth]{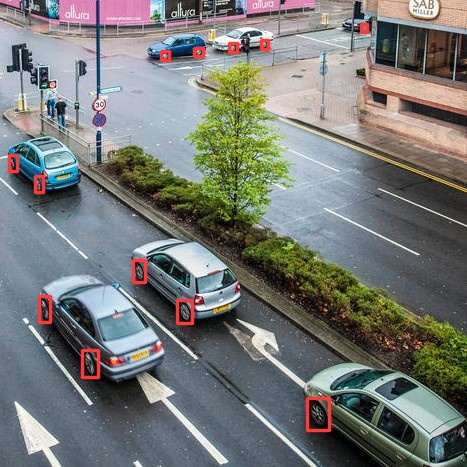}
    \caption{Final Model Test 9}
    \label{fig:test9}
\end{figure}
\begin{figure}[H]
    \centering
    \includegraphics[width=\linewidth]{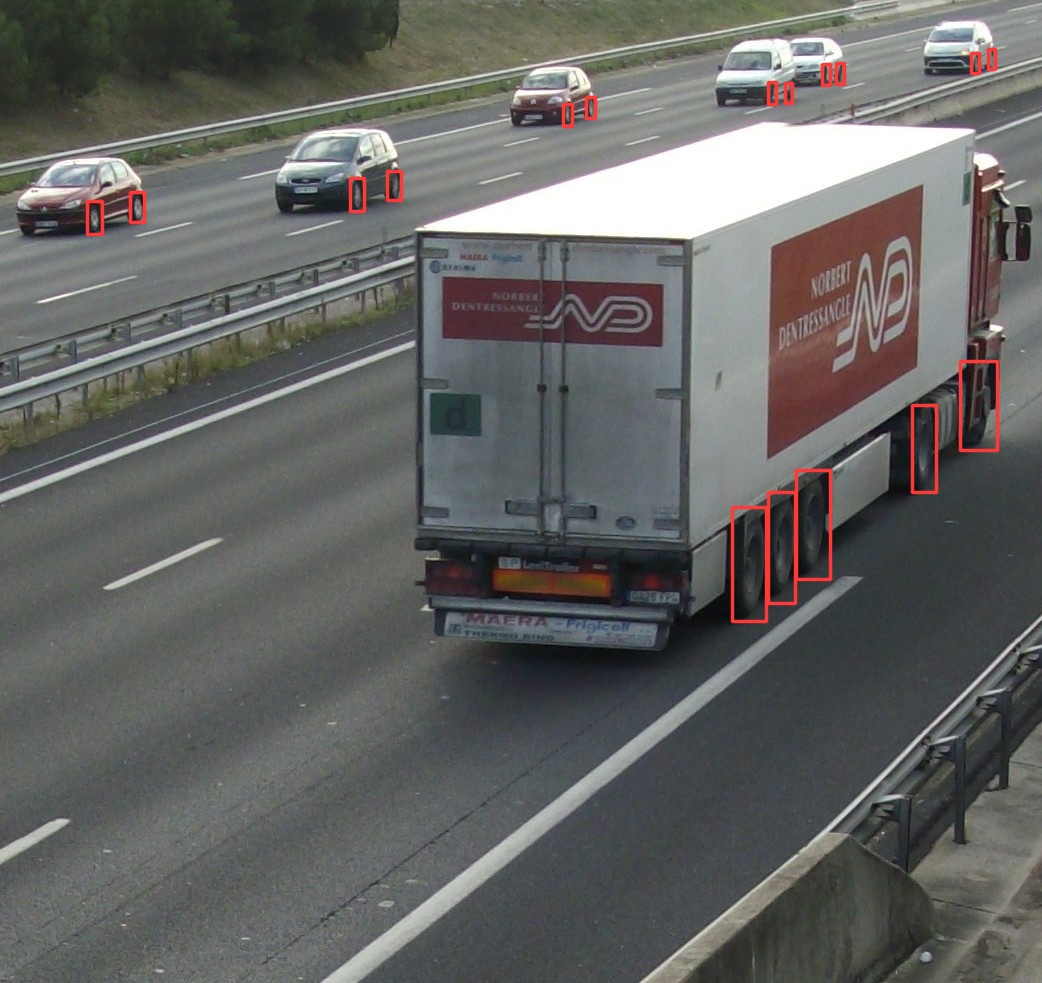}
    \caption{Final Model Test 10}
    \label{fig:test10}
\end{figure}
\begin{figure}[H]
    \centering
    \includegraphics[width=\linewidth]{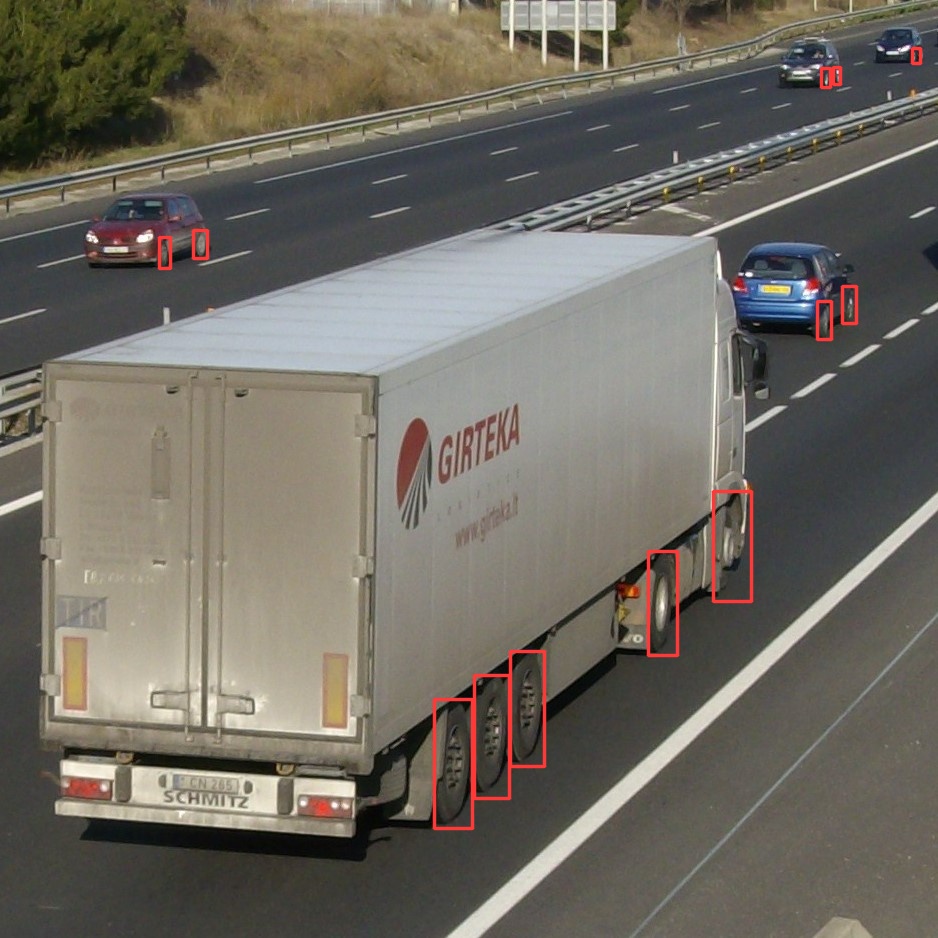}
    \caption{Final Model Test 11}
    \label{fig:test11}
\end{figure}
\begin{figure}[H]
    \centering
    \includegraphics[width=\linewidth]{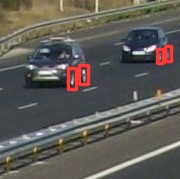}
    \caption{Final Model Test 12}
    \label{fig:test12}
\end{figure}

\vspace{12pt}
\color{red}

\begin{thebibliography}{00}
\bibitem{b1} J. Redmon, S. Divvala, R. Girshick and A. Farhadi, "You Only Look Once: Unified, Real-Time Object Detection," 2016 IEEE Conference on Computer Vision and Pattern Recognition (CVPR), 2016, pp. 779-788, doi: 10.1109/CVPR.2016.91.
\bibitem{b2} Redmon, Joseph and Ali Farhadi. “YOLO9000: Better, Faster, Stronger.” 2017 IEEE Conference on Computer Vision and Pattern Recognition (CVPR) (2017): 6517-6525.
\bibitem{b3} Redmon, Joseph and Ali Farhadi. “YOLOv3: An Incremental Improvement.” ArXiv abs/1804.02767 (2018): n. pag.
\bibitem{b4}Bochkovskiy, Alexey et al. “YOLOv4: Optimal Speed and Accuracy of Object Detection.” ArXiv abs/2004.10934 (2020): n. pag.
\bibitem{b5}YOLOv5 in PyTorch  https://github.com/ultralytics/yolov5
\bibitem{b6}Xu, Renjie et al. “A Forest Fire Detection System Based on Ensemble Learning.” Forests 12.2 (2021): 217. Crossref. Web.
\bibitem{b7} LabelImg is a graphical image annotation tool and label object bounding boxes in images https://github.com/tzutalin/labelImg
\bibitem{b8} Hum3D.com, a ParaART, LLC company
\bibitem{b9} The Comprehensive Cars (CompCars) dataset 
\bibitem{b10} Open Images Dataset V6 
\bibitem{b11} CARLA https://carla.org/
\end{thebibliography}
\end{document}